\documentclass{bmvc2k}

\title{Provably Contractive and High-Quality \\ Denoisers for Convergent Restoration}

\addauthor{Shubhi Shukla}{ee25s079@smail.iitm.ac.in}{1}
\addauthor{Pravin Nair}{pravinnair@ee.iitm.ac.in}{1}

\addinstitution{
Department of Electrical Engineering\\
Indian Institute of Technology Madras\\
Chennai, India
}

\runninghead{Shukla, Nair}{Contractive Denoisers for Convergent Restoration}

\usepackage{amsmath,amssymb,amsthm}
\usepackage{booktabs}
\usepackage{multirow}
\usepackage{graphicx}
\usepackage{listings}
\usepackage{orcidlink}
\usepackage{xcolor}
\usepackage{multirow}

\def\x{\boldsymbol{x}}

\def\y{\boldsymbol{y}}

\def\z{\boldsymbol{z}}

\usepackage{placeins} 
\def\A{\mathbf{A}}
\def\W{\mathbf{W}}

\def\B{\mathbf{B}}

\def\Z{\mathbf{Z}}
\def\D{\mathrm{D}}

\def\T{\mathrm{T}}
\def\K{\mathrm{K}}
\def\R{\mathbb{R}}
\def\S{\mathbf{S}}
\def\U{\mathrm{U}}

\def\V{\mathrm{V}}

\def\prox{\mathrm{Prox}}

\newcommand*{\tri}{\Lambda}
\newcommand{\norm}[1]{\left\lVert#1\right\rVert}

\DeclareMathOperator*{\argmin}{\arg\!\min}
\renewcommand{\Re}{\mathbb{R}}

\newtheorem{lemma}{Lemma}
\newtheorem{proposition}{Proposition}

\theoremstyle{definition}
\newtheorem{definition}{Definition}

\usepackage{booktabs}
\usepackage{multirow}
\theoremstyle{remark}

\begin{document}

\maketitle

\begin{abstract}
Image restoration, the recovery of clean images from degraded measurements, has applications in various domains like surveillance, defense, and medical imaging. While modern convolutional and attention-based networks achieve strong restoration quality, they typically lack explicit stability guarantees under small input perturbations. In this work, we develop provably contractive denoiser networks with global Lipschitz constant smaller than one. Our design composes proximal layers obtained from unfolding techniques, with Lipschitz-controlled convolutional refinements. By contractivity, the proposed denoiser guarantees that input perturbations of strength $\|\delta\|\leq \varepsilon$ induce at most $\varepsilon$ change at the output, whereas strong unconstrained baselines such as DnCNN and Restormer can exhibit larger deviations under the same perturbations. On image denoising, our model substantially improves over prior provably 1-Lipschitz / contractive baselines and narrows the gap to strong unconstrained denoisers. In addition, when used as a prior in Plug-and-Play restoration, the proposed denoiser yields provable convergence and strong empirical performance. Overall, our results show that strict Lipschitz control need not preclude practically competitive restoration quality, and suggest that stable, verifiable restoration models can be achieved without a severe loss in performance. Codes and pretrained models are available at
\url{https://github.com/SHUBHI1553/Contractive-Denoisers}.
\end{abstract}

\section{Introduction}
Image restoration aims at estimating a clean image from degraded observations produced by sensing and acquisition pipelines. Typical degradations include additive noise (denoising), motion/defocus blur (deblurring), point-spread effects (deconvolution), loss of spatial resolution (superresolution), and missing samples (inpainting/demosaicking) \cite{ribes2008linear}. Classical methods pose restoration as:
\begin{equation}
\label{mainopt}
\underset{\x \in \Re^n}{\min} \  f(\x) + g(\x),
\end{equation}
where $f$ is data-fidelity term derived by forward model and $g$ is the prior/regularizer on the estimated image \cite{engl2015regularization}. For applications like deblurring, superresolution, inpainting, and compressed sensing \cite{dong2011image,jagatap2019algorithmic},  $f(\x)={\lVert \y - \A\x \rVert}^2$, where $\A \in \Re^{m \times n}$ is the measurement matrix and $\y$ the observation. Common choices of regularizer $g(x)$ include total variation and related variational models, sparse representations, and patch-based nonlocal priors. The problem in Eq.~\eqref{mainopt} is usually solved by proximal or splitting  algorithms~\cite{ParikhBoyd2014Proximal,ROF92,CombettesPesquet2011}. These methods either assume that the degradation operator $\A$ is known (non-blind restoration), or require an additional procedure to estimate $\A$ from the data (blind restoration).

\begin{figure*}[t]
\begin{center}
\setlength{\tabcolsep}{4pt}
\renewcommand{\arraystretch}{1.0}
\scriptsize

\begin{tabular}{c c c c}

\multirow{3}{*}{
\begin{tabular}{c}
\includegraphics[width=0.22\linewidth]{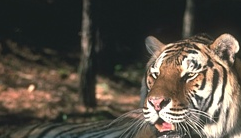} \\
Reference
\end{tabular}
}
&
\includegraphics[width=0.22\linewidth]{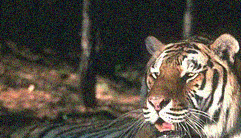} &
\includegraphics[width=0.22\linewidth]{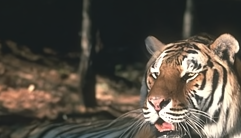} &
\includegraphics[width=0.22\linewidth]{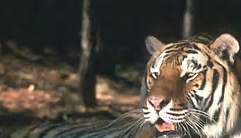} \\

&
Noisy ($24.74$) &
Restormer ($34.2$) &
Ours ($33.2$) \\[6pt]

&
\includegraphics[width=0.22\linewidth]{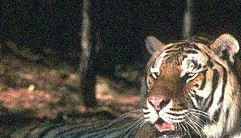} &
\includegraphics[width=0.22\linewidth]{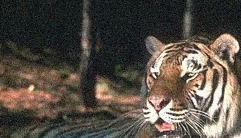} &
\includegraphics[width=0.22\linewidth]{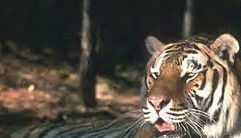} \\

&
Noisy+JPEG ($27.19$) &
Restormer ($27.9$) &
Ours ($32.2$)

\end{tabular}
\end{center}
\caption{JPEG robustness at $90\%$ quality. Our contractive denoiser remains stable, while Restormer's denoising quality drops under mild compression. PSNR values are shown in parentheses.}
\label{fig:jpeg_robustness}
\end{figure*}

\begin{figure*}[t]
\begin{center}
\setlength{\tabcolsep}{4pt}
\renewcommand{\arraystretch}{1.0}
\scriptsize

\begin{tabular}{c c c c}

\multirow{3}{*}{
\begin{tabular}{c}
\includegraphics[width=0.22\linewidth]{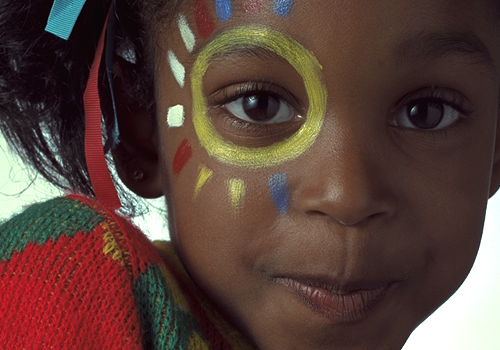} \\
Reference
\end{tabular}
}
&
\includegraphics[width=0.22\linewidth]{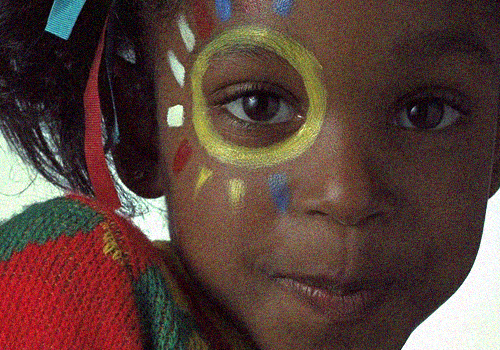} &
\includegraphics[width=0.22\linewidth]{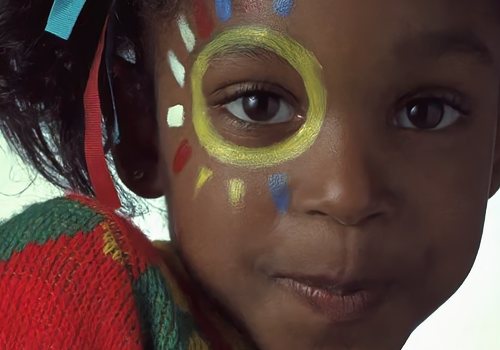} &
\includegraphics[width=0.22\linewidth]{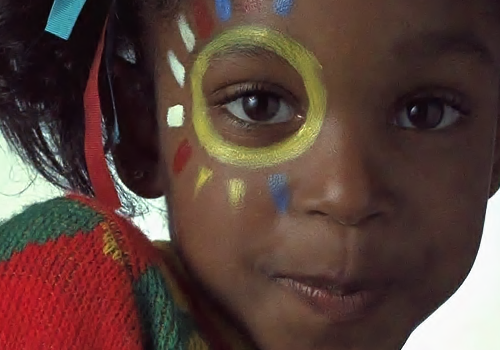} \\

&
Noisy ($24.83$) &
DnCNN ($34.75$) &
Ours ($34.01$) \\[6pt]

&
\includegraphics[width=0.22\linewidth]{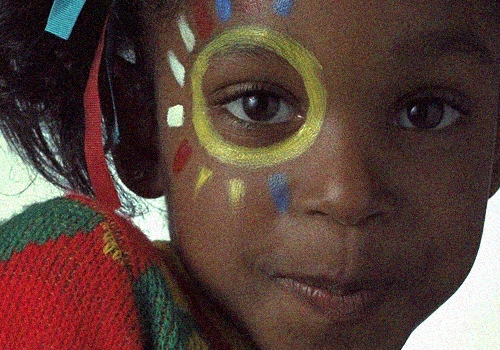} &
\includegraphics[width=0.22\linewidth]{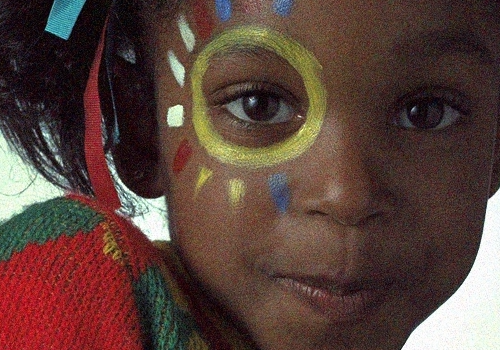} &
\includegraphics[width=0.22\linewidth]{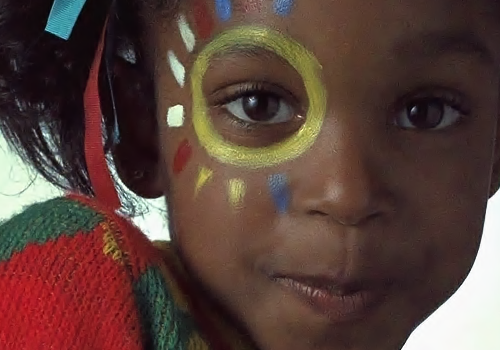} \\

&
Noisy+ChromaSub. ($27.88$) &
DnCNN ($28.71$) &
Ours ($33.47$)

\end{tabular}
\end{center}
\caption{Robustness to chroma subsampling. Our contractive denoiser retains stable reconstruction quality, while DnCNN suffers significant degradation under chroma perturbation. PSNR values are shown in parentheses.}
\label{fig:chroma_robustness}
\end{figure*}


With the advent of deep learning, one line of work replaces optimization in Eq.~\eqref{mainopt} with end-to-end supervised networks trained per task. Early CNNs demonstrated supervised superresolution and led to the first widely adopted learned networks in the form of SRCNN \cite{SRCNN14}. From deeper residual networks such as VDSR~\cite{Kim2016VDSR} and EDSR~\cite{Lim2017EDSR} to transformer-based SwinIR~\cite{Liang2021SwinIR}, architectural changes have improved superresolution. For denoising, {DnCNN} \cite{DnCNN17} and {FFDNet} \cite{FFDNet18} advanced accuracy and efficiency, while more recent methods, including iterative dynamic filtering \cite{kim2025idf} and self-calibrated variance-stabilizing transformations for real-world denoising \cite{herbreteau2025self}, further extend learned denoising. More recently, transformer-based and activation-free designs, such as Restormer \cite{Restormer22} and \mbox{NAFNet} \cite{NAFNet22}, deliver state-of-the-art performance for various restoration tasks. However, this entire class of end-to-end supervised models typically lacks explicit stability or Lipschitz guarantees.

A complementary line of work retains the inverse-problem structure in Eq.~\eqref{mainopt} via plug-and-play priors~\cite{Gregor2010LISTA,Monga2021UnrollingSurvey}. 
In \emph{plug-and-play} (PnP) \cite{PnP13} and \emph{regularization-by-denoising} (RED) \cite{RED17}, a trained denoiser is used as a learned operator within iterative algorithms like FBS and ADMM for \eqref{mainopt}, replacing the proximal step of the regularizer \(g\) \cite{CombettesPesquet2011,Boyd2011ADMM} (refer Sec.~\ref{sec:Preliminaries}). This induces an implicit prior on the images to be restored. Under standard conditions on the denoiser (e.g., nonexpansiveness ~\cite{Ryu2019PnPConvergence,nair2024averaged}, boundedness ~\cite{Chan2016PlugAndPlay}, or denoisers realized as proximal operator or gradient-step~\cite{hurault2022proximal,hurault2022gradient}), these iterative methods admit robustness and convergence guarantees. Empirically, PnP/RED achieve state-of-the-art reconstructions when the forward model is known or accurately estimated~\cite{renaud2024snore,ebner2024pnpconv,xu2024provably,nair2024averaged}.

\noindent\textbf{Motivation:}
Despite rapid progress in image restoration, stability under realistic input perturbations remains underexplored. Real acquisition pipelines inevitably introduce perturbations arising from sensor noise, optics, compression, and sampling. Yet many state-of-the-art restoration models are trained on curated supervised datasets, leaving their behavior under such perturbations unclear. Figures~\ref{fig:jpeg_robustness} and \ref{fig:chroma_robustness} illustrate this issue: even strong denoisers such as DnCNN and Restormer can be noticeably sensitive to small, common perturbations, including mild JPEG compression and chroma subsampling, leading to amplified artifacts and degraded denoising quality. This motivates the design of denoisers that remain stable under realistic perturbations while preserving structure and color fidelity. A second challenge arises in plug-and-play (PnP) restoration, where algorithmic convergence typically requires the denoiser to satisfy nonexpansive or contractive properties (see Sec.~\ref{sec:Preliminaries}). In practice, however, training denoisers that are both provably contractive and competitive with modern unconstrained architectures remains difficult. This leads to the central question addressed in this work: can we design provably contractive denoisers with sufficient expressive power to remain practically competitive, while also enabling convergent and high-quality restoration within PnP algorithms?

\noindent\textbf{Contributions:}
We show that provably contractive denoisers can approach the performance of strong unconstrained restoration models closely enough to be practically competitive. 
Our main contributions are:
\begin{itemize}
\item \textit{Contractive architecture.} We design an image-denoising layer in Sec.~\ref{subsec:Contr} that is provably contractive, identify its key components (gradient and proximal-wavelet blocks together with Lipschitz-controlled convolutions), and compose them into a contractive network with practical restoration performance.
\item \textit{Theory and empirical performance.} We prove contractivity of the proposed architecture and compare it against prior provably \(1\)-Lipschitz / contractive networks. Experiments in Sec.~\ref{sec:denoising} show that our method substantially improves over prior provable baselines and narrows the gap to strong recent unconstrained denoisers.
\item \textit{PnP restoration with guarantees.} When used as a prior in Plug-and-Play restoration (Sec.~\ref{sec:pnp}), our denoiser combines the benefits of contractivity and strong denoising quality, yielding provable convergence together with high-quality reconstructions.
\end{itemize}

\section{Related Work}

Image restoration has a broad literature spanning classical variational priors, supervised CNN/Transformer restoration, and plug-and-play/RED methods; we refer the reader to recent surveys for broader coverage
\cite{jiang2025survey,kamilov2023plug,tan2025pnpOverview}. Recently, generative and diffusion models have also been used as powerful priors for inverse problems and plug-and-play restoration 
\cite{graikos2022diffusion,zhu2023denoising,daras2024survey,park2026stochastic}. These methods are complementary to our work and often perform well under severe degradations; accordingly, we include SGPnP~\cite{park2026stochastic} and DiffPIR \cite{zhu2023denoising} as
generative PnP baselines. However, unlike our goal, they do not explicitly enforce a deterministic globally nonexpansive/contractive denoising operator for convergent PnP restoration. We therefore focus next on Lipschitz-controlled  models. Lipschitz-controlled neural networks are important in applications requiring stability, robustness, and convergence guarantees. In classification, Lipschitz constraints are closely tied to adversarial robustness, certified defenses, and invertible architectures  \cite{miyato2018spectral,behrmann2019invertible,Tsuzuku2018LMT,Gouk2021LipschitzSurvey}. In inverse problems, they are particularly useful in plug-and-play and operator-splitting methods, where nonexpansive or \(1\)-Lipschitz denoisers enable convergence guarantees \cite{Ryu2019PnPConvergence,nair2024averaged}. Relatedly, CRR-NN~\cite{goujon2023neural} learns a convex-ridge regularizer and uses gradient Lipschitz bounds for convergent restoration, rather than enforcing layer-wise contractivity of the denoiser. These motivations have led to a growing body of work on training Lipschitz-constrained networks.

Several works aim to train \(1\)-Lipschitz networks by constraining layer norms, but mostly without provable end-to-end guarantees. Spectral normalization (SN) constrains the largest singular value of linear and convolutional layers using a few power iterations during training \cite{miyato2018spectral,Ryu2019PnPConvergence,behrmann2019invertible}. However, the use of finite power iterations makes SN approximate and hence insufficient to certify an end-to-end \(1\)-Lipschitz network (see \cite{nair2024averaged} for a counterexample). Parseval networks \cite{Cisse2017Parseval} and orthogonality-regularized CNNs \cite{wang2020orthogonal} similarly promote orthogonal weights. While exact orthogonality would yield nonexpansive layers, in practice these constraints are typically imposed softly through regularization terms, resulting only in approximate Lipschitz control. More generally, heuristics like margin-based training and norm penalties improve stability, but they do not provide network-level guarantees \cite{Tsuzuku2018LMT,Gouk2021LipschitzSurvey}.

Provably \(1\)-Lipschitz models can instead be built by combining exactly normalized linear or convolutional layers with \(1\)-Lipschitz activations. GroupSort activations were introduced in \cite{Anil2019GroupSort} to construct end-to-end \(1\)-Lipschitz networks with improved expressivity, and recent learnable \(1\)-Lipschitz activation families further enhance approximation power while preserving the guarantee \cite{Ducotterd2024JMLR}. Exact frequency-domain normalization can enforce a convolutional layer to be \(1\)-Lipschitz, although the required singular-value computations incur substantial training-time overhead \cite{Sedghi2019Singular}. In the context of inverse problems, denoisers based on wavelet-based unrolled splitting have been shown to satisfy \(1\)-Lipschitz guarantees and thereby ensure plug-and-play convergence \cite{nair2024averaged}, but they do not yet match the restoration quality of unconstrained state-of-the-art models. However, provably \(1\)-Lipschitz architectures remain relatively scarce and have been developed primarily for classification \cite{Anil2019GroupSort,Tsuzuku2018LMT,Prach2024OneLipLayers}. Only a few recent works \cite{bredies2024learning,nair2024averaged} study such constrained models for inverse problems. Overall, while provably $1$-Lipschitz architectures offer strong guarantees, strict enforcement can reduce flexibility or increase cost, and it remains unclear whether they can match unconstrained restoration models. Our work addresses this question by investigating whether provable contractivity can be retained while substantially reducing the performance gap in restoration quality.

\section{Preliminaries}
\label{sec:Preliminaries}
In this section, we discuss the preliminary results required to explain our architecture. We start with the Lipschitz property of functions. Throughout this work, we use Euclidean norm.  
\begin{definition}[Lipschitz continuity]
Let $f:\mathbb{R}^n \to \mathbb{R}^m$ and let $\|\cdot\|$ denote the Euclidean norm.
We say $f$ is Lipschitz continuous if there exists $L \ge 0$ such that
$\| f(x) - f(y) \| \le L \, \| x - y \|$, for all $x,y \in \mathbb{R}^n$.
Such an $L$ is called the Lipschitz constant of $f$ and $f$ is said to be $L$-Lipschitz.
If this constant satisfies $L<1$, then $f$ is said to be \emph{contractive} or \emph{$L$-contractive}. If $L \le 1$, $f$ is said to be \emph{nonexpansive}.

\end{definition}
Recall the inverse problem in Eq.~\eqref{mainopt}. For a convex regularizer $g: \Re^n \to \Re \cup \{\infty\}$, the proximal operator is defined as
\begin{equation}
\label{prox}
\prox_{g}(\y) = \argmin_{\x \in \Re^n}\  \left\{\frac{1}{2} \norm{\x - \y}^2 +  g(\x)\right\}.
\end{equation}
The proximal operator is the denoiser induced by the regularizer $g$. Suppose $\y$ is a noisy observation of a clean image $\x_0$ under the additive white Gaussian noise (AWGN) model:
\begin{equation}
\label{den:forward}
\y = \x_0 + \boldsymbol{\eta}, \qquad \boldsymbol{\eta} \sim \mathcal{N}(0, \mathbf{I}),
\end{equation}
Assume prior density $p(\x) \propto \exp(-g(\x))$. Then
Equation~\eqref{prox} returns maximum a posteriori (MAP) estimate of $\x_0$ given $\y$~\cite{beck2009fast}. 

One widely used algorithm to solve the problem in Eq.~\eqref{mainopt} is Forward-Backward Splitting (FBS), which alternates between a gradient step on $f$ with a proximal step on $g$. In particular, the FBS update  is 
\begin{align}
\label{fbs}
{\z}_{k+1} &= {\x}_{k} -  \alpha \nabla \! f ({\x}_{k}), \nonumber \\
{\x}_{k+1} &= \prox_{\alpha g} (\z_{k+1}). 
\end{align}
where $\alpha > 0$  and $\x_0 \in \Re^n$. Since $\prox_{\alpha g}(\z_{k+1})$ in Eq.~\eqref{fbs} coincides with the MAP estimate of Gaussian denoising, with $g$ acting as the log-prior, \emph{plug-and-play} (PnP) methods, propose to replace $\prox_{\alpha g}$ in algorithms like FBS by a generic denoiser. In particular, the PnP variant of forward-backward splitting (PnP-FBS) performs
\begin{equation}
\label{eq:pnpfbs}
\x_{k+1} = \D\big(\x_k - \alpha \nabla f(\x_k)\big),
\end{equation}
where $\D$ is a denoising operator. Modern PnP and \emph{regularization by denoising} (RED) methods typically employ powerful learned denoisers as $\D$, enabling near state-of-the-art performance across a range of image restoration tasks. We next recall a result from \cite{nair2024averaged} that analyzes the FBS update, viewed as an operator, for the Gaussian denoising setting. 
\begin{lemma}[Single FBS iteration]
\label{thm:fbs_contraction}
Let $f:\mathbb{R}^n\!\to\!\mathbb{R}$ be $f(\x) = 1/2 \|\y - \x\|^2$ for fixed $\y \in \R^n$, and let $g:\mathbb{R}^n\!\to\!\mathbb{R}\cup\{\infty\}$
be proper, closed, and convex. For $\alpha>0$, define
\[
T_\alpha(x)\;=\;\prox_{\alpha g} \big(x-\alpha\nabla f(x)\big).
\]
Then $T_\alpha$ is $(1-\alpha)$-contractive for any $\alpha\in(0,1)$.
\end{lemma}

We next analyze the Lipschitz constant of the convolutional layer. Let $\V_{\K} : \mathbb{R}^{C_{\mathrm{in}}\times H\times W} \to \mathbb{R}^{C_{\mathrm{out}}\times H\times W}$ denote the linear operator of a multi-channel 2D convolution with kernel $\K$ 
of dimension ${C_{\mathrm{out}}\times C_{\mathrm{in}}\times k_h\times k_w}$. Hence, the Lipschitz constant of the convolutional layer equals the operator (spectral) norm of $\V_{\K}$, denoted as $\|\V_{\K}\|_{\mathrm{op}}$.  Under circular boundary conditions on an $H\times W$ grid, $\V_{\K}$ is block-diagonalized by the 2D Discrete Fourier Transform. 
Thus, by ~\cite{Sedghi2019Singular}, a convolutional layer's Lipschitz constant admits a closed form:
\begin{equation}
\label{eq:conv_lip_closed_form}
\|\V_{\K}\|_{\mathrm{op}}
\;=\;
\max_{\omega\in\Omega}\,\sigma_{\max} \big(\widehat H(\omega)\big),
\end{equation}
where $\Omega$ is a grid of size $H \times W$ and $\widehat H(\omega)\in\mathbb{C}^{C_{\mathrm{out}}\times C_{\mathrm{in}}}$ stacks the per-channel 2D FFTs of the spatial kernels:
\begin{equation}
\label{eq:freq_response_def}
\big[\widehat H(\omega)\big]_{o,i}
\;=\;
\mathcal{F}_{H,W}\big(\K_{o,i,:,:}\big)[\omega],
\end{equation}
where $o\in\{1,\ldots,C_{\mathrm{out}}\}$ and $i\in\{1,\ldots,C_{\mathrm{in}}\}$ index the output and input channels. Thus, the convolution layer's Lipschitz constant is the maximum of frequency-response matrices $\widehat H(\omega)$ over all frequencies. Consequently, for a $1$-Lipschitz convolution layer, we need to enforce 
\begin{equation}
\label{eq:conv_contractive_condition}
\max_{\omega\in\Omega}\sigma_{\max}\big(\widehat H(\omega)\big)\;\le\;1.
\end{equation}

We also need the following property of Lipschitz functions  \cite{bauschke2017convex} to constrain our entire architecture to be contractive. 
\begin{lemma}[Lipschitz constant of compositions]
\label{lem:composition_product}
Let $F_i:\mathbb{R}^{n_{i-1}}\to\mathbb{R}^{n_i}$, $i=1,\dots,M$, be $L_i$-Lipschitz. Then the composition $F:=F_M\circ F_{M-1}\circ\cdots\circ F_1$ is Lipschitz and, for all $x,y \in \R^{n_1}$,
\[
\|F(x)-F(y)\|\;\le\;\Big(\prod_{i=1}^M L_i\Big)\,\|x-y\|.
\]
\end{lemma}
More importantly, the result in Lemma \ref{lem:composition_product} ensures that finite compositions of contractive operators are contractive. 

\section{Proposed Method}
\label{sec:proposed}

In this section, we present our contractive architecture: we build a single contractive layer from carefully constrained blocks, form the full model by composing such layers, and discuss implementation details.

At a high level, the proposed layer combines a contractive restoration step with a learnable refinement step. The prox-wavelet block acts as a structured denoising operator that suppresses unstable high-frequency perturbations while preserving the low-frequency image content needed for faithful reconstruction. The Lipschitz-controlled convolution then restores expressive power by learning task-adaptive feature refinements without violating the Lipschitz constraint. Since each component is contractive or nonexpansive with explicit scaling, their composition yields a contractive layer that remains both trainable and practically effective for restoration.

\subsection{Gradient and prox-wavelet layer}
\label{subsec:Gradprox}
We consider the denoising model in Eq.~\eqref{den:forward} and solve the inverse problem via FBS in Eq.~\eqref{fbs}, using a convex wavelet-based regularizer. Let $\W$ be an orthonormal wavelet transform and let $\mathcal{H}$ denote the index set of {high-frequency} detail coefficients; its complement $\mathcal{L}$ indexes the low-frequency coefficients. We choose to regularize {only} the
high-frequency coefficients for image denoising:
\begin{equation}
\label{eq:prox_obj}
\min_{\x\in\mathbb{R}^p}\;\; 
\frac{1}{2}\,\|\x-\y\|^2 \;+\; 
\sum_{i\in\mathcal{H}} \lambda_i\,\big|(\W\x)_i\big|
\end{equation}
Let $\Lambda=\{\lambda_i\}_{i\in\mathcal{H}}$ collect all the coefficient thresholds with constraints that $\lambda_i>0$ i.e. thresholds are positive. This regularizer is motivated from~\cite{nair2024averaged}, where, contrary to their work, we threshold only the high-frequency components. In contrast to BayesShrink~\cite{Chang2000Adaptive} and SUREShrink~\cite{Donoho1995SureShrink}, which set subband-wise thresholds by analytic rules, our approach learns thresholds for the high-frequency subbands.

\begin{figure*}[t]
\begin{center}
\includegraphics[width=0.95\linewidth]{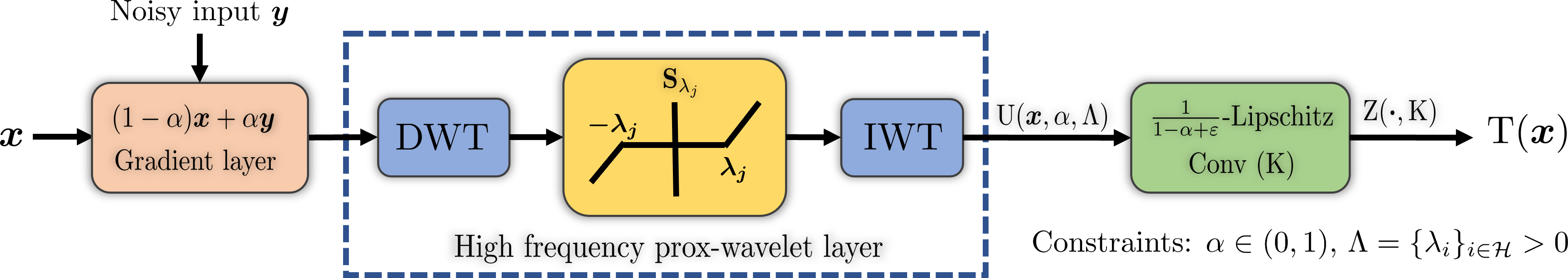}
\end{center}
\caption{Architecture of contractive layer $\mathcal{T}$. The layer performs a gradient step and prox-wavelet operation, followed by a scaled convolution.}
\label{fig:net}
\end{figure*}

A single FBS update for solving Eq.~\eqref{eq:prox_obj} can be written as
\begin{align*}
\text{(Gradient step)}\quad
\z_{k+1} &= (1-\alpha)\,\x_k + \alpha\,\y, \\
\text{(Prox-wavelet step)}\quad
\x_{k+1} &= \W^\top \mathcal{S}_{\Lambda}(\W \z_{k+1}),
\end{align*}
where $\alpha\in(0,1)$ is the step size, and $\mathcal{S}_{\Lambda}:\mathbb{R}^p\!\to\!\mathbb{R}^p$ soft-thresholds only the high-frequency coefficients: 
\begin{equation*}
\big(\mathcal{S}_{\Lambda}(\z)\big)_i =
\begin{cases}
\operatorname{sign}(z_i)\,\max\{|z_i|-\lambda_i,\,0\}, & i \in \mathcal{H},\\[3pt]
z_i, & i \in \mathcal{L}.
\end{cases}
\end{equation*}
We define our trainable layer $\U$ as this single FBS update:
\begin{equation}
\label{eq:prox_layer}
\U(\x;\alpha,\Lambda)
= \W^\top \mathcal{S}_{\Lambda}\!\left(\W\big((1-\alpha)\,\x + \alpha\,\y\big)\right),
\end{equation}
and learn $\alpha$ and the thresholds $\Lambda$ under the explicit constraints $\alpha \in (0,1)$ and $\lambda_i > 0$ for all $i \in \mathcal{H}$. As shown next, this parametrization yields a provably contractive layer.
\begin{proposition}
\label{prop:ctr}
For all $\alpha \in (0,1]$ and $\tri >0$, the map $\x \mapsto \U(\x; \alpha, \tri)$ is $(1-\alpha)$-contractive.
\end{proposition} 
This follows because the  wavelet-based regularizer is closed, proper, and convex \cite{Mallat2009WaveletTour}, so Lemma~\ref{thm:fbs_contraction} applies.

\subsection{Scaled convolution layer}
\label{subsec:Scalconv}
From Eq.\eqref{eq:conv_contractive_condition}, a convolution is contractive iff its operator norm is less than $1$. We keep the kernel $\K$ as the learnable parameter and induce the Lipschitz scaling into the filter operation. Let $s(\K):=\|\V_{\K}\|_{\mathrm{op}}$ denote the operator norm of the convolution as defined in Eq.~\eqref{eq:conv_lip_closed_form}. Define the normalized operator $\overline{\V}_{\K}\;:=\;\V_{\K} / s(\K)$. By construction, $\|\overline{\V}_{\K}\|_{\mathrm{op}} \le 1$, and hence $\overline{\V}_{\K}$ is nonexpansive i.e. Lipschitz constant $\le 1$.  Thus $1$-Lipschitz convolutional layer is defined as 
\begin{equation}
\label{eq:gamma_outside}
\Z(\x;\K)\;=\;\overline{\V}_{\K}\x
\;=\;\,\frac{\V_{\K}\x}{s(\K)}
\end{equation}
Replacing $s(\K)$ by $s(\K)+\varepsilon$ with $\varepsilon>0$, constrains $\Z(\x;\K)$ to be contractive. Given a target Lipschitz budget $\gamma>0$, we can use $\gamma \Z(\x;\K)$ as 
$\gamma$-Lipschitz convolution layer. 

\label{subsec:Contr}
The proposed layer architecture is shown in Fig~\ref{fig:net}. To form a contractive layer, we combine a gradient step and a prox-wavelet block (one FBS update), with a Lipschitz-controlled convolution to refine the features. We define the layer $\T(\x; \alpha,\Lambda,\K)$ as, 
\begin{equation}
\label{eq:layer_def}
\tfrac{1}{(1-\alpha)+\varepsilon}\,\Z\big(\U(\x;\alpha,\Lambda);\K\big)
\end{equation}
where $\K$ is the convolution kernel, and $\varepsilon>0$ enforces the convolution has a Lipschitz constant less than $1 / (1-\alpha)$. Here $\U(\cdot;\alpha,\Lambda)$ is the FBS-based block defined in Eq.~\eqref{eq:prox_layer}, and $\Z(\cdot;\K)$ is the $1$-Lipschitz convolutional layer defined in Eq.~\eqref{eq:gamma_outside}.

The FBS-based block \(\U(\cdot;\alpha,\Lambda)\) is \((1-\alpha)\)-contractive by Proposition \ref{prop:ctr}. We therefore compose FBS-based block with a \(\Z(\cdot;\K)\), $1$-Lipschitz convolution with kernel \(\K\), and scale it by \(1/((1-\alpha)+\varepsilon)\). This preserves  contractivity of the layer while relaxing the constraint to scale the main branch convolutional layer to be \(1\)-Lipschitz. This construction is motivated by the contractive FBS blocks in~\cite[Sec.~4.2, Props.~7--8]{nair2024averaged}. We formalize the layer contractivity result next.
\begin{proposition}
\label{prop:ctrnet}
For all $\alpha \in (0,1)$, $\epsilon>0$, and $\tri>0$, the map
$\x \mapsto \T(\x; \alpha,\tri,\K)$ is contractive.
\end{proposition}
We define the depth-$M$ network as the composition
\(
F = \T_M \circ \T_{M-1} \circ \cdots \circ \T_1.
\)
For each layer $\ell = 1,\dots,M$, let
\[
\T_\ell(\x)
=
\T\big(\x;\alpha_\ell,\Lambda_\ell,\K_{\ell}\big),
\]
where $\alpha_\ell \in (0,1)$, $\Lambda_\ell = \{\lambda_{\ell,i}\}_{i\in\mathcal{H}}$ with $\lambda_{\ell,i}>0$, and $\K_{\ell}$ are the layer-specific convolution kernels. By Proposition~\ref{prop:ctrnet}, each $\T_\ell$ is contractive and therefore, by Lemma~\ref{lem:composition_product}, the overall network $F$ is contractive.

\noindent \textbf{Implementation details:} We distinguish between constraints required for the contractivity guarantee and empirical design choices used for performance. To satisfy the layerwise conditions in Proposition~\ref{prop:ctrnet}, for each layer $\ell$ we constrain the stepsize $\alpha_\ell$ to lie in $(0,1)$ by clipping after each gradient step, enforce positivity of the high-frequency wavelet thresholds $\lambda_{\ell,i}$, $i\in\mathcal{H}$, using the softplus function, and control the Lipschitz constants of the convolutions $\K_\ell$ via the exact normalization described in Sec.~\ref{subsec:Scalconv}. The PyTorch reference implementation for convolutional normalization is provided below.

\begin{lstlisting}[language=Python,
basicstyle=\ttfamily\footnotesize,
columns=fullflexible,
keepspaces=true,
breaklines=true,
breakatwhitespace=false,
xleftmargin=0.5em,
aboveskip=0.5em,
belowskip=0.5em]
def clip_norm(kernel, input_shape, Lip):
    # kernel: [C_out, C_in, k_h, k_w]
    H, W = input_shape
    C_out, C_in = kernel.shape[:2]
    eps = 1e-12

    # FFT-domain convolution operator
    K_fft = torch.fft.fft2(kernel, s=(H, W))
    Huv = K_fft.permute(2, 3, 0, 1).reshape(H * W, C_out, C_in)
    s_max = torch.linalg.norm(Huv, ord=2, dim=(-2, -1)).max()

    # Rescale if the target Lipschitz bound is exceeded
    if s_max > Lip:
        kernel = kernel * (Lip / (s_max + eps))
    return kernel
\end{lstlisting}

The remaining choices are made for empirical effectiveness. We fix the network depth to $M=30$ and cycle the orthogonal wavelet transforms across layers as Haar, Daubechies-4 (dB4), and Symlet-4 (sym4). The model is trained on fixed-size patches using Adam with an initial learning rate of $1\times10^{-4}$ and a three-stage decay at epochs 10 and 20, for 50 epochs with batch size 256 on $64\times64$ BSD500 patches extracted with stride 4. Data augmentation consists of random horizontal/vertical flips and rotations by $90^\circ$, $180^\circ$, and $270^\circ$. At test time, to handle arbitrary image resolutions, we use patchwise inference similar to restoration transformers~\cite{Liang2021SwinIR,chen2021pre}. The image is processed using overlapping $P\times P$ patches ($P=64$) with stride $s$ such that $P \bmod s = 0$, and the outputs are aggregated by overlap-add using a Tukey window~\cite{pielawski2020introducing}, which smoothly blends neighboring patches and avoids blocking artifacts.

\section{Experiments and Analysis}
\label{Experiments}
We evaluate our denoiser in both standalone image denoising and as a regularizer within the  iterative scheme PnP-FBS in Eq.~\eqref{eq:pnpfbs}, for deblurring and superresolution. Across all settings, we compare against provably $1$-Lipschitz/contractive baselines and strong unconstrained networks. In addition to visual comparisons, we report quantitative results (PSNR and SSIM) on multiple standard benchmarks. All experiments were run on an NVIDIA A40 GPU. Additional experimental results are provided in Appendix~\ref{sec:appendix}.

\noindent \textbf{Evaluation methods:} We evaluate the proposed contractive denoiser from two perspectives: (i) as a standalone denoiser, and (ii) as a prior within Plug-and-Play (PnP) restoration, where contractivity is relevant for convergence guarantees. In both settings, the most important comparisons are against prior provably \(1\)-Lipschitz / contractive denoisers. To this end, we include the provably \(1\)-Lipschitz deep denoisers from~\cite{nair2024averaged}, \textbf{D-FBS} and \textbf{D-DRS}. We also construct two \(1\)-Lipschitz variants of DnCNN~\cite{DnCNN17}, namely \textbf{eSN-ReLU} and \textbf{eSN-GS}, by rescaling all convolutional layers using exact spectral norms, as in Sec.~\ref{subsec:Scalconv}. For these models, we remove batch normalization and use \(1\)-Lipschitz activations, namely ReLU and GroupSort~\cite{Anil2019GroupSort}. To assess the remaining performance gap to unconstrained restoration models, we further compare with strong unconstrained baselines: \textbf{DnCNN}~\cite{DnCNN17}, \textbf{Restormer}~\cite{Restormer22}, \textbf{IDF}~\cite{kim2025idf}, and \textbf{Noise2VST}~\cite{herbreteau2025self}. In our restoration experiments, we use PnP-FBS as the default base algorithm and plug in different denoisers to isolate the effect of the prior. For completeness, Table~\ref{tab:pnp_algorithms_denoisers} also reports results obtained by plugging our denoiser into several PnP algorithms, highlighting its versatility across reconstruction schemes.

\begin{figure}[t]
\begin{center}
\setlength{\tabcolsep}{3pt}
\renewcommand{\arraystretch}{1.0}
\scriptsize

\begin{tabular}{cccccc}

\includegraphics[width=0.15\linewidth]{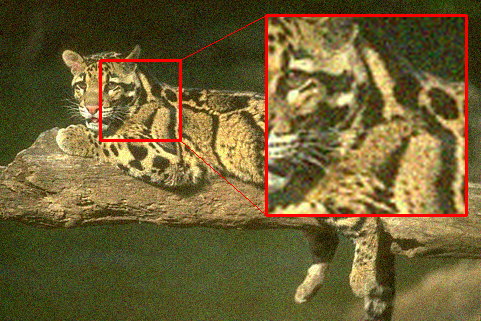} &
\includegraphics[width=0.15\linewidth]{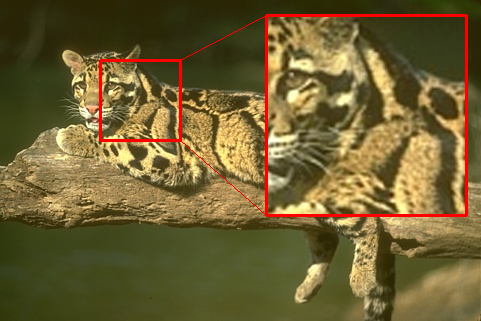} &
\includegraphics[width=0.15\linewidth]{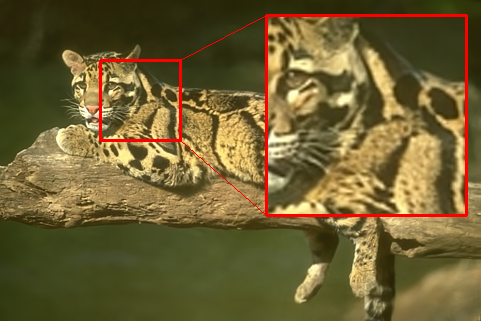} &
\includegraphics[width=0.15\linewidth]{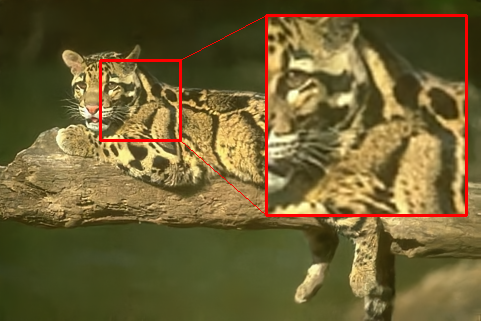} &
\includegraphics[width=0.15\linewidth]{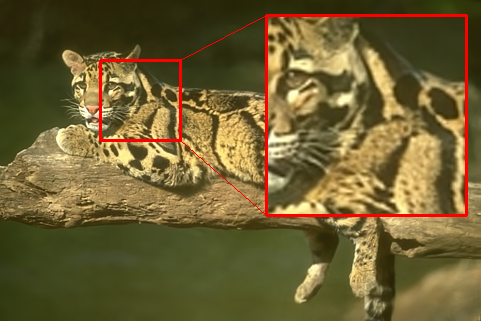} &
\includegraphics[width=0.15\linewidth]{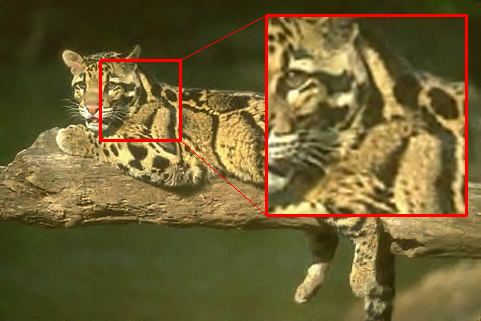} \\

Noisy &
Reference &
NOISE2VST &
DnCNN &
Restormer &
Ours \\

$(24.77/0.501)$ &
&
$(34.77/0.958)$ &
$(34.73/0.964)$ &
$(34.86/0.965)$ &
$(33.90/0.951)$

\end{tabular}
\end{center}
\caption{
Color Gaussian denoising ($\sigma=15$). Our contractive denoiser preserves fine structures similar to unconstrained models, while closely matching Restormer.
}
\label{fig:gauss_color_visual}
\end{figure}

\begin{figure}[t]
\begin{center}
\setlength{\tabcolsep}{3pt}
\renewcommand{\arraystretch}{1.0}
\scriptsize

\begin{tabular}{cccccc}

\includegraphics[width=0.15\linewidth]{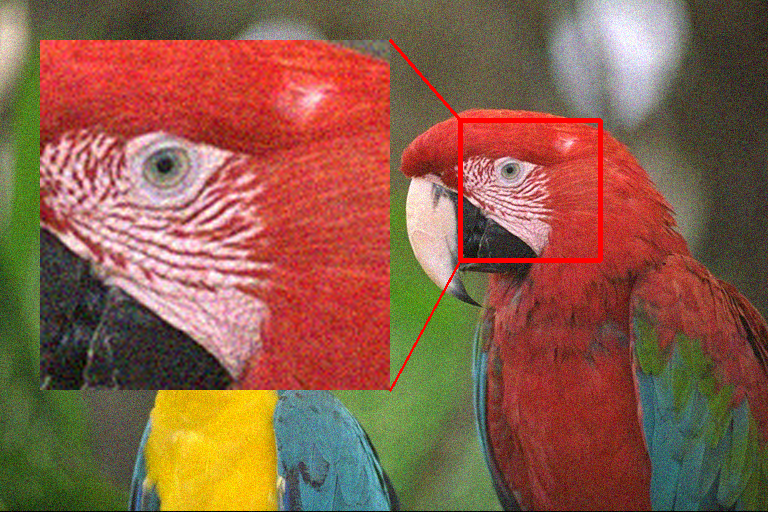} &
\includegraphics[width=0.15\linewidth]{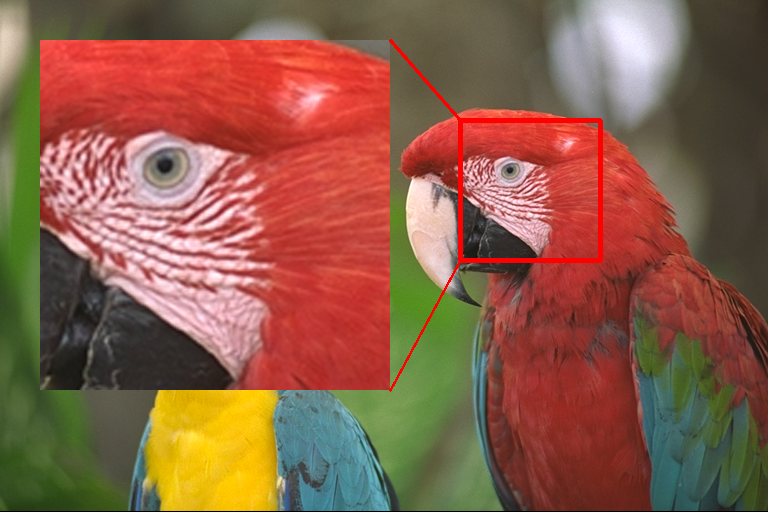} &
\includegraphics[width=0.15\linewidth]{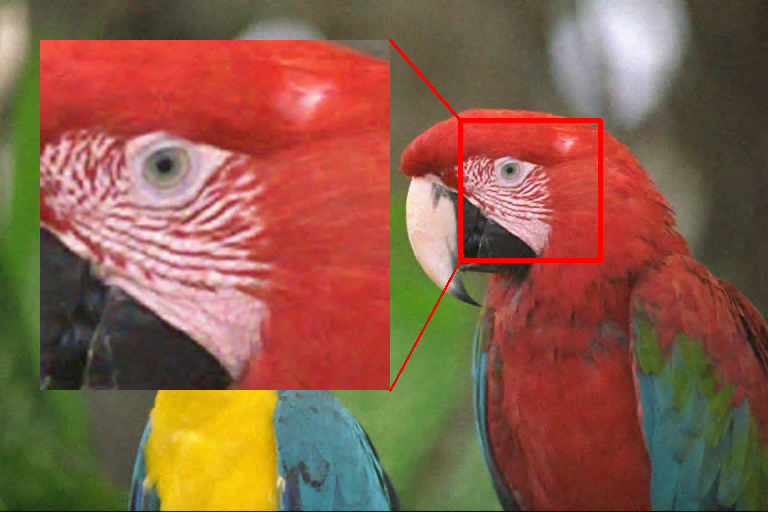} &
\includegraphics[width=0.15\linewidth]{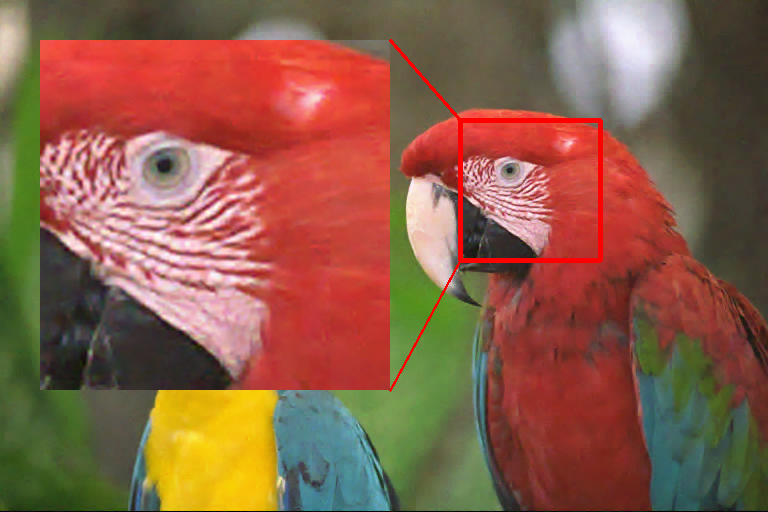} &
\includegraphics[width=0.15\linewidth]{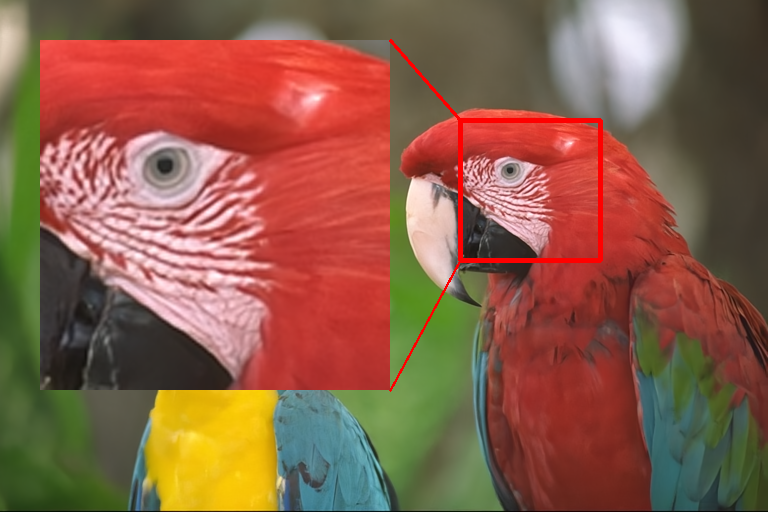} &
\includegraphics[width=0.15\linewidth]{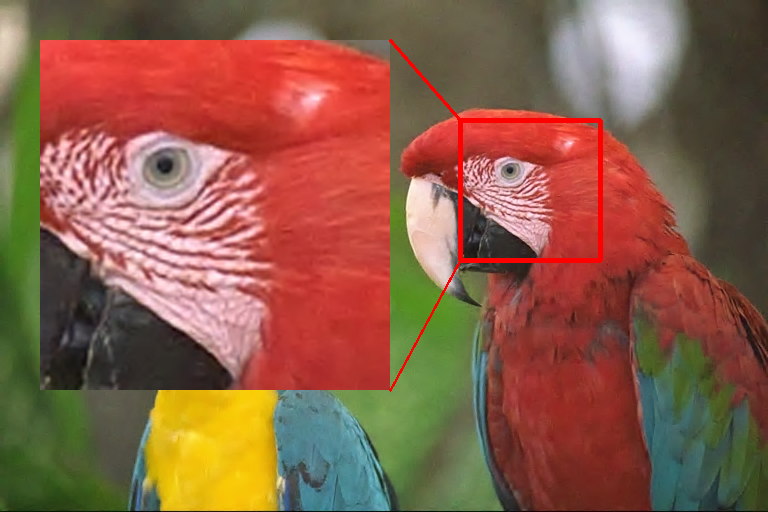} \\

Noisy &
Reference &
D-FBS &
IDF &
Restormer &
Ours \\

$(24.70/0.3795)$ &
&
$(33.41/0.8532)$ &
$(34.67/0.8917)$ &
$(36.96/0.9355)$ &
$(35.88/0.9191)$

\end{tabular}
\end{center}
\caption{
Color Gaussian denoising ($\sigma=15$). Our contractive model preserves feather and eye-ring textures similar to the best-performing Restormer.
}
\label{fig:color_gaussian2}
\end{figure}

\begin{table}[!t]

\setlength{\tabcolsep}{2.5pt}
\small
\begin{center}
\resizebox{\textwidth}{!}{
\begin{tabular}{lccccccccc}
\toprule
\multirow{2}{*}{Method} &
\multicolumn{3}{c}{CBSD68} &
\multicolumn{3}{c}{Kodak24} &
\multicolumn{3}{c}{Urban100} \\
\cmidrule(lr){2-4}
\cmidrule(lr){5-7}
\cmidrule(lr){8-10}
& $\sigma{=}15$ & $\sigma{=}25$ & $\sigma{=}50$
& $\sigma{=}15$ & $\sigma{=}25$ & $\sigma{=}50$
& $\sigma{=}15$ & $\sigma{=}25$ & $\sigma{=}50$ \\
\midrule

eSN-ReLU
  & {\scriptsize $24.84/0.5919$} & {\scriptsize $22.54/0.5174$} & {\scriptsize $19.76/0.2189$}
  & {\scriptsize $24.76/0.5265$} & {\scriptsize $20.44/0.350$}  & {\scriptsize $17.87/0.2713$}
  & {\scriptsize $24.95/0.6291$} & {\scriptsize $20.68/0.5802$} & {\scriptsize $17.17/0.2922$} \\

eSN-GS
  & {\scriptsize $24.98/0.6016$} & {\scriptsize $22.78/0.5765$} & {\scriptsize $20.43/0.2245$}
  & {\scriptsize $24.87/0.5364$} & {\scriptsize $20.86/0.350$}  & {\scriptsize $17.95/0.2856$}
  & {\scriptsize $25.21/0.6267$} & {\scriptsize $20.85/0.5832$} & {\scriptsize $17.32/0.3214$} \\

D-FBS
  & {\scriptsize $30.12/0.8409$} & {\scriptsize $27.54/0.7577$} & {\scriptsize $23.87/0.6206$}
  & {\scriptsize $31.20/0.8315$} & {\scriptsize $28.44/0.7558$} & {\scriptsize $24.90/0.6380$}
  & {\scriptsize $29.53/0.8519$} & {\scriptsize $26.34/0.7729$} & {\scriptsize $22.02/0.6273$} \\

D-DRS
  & {\scriptsize $30.07/0.8526$} & {\scriptsize $27.51/0.7695$} & {\scriptsize $24.11/0.6266$}
  & {\scriptsize $31.05/0.8465$} & {\scriptsize $28.58/0.7710$} & {\scriptsize $25.15/0.6484$}
  & {\scriptsize $29.73/0.8687$} & {\scriptsize $26.70/0.7934$} & {\scriptsize $22.52/0.6438$} \\

Ours
  & {\scriptsize $\mathbf{32.74/0.9115}$} & {\scriptsize $\mathbf{29.78/0.8501}$} & {\scriptsize $\mathbf{25.61/0.7211}$}
  & {\scriptsize $\mathbf{33.42/0.9020}$} & {\scriptsize $\mathbf{30.67/0.8446}$} & {\scriptsize $\mathbf{26.49/0.7285}$}
  & {\scriptsize $\mathbf{31.96/0.9092}$} & {\scriptsize $\mathbf{28.94/0.8585}$} & {\scriptsize $\mathbf{24.40/0.7441}$} \\

\midrule
\midrule

DnCNN
  & {\scriptsize $33.63/0.9296$} & {\scriptsize $30.72/0.8806$} & {\scriptsize $26.52/0.7714$}
  & {\scriptsize $34.37/0.9219$} & {\scriptsize $31.70/0.8837$} & {\scriptsize $27.63/0.7785$}
  & {\scriptsize $32.65/0.9199$} & {\scriptsize $30.11/0.8837$} & {\scriptsize $25.74/0.7990$} \\

IDF
  & {\scriptsize $32.18/0.8915$} & {\scriptsize $29.80/0.8435$} & {\scriptsize $25.73/0.7188$}
  & {\scriptsize $32.87/0.8839$} & {\scriptsize $30.64/0.8410$} & {\scriptsize $26.60/0.7224$}
  & {\scriptsize $31.42/0.8929$} & {\scriptsize $29.28/0.8571$} & {\scriptsize $25.06/0.7547$} \\

Noise2VST
  & {\scriptsize $33.71/0.9288$} & {\scriptsize $\mathbf{31.03/0.8877}$} & {\scriptsize $25.93/0.7450$}
  & {\scriptsize $\mathbf{34.64/0.9221}$} & {\scriptsize $\mathbf{32.33/0.8898}$} & {\scriptsize $27.44/0.7450$}
  & {\scriptsize $\mathbf{33.58/0.9253}$} & {\scriptsize $\mathbf{31.43/0.9032}$} & {\scriptsize $\mathbf{26.48/0.8234}$} \\

Restormer
  & {\scriptsize $\mathbf{33.85/0.9316}$} & {\scriptsize $30.90/0.8837$} & {\scriptsize $\mathbf{26.65/0.7765}$}
  & {\scriptsize $34.62/0.9237$} & {\scriptsize $31.93/0.8814$} & {\scriptsize $\mathbf{27.82/0.7901}$}
  & {\scriptsize $33.16/0.9229$} & {\scriptsize $30.55/0.8900$} & {\scriptsize $26.07/0.8129$} \\
\bottomrule
\end{tabular}}
\end{center}
\caption{\textbf{Color denoising (PSNR(dB)/SSIM)}. Our model dominates $1$-Lipschitz models (in top) and narrows the gap with unconstrained models (in bottom).}
\label{tab:color_denoising}
\end{table}

\subsection{Image denoising}
\label{sec:denoising}

We evaluate our contractive denoiser on synthetic noise datasets. For synthetic denoising, we use Set12~\cite{DnCNN17}, BSD68~\cite{martin2001database}, Urban100~\cite{huang2015single}, and Kodak24~\cite{Kodak24}, generating noisy inputs by adding AWGN at varying noise levels.

\noindent \textbf{Gaussian denoising:}
For color Gaussian denoising, we evaluate on CBSD68, Kodak24, and Urban100 at $\sigma \in \{15,25,50\}$. For a fair comparison, every compared model is trained on BSD500, with hyperparameters tuned to its best validation performance. Table~\ref{tab:color_denoising} summarizes the results. Across all tested settings, our model is the strongest provably \(1\)-Lipschitz / contractive denoiser in terms of PSNR/SSIM. In contrast, the \(1\)-Lipschitz variants of DnCNN (eSN-ReLU and eSN-GS) fail to learn an effective denoiser, exhibiting severe degradation in PSNR/SSIM. While a gap to the best unconstrained model, Restormer, remains, it is relatively modest (typically around $0.8$--$1.5$ dB in PSNR and $0.02$--$0.06$ in SSIM), despite the proposed model being strictly contractive. The qualitative examples in Fig.~\ref{fig:gauss_color_visual} and Fig.~\ref{fig:color_gaussian2} further show that our reconstructions preserve natural textures and clean flat regions in a manner closer to strong unconstrained models, whereas other provable baselines tend to oversmooth. We next evaluate grayscale denoising with AWGN at $\sigma \in \{15,25,50\}$ on BSD68, Set12, and Urban100. Table~\ref{tab:gray_denoising} reports average PSNR/SSIM for all baselines. The same trend holds in the grayscale setting: among provably \(1\)-Lipschitz / contractive methods, our model consistently achieves the best performance. Moreover, the remaining gap to unconstrained networks is substantially smaller than that of prior provable baselines, amounting to about $0.7$--$1.3$ dB on Set12 and BSD68 and about $2$ dB on Urban100.

\begin{table}[!t]
\setlength{\tabcolsep}{2.5pt}
\small
\begin{center}
\resizebox{\textwidth}{!}{
\begin{tabular}{lccccccccc}
\toprule
\multirow{2}{*}{Method} &
\multicolumn{3}{c}{BSD68} &
\multicolumn{3}{c}{Set12} &
\multicolumn{3}{c}{Urban100} \\
\cmidrule(lr){2-4}
\cmidrule(lr){5-7}
\cmidrule(lr){8-10}
& $\sigma{=}15$ & $\sigma{=}25$ & $\sigma{=}50$
& $\sigma{=}15$ & $\sigma{=}25$ & $\sigma{=}50$
& $\sigma{=}15$ & $\sigma{=}25$ & $\sigma{=}50$ \\
\midrule
eSN-ReLU
  & {\scriptsize $24.89/0.6899$} & {\scriptsize $20.48/0.4159$} & {\scriptsize $14.91/0.2179$}
  & {\scriptsize $25.67/0.6618$} & {\scriptsize $21.34/0.3941$} & {\scriptsize $15.75/0.2625$}
  & {\scriptsize $24.89/0.6262$} & {\scriptsize $20.61/0.4774$} & {\scriptsize $14.78/0.2705$} \\
eSN-GS
  & {\scriptsize $24.93/0.6976$} & {\scriptsize $20.68/0.4313$} & {\scriptsize $15.21/0.2231$}
  & {\scriptsize $25.73/0.6743$} & {\scriptsize $21.43/0.3854$} & {\scriptsize $15.87/0.2634$}
  & {\scriptsize $24.91/0.6341$} & {\scriptsize $20.78/0.5236$} & {\scriptsize $14.86/0.2823$} \\
D-FBS
  & {\scriptsize $29.89/0.8347$} & {\scriptsize $27.31/0.7537$} & {\scriptsize $24.21/0.6290$}
  & {\scriptsize $30.41/0.8451$} & {\scriptsize $28.23/0.7924$} & {\scriptsize $24.53/0.6788$}
  & {\scriptsize $29.21/0.8478$} & {\scriptsize $26.48/0.7803$} & {\scriptsize $22.43/0.6461$} \\
D-DRS
  & {\scriptsize $29.79/0.8560$} & {\scriptsize $27.24/0.7600$} & {\scriptsize $24.18/0.6368$}
  & {\scriptsize $30.39/0.8535$} & {\scriptsize $28.15/0.8136$} & {\scriptsize $24.48/0.6815$}
  & {\scriptsize $29.18/0.8774$} & {\scriptsize $26.47/0.8058$} & {\scriptsize $22.39/0.6339$} \\

 Ours
  & {\scriptsize $\mathbf{30.52/0.8690}$} & {\scriptsize $\mathbf{27.82/0.7883}$} & {\scriptsize $\mathbf{24.40/0.6460}$}
  & {\scriptsize $\mathbf{31.43/0.8816}$} & {\scriptsize $\mathbf{28.71/0.8219}$} & {\scriptsize $\mathbf{25.19/0.7160}$}
  & {\scriptsize $\mathbf{30.11/0.8851}$} & {\scriptsize $\mathbf{26.94/0.8118}$} & {\scriptsize $\mathbf{22.93/0.6690}$} \\
\midrule
\midrule
DnCNN
  & {\scriptsize $\mathbf{31.58/0.8934}$} & {\scriptsize $\mathbf{28.87/0.8254}$} & {\scriptsize $\mathbf{25.12/0.6846}$}
  & {\scriptsize $\mathbf{32.82/0.9044}$} & {\scriptsize $\mathbf{30.32/0.8609}$} & {\scriptsize $\mathbf{26.54/0.7654}$}
  & {\scriptsize $\mathbf{32.36/0.9159}$} & {\scriptsize $\mathbf{29.36/0.8677}$} & {\scriptsize $\mathbf{24.75/0.7481}$} \\
Restormer
  & {\scriptsize $31.47/0.8930$} & {\scriptsize $28.75/0.8233$} & {\scriptsize $24.35/0.6288$}
  & {\scriptsize $32.69/0.9020$} & {\scriptsize $30.13/0.8574$} & {\scriptsize $26.15/0.6950$}
  & {\scriptsize $32.20/0.9139$} & {\scriptsize $29.13/0.8630$} & {\scriptsize $24.43/0.6798$} \\
\bottomrule
\end{tabular}}
\end{center}
\caption{\textbf{Grayscale denoising (PSNR(dB)/SSIM)}. Our model is best among $1$-Lipschitz models (top) and competitive with unconstrained models (bottom).}
\label{tab:gray_denoising}
\end{table}

\begin{table}[t]
\centering
\scriptsize
\renewcommand{\arraystretch}{1.05}

\setlength{\tabcolsep}{12pt}
\begin{tabular}{lccccc}
\toprule
Dataset & PGD & DRS-Diff & GS & FBS & DRS \\
\midrule
CBSD68
& $26.13/0.7062$ & $26.03/0.663$
& $26.23/0.713$  & $26.18/0.707$
& $24.26/0.623$ \\

Kodak24
& $27.01/0.7116$ & $26.12/0.603$
& $27.11/0.720$  & $27.05/0.712$
& $25.19/0.649$ \\

Urban100
& $23.41/0.7167$ & $25.17/0.722$
& $23.51/0.722$  & $23.47/0.718$
& $21.20/0.602$ \\
\bottomrule
\end{tabular}

\vspace{2pt}

{\scriptsize\textbf{(a) Different PnP algorithms with our denoiser}}

\vspace{4pt}

\setlength{\tabcolsep}{10pt}
\begin{tabular}{lccccc}
\toprule
PnP method
& DRUNet
& CRR-NN (prox.)
& CRR-NN ($t$-step)
& WCRR
& Ours \\
\midrule

GS-PnP
& $26.12/0.701$
& $25.11/0.663$
& $25.13/0.667$
& $25.21/0.678$
& $\mathbf{26.23/0.713}$ \\

DRS-PnP
& $24.17/0.610$
& $23.91/0.587$
& $23.94/0.589$
& $24.01/0.598$
& $\mathbf{24.26/0.623}$ \\

\bottomrule
\end{tabular}

\vspace{2pt}

{\scriptsize\textbf{(b) Different denoisers on CBSD68}}

\caption{\textbf{PnP deblurring with a $90\%$ random sparse kernel
(PSNR(dB)/SSIM).} (a) Comparison of PnP algorithms using our denoiser.
(b) Comparison of different denoisers within GS-PnP and DRS-PnP,
including convergent baselines.}
\label{tab:pnp_algorithms_denoisers}
\end{table}

\subsection{Iterative restoration with denoisers}
\label{sec:pnp}

We next demonstrate the practical payoff of contractivity in iterative restoration, focusing on deblurring and super-resolution. In this setting, contractivity is useful not only because it yields convergence guarantees within Plug-and-Play (PnP) algorithms, but also because it can lead to strong restoration quality when paired with an expressive denoiser. The forward model is
\(
\y = \S\B\x + \eta,
\)
where $\S$ is a fixed downsampling operator, $\B$ is a known blur, and $\eta$ is AWGN for super-resolution. For deblurring, $\S$ is the identity. We use PnP-FBS in Eq.~\eqref{eq:pnpfbs} to invert this model; for contractive denoisers, PnP-FBS is guaranteed to converge~\cite{nair2024averaged}. 

\begin{figure}[!t]
\begin{center}
\setlength{\tabcolsep}{4pt}

\begin{tabular}{cc}
\includegraphics[width=0.40\linewidth]{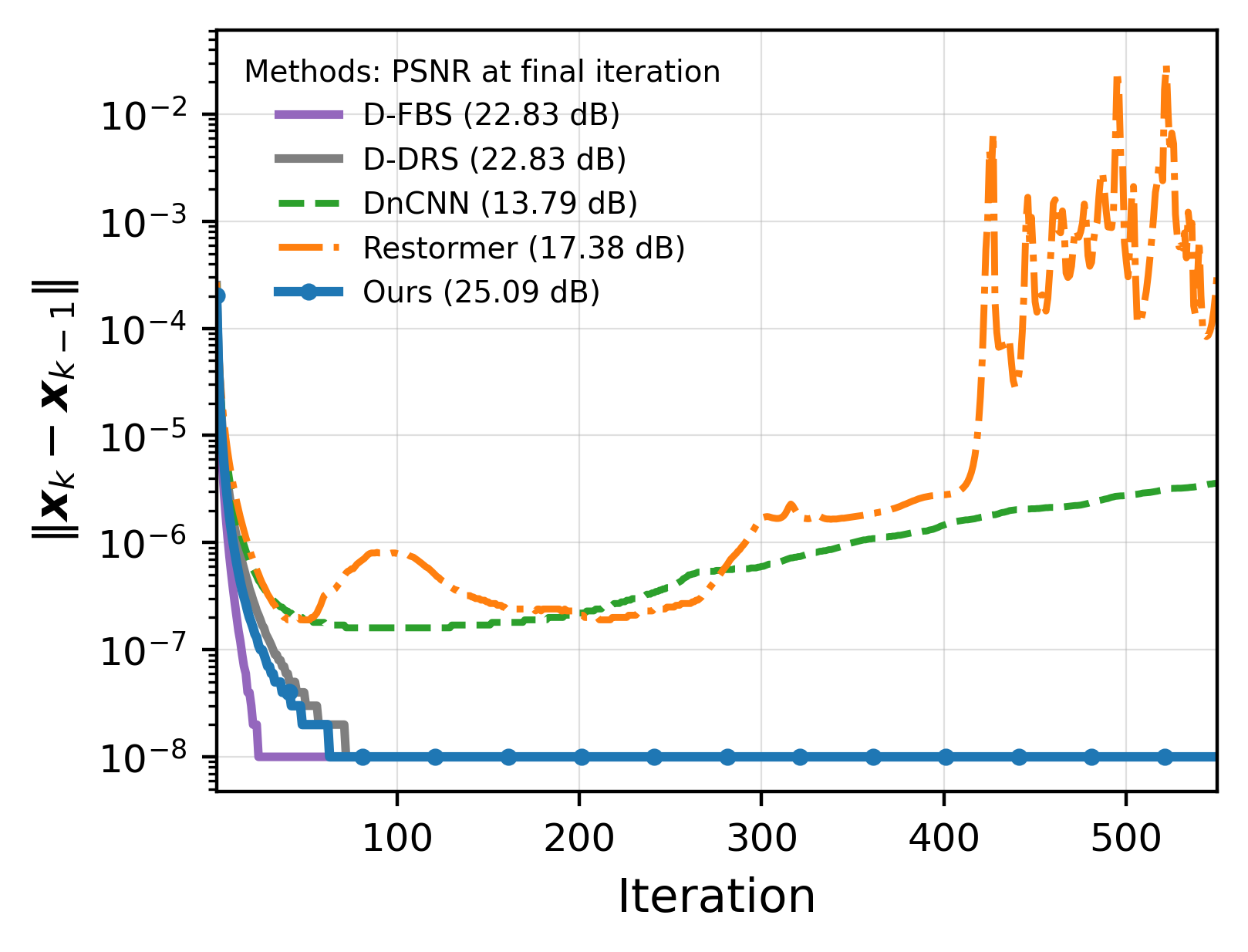} &
\includegraphics[width=0.40\linewidth]{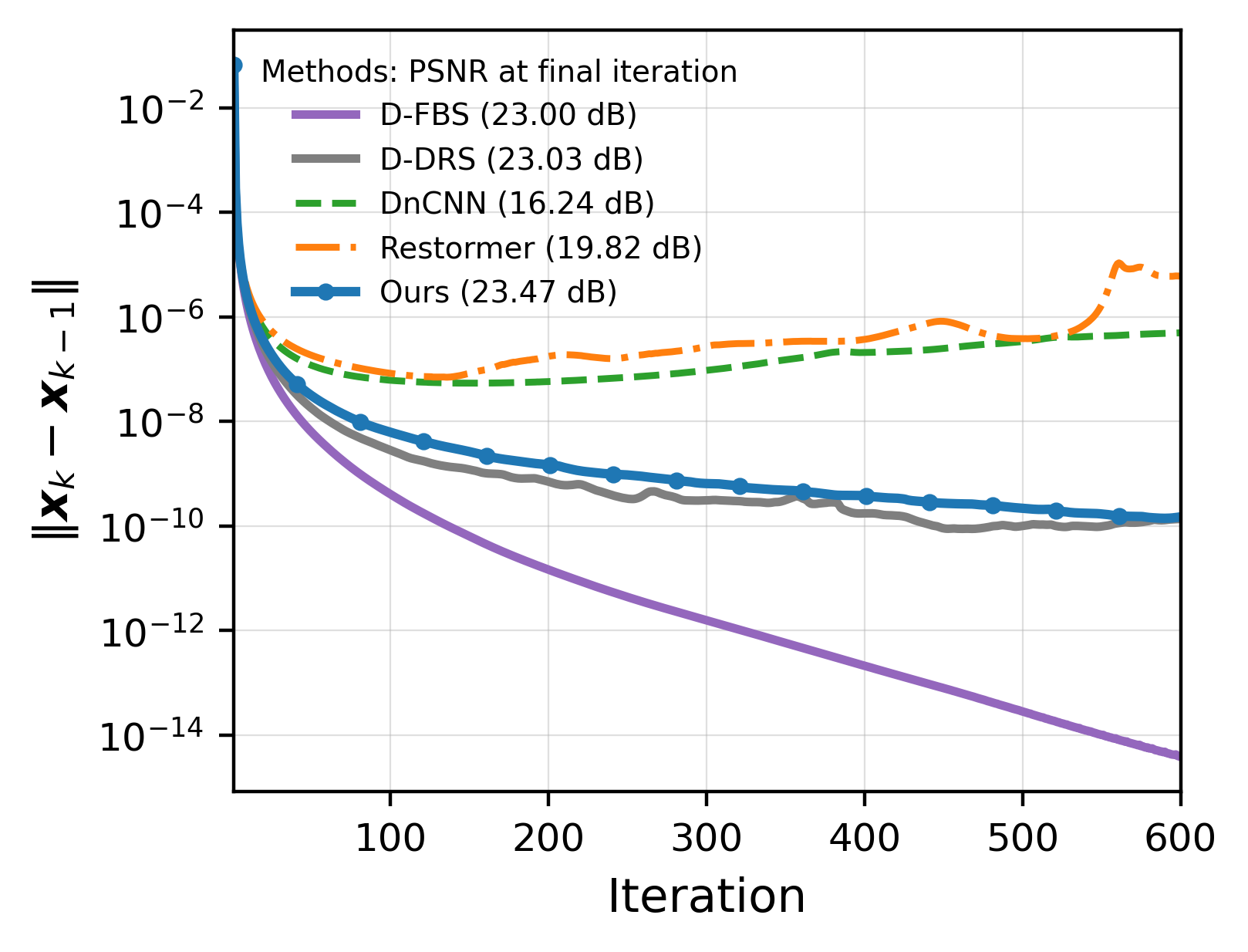}
\end{tabular}
\end{center}
\caption{PnP convergence. All $1$-Lipschitz denoisers converge; unconstrained models diverge, while ours achieves the best PSNR.}
\label{fig:pnp_conv}
\end{figure}

We first study convergence in Fig.~\ref{fig:pnp_conv} by running PnP-FBS with different denoisers on fixed images for super-resolution and deblurring. Across settings, all \(1\)-Lipschitz / contractive models exhibit monotone convergence, whereas unconstrained models can diverge. Importantly, our method combines this convergence behavior with the strongest asymptotic restoration performance among the provable baselines. We next present visual comparisons and quantitative results on deblurring and super-resolution. We omit the \(1\)-Lipschitz DnCNN variants in this section because, as shown earlier, they do not provide competitive denoising quality. In all tables, the best score is highlighted in \textbf{bold}. For visual comparison, we include DnCNN, Restormer, and the stronger of D-FBS and D-DRS.

\begin{figure}[!t]
\begin{center}
\setlength{\tabcolsep}{3pt}
\renewcommand{\arraystretch}{1.0}
\scriptsize

\begin{tabular}{cccccc}

\includegraphics[width=0.15\linewidth]{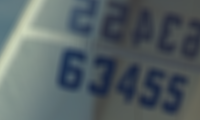} &
\includegraphics[width=0.15\linewidth]{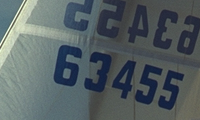} &
\includegraphics[width=0.15\linewidth]{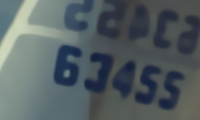} &
\includegraphics[width=0.15\linewidth]{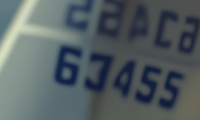} &
\includegraphics[width=0.15\linewidth]{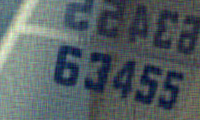} &
\includegraphics[width=0.15\linewidth]{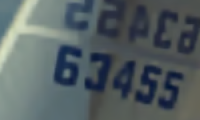} \\

Blurred &
Reference &
IDF &
Noise2VST &
Restormer &
Ours \\

$(22.23/0.6061)$ &
&
$(26.62/0.7244)$ &
$(26.89/0.7309)$ &
$(25.40/0.4712)$ &
$(27.99/0.7707)$

\end{tabular}
\end{center}
\caption{
PnP box deblurring with a $9\times9$ kernel. IDF and Noise2VST oversmooth fine details, and Restormer introduces artifacts. Our method restores sharper structures.
}
\label{fig:deblur_motion_example}
\end{figure}

\begin{figure}[!t]
\begin{center}
\setlength{\tabcolsep}{3pt}
\renewcommand{\arraystretch}{1.0}
\scriptsize

\begin{tabular}{cccccc}

\includegraphics[width=0.15\linewidth]{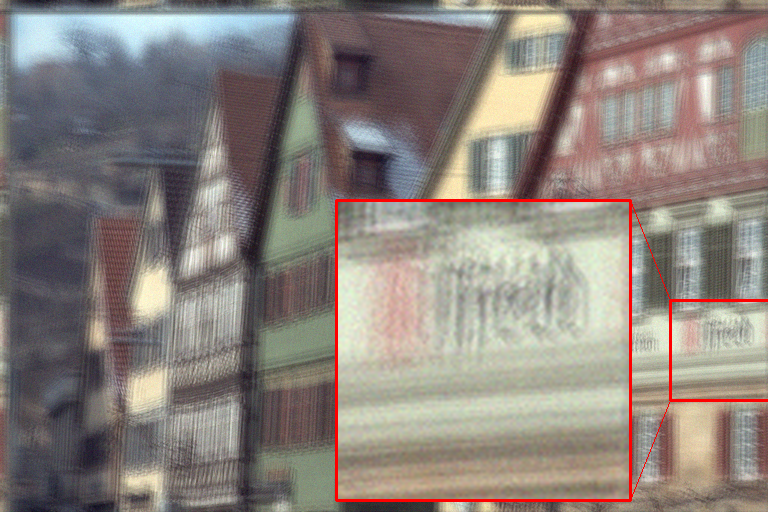} &
\includegraphics[width=0.15\linewidth]{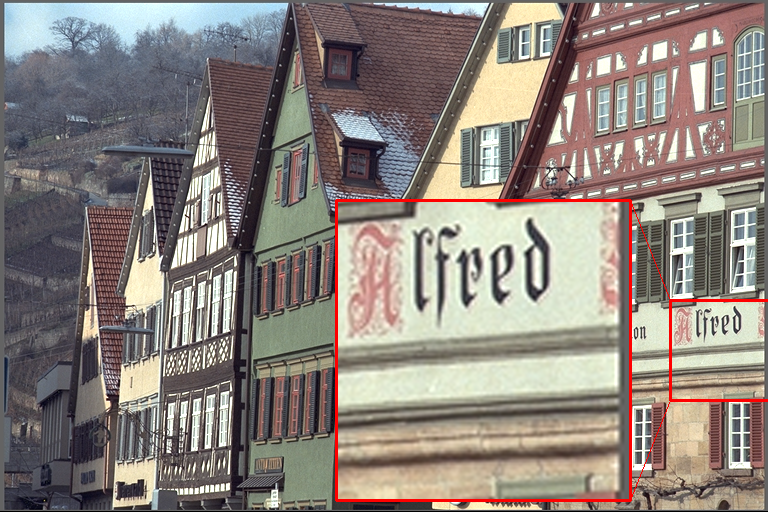} &
\includegraphics[width=0.15\linewidth]{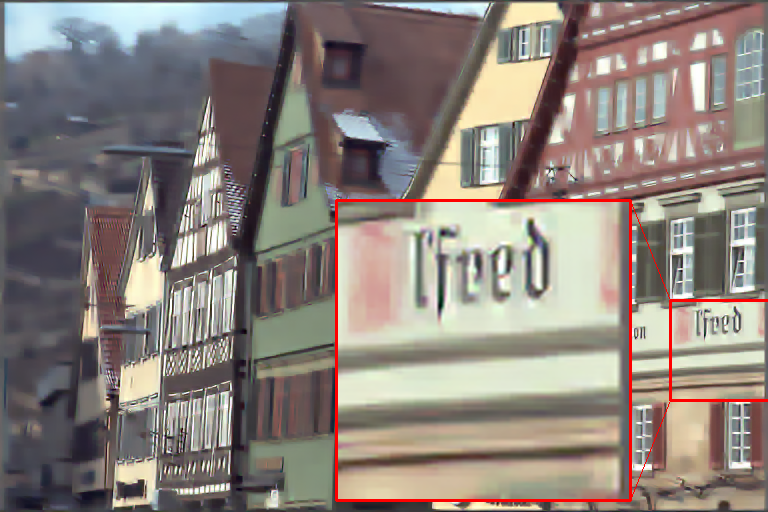} &
\includegraphics[width=0.15\linewidth]{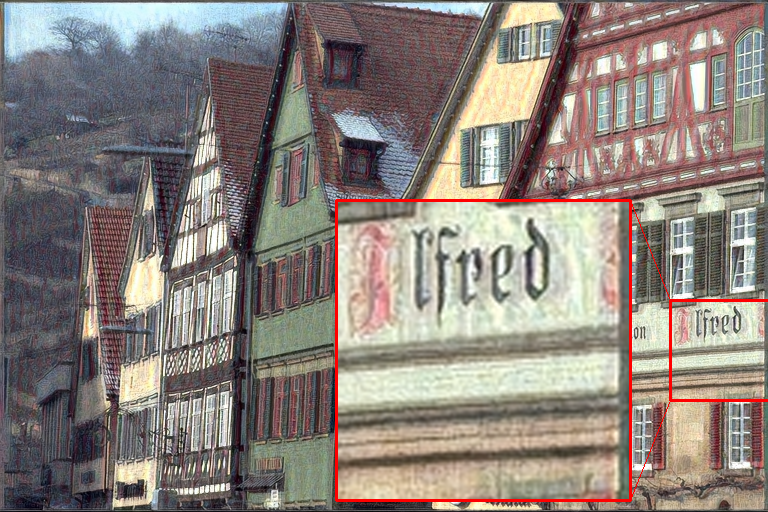} &
\includegraphics[width=0.15\linewidth]{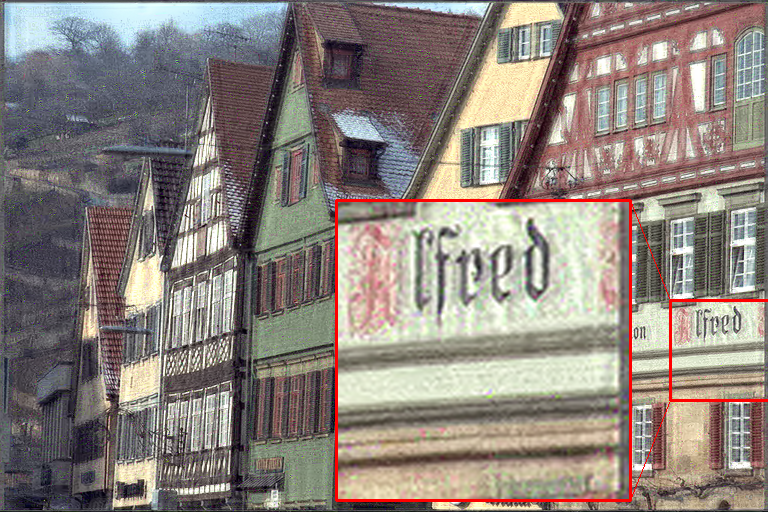} &
\includegraphics[width=0.15\linewidth]{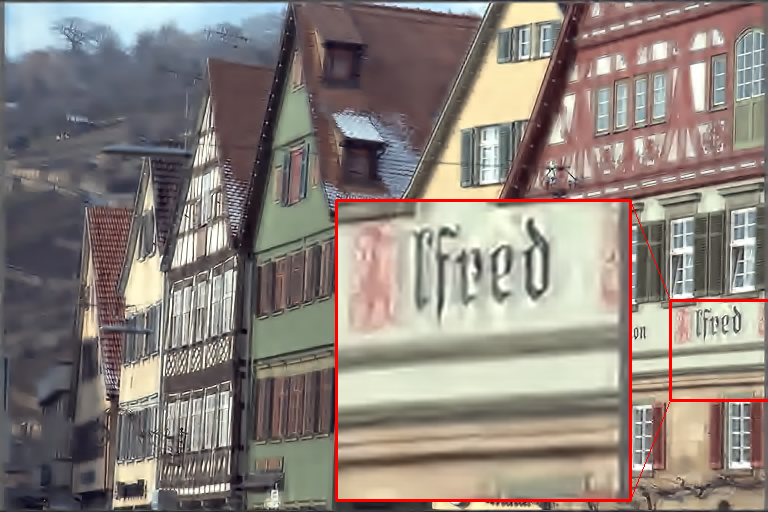} \\

Input &
Reference &
D-DRS &
DnCNN &
Restormer &
Ours \\

$(13.73 / 0.150)$ &
&
$(22.49 / 0.616)$ &
$(23.61 / 0.727)$ &
$(23.38 / 0.714)$ &
$(24.00 / 0.757)$

\end{tabular}
\end{center}
\caption{
PnP deblurring with $90\%$ sparse kernel. Our model preserves edges and letters with fewer artifacts than DnCNN and Restormer, without D-DRS-like oversmoothing.
}
\label{fig:deblur_sparse}
\end{figure}

\begin{figure}[t]
\begin{center}
\setlength{\tabcolsep}{3pt}
\renewcommand{\arraystretch}{1.0}
\scriptsize

\begin{tabular}{cccccc}

\includegraphics[width=0.15\linewidth]{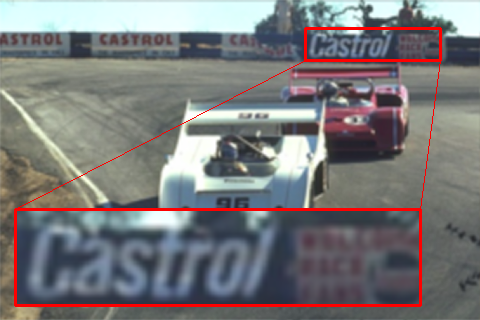} &
\includegraphics[width=0.15\linewidth]{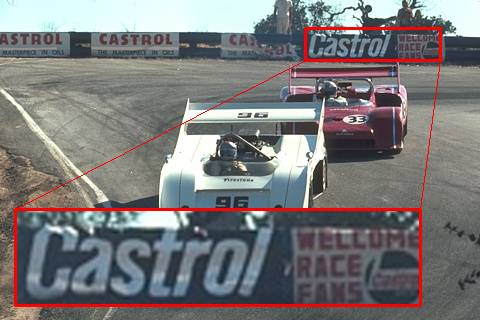} &
\includegraphics[width=0.15\linewidth]{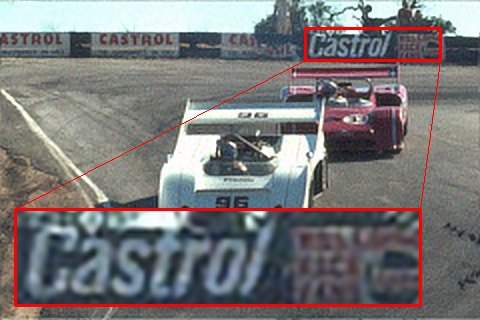} &
\includegraphics[width=0.15\linewidth]{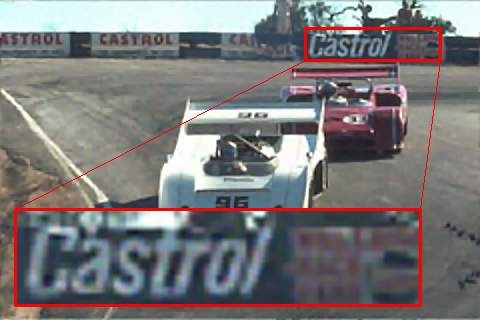} &
\includegraphics[width=0.15\linewidth]{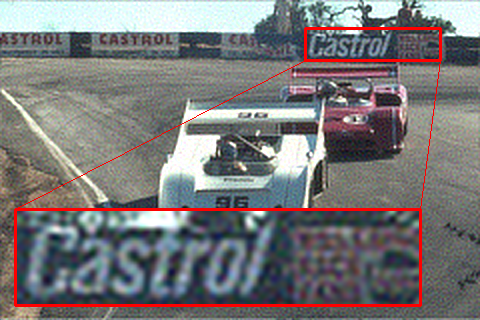} &
\includegraphics[width=0.15\linewidth]{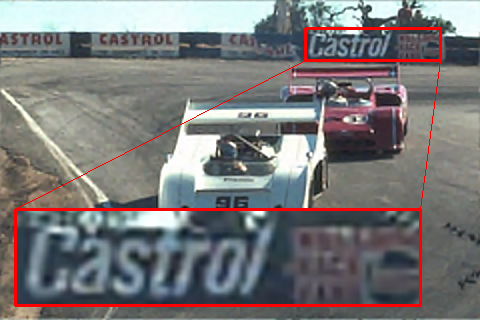} \\

Bicubic &
Reference &
DnCNN &
D-DRS &
Restormer &
Ours \\

$(23.79 / 0.696)$ &
&
$(26.78 / 0.759)$ &
$(26.40 / 0.754)$ &
$(25.23 / 0.732)$ &
$(27.44 / 0.786)$

\end{tabular}
\end{center}
\caption{
$2\times$ PnP-FBS superresolution ($\mathbf{B}$: $9\times9$ Gaussian blur, $\sigma=1$). Our model restores sharper text and edges with fewer artifacts as compared to DnCNN and less oversmoothing.
}
\label{fig:sr_color_visual}
\end{figure}

\begin{figure}[!t]
\begin{center}
\setlength{\tabcolsep}{3pt}
\renewcommand{\arraystretch}{1.0}
\scriptsize

\begin{tabular}{cccccc}

\includegraphics[width=0.15\linewidth]{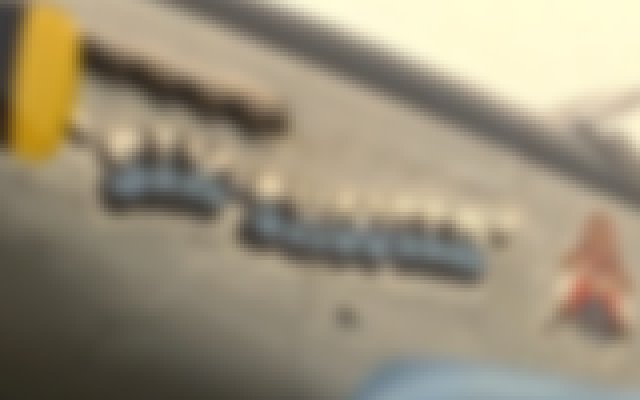} &
\includegraphics[width=0.15\linewidth]{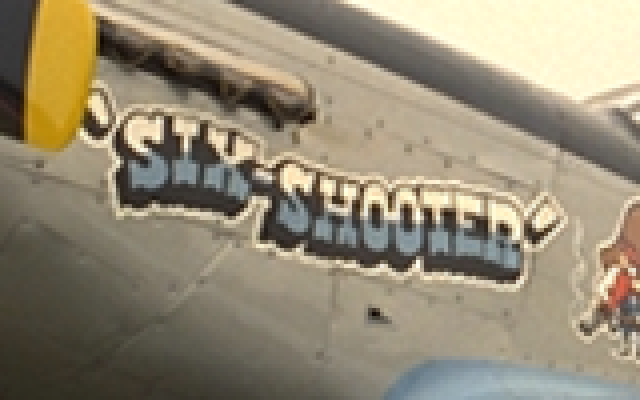} &
\includegraphics[width=0.15\linewidth]{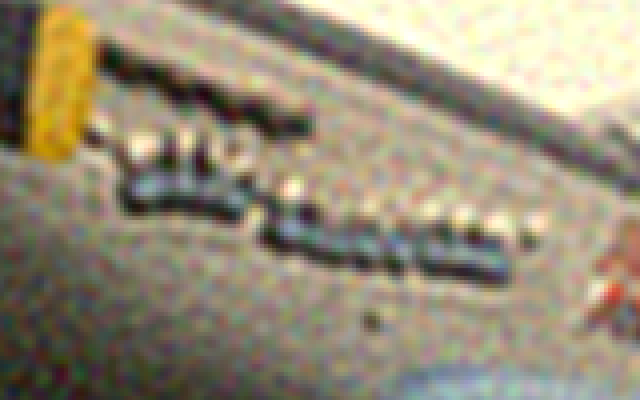} &
\includegraphics[width=0.15\linewidth]{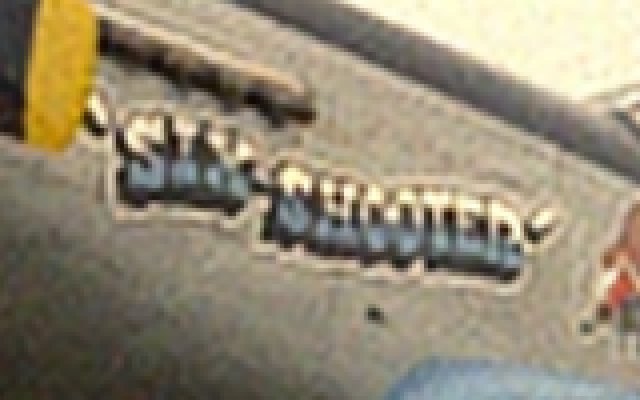} &
\includegraphics[width=0.15\linewidth]{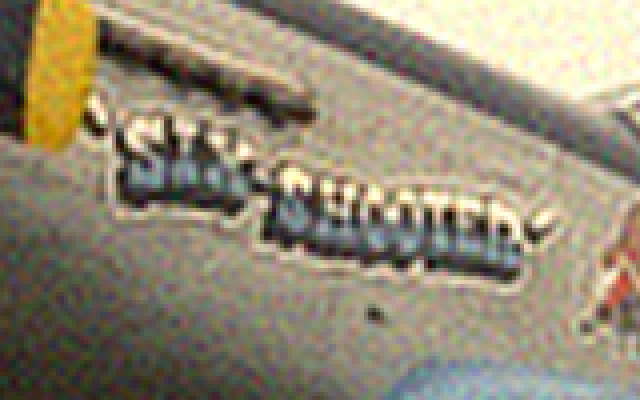} &
\includegraphics[width=0.15\linewidth]{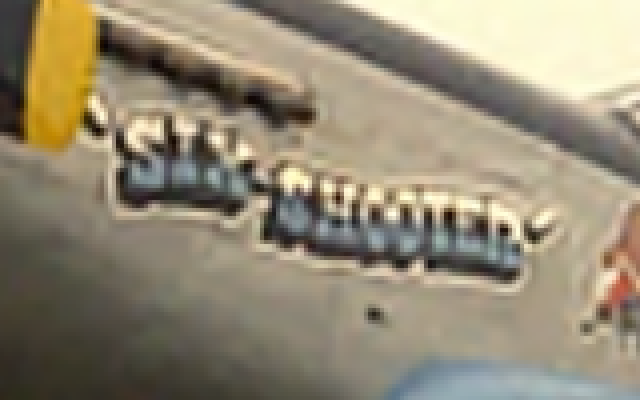} \\

Bicubic &
Reference &
Noise2VST &
DnCNN &
Restormer &
Ours \\

$(22.12/0.8652)$ &
&
$(27.14/0.6267)$ &
$(29.15/0.8072)$ &
$(27.21/0.6478)$ &
$(29.40/0.8692)$

\end{tabular}
\end{center}
\caption{
$2\times$ PnP-FBS superresolution ($\mathbf{B}$: Gaussian blur $25\times25$, $\sigma=1.6$). Our denoiser yields sharper details and fewer artifacts than unconstrained models.
}
\label{fig:superres1}
\end{figure}
\begin{table}[!t]
\scriptsize
\setlength{\tabcolsep}{4pt}
\renewcommand{\arraystretch}{0.65}

\begin{center}
\resizebox{1\linewidth}{!}{
\begin{tabular}{lcc|ccccccc}
\toprule

\multirow{2}{*}{Dataset}
& \multicolumn{2}{c|}{Diffusion-based PnP}
& \multicolumn{7}{c}{PnP-FBS} \\
\cmidrule(lr){2-3}
\cmidrule(lr){4-10}

& DiffPIR
& SGPnP
& FBS
& DRS
& IDF
& Noise2VST
& DnCNN
& Restormer
& \textbf{OURS} \\

\midrule

CBSD68
& $27.86/0.725$
& $28.12/0.8001$
& $26.62/0.7403$
& $27.03/0.7438$
& $24.26/0.6009$
& $24.72/0.6130$
& $27.11/0.7179$
& $25.35/0.7013$
& $\mathbf{28.40}/\mathbf{0.8063}$ \\

Kodak24
& $28.69/0.6967$
& $29.08/0.8021$
& $27.45/0.7456$
& $27.91/0.7451$
& $25.24/0.6430$
& $25.82/0.6490$
& $27.45/0.6772$
& $25.24/0.6692$
& $\mathbf{29.21}/\mathbf{0.8102}$ \\

Urban100
& $\mathbf{28.31}/0.8012$
& $28.02/0.8059$
& $24.59/0.7600$
& $25.48/0.7741$
& $22.56/0.6390$
& $23.33/0.7369$
& $25.86/0.7419$
& $25.26/0.7397$
& $26.33/\mathbf{0.8099}$ \\

\bottomrule
\end{tabular}
}
\end{center}

\caption{PSNR/SSIM for deblurring with a $90\%$ sparse random blur. Diffusion-based methods, which do not provide contractivity guarantees, are reported separately.}
\label{tab:deblur_sparse_pnp}
\end{table}

\begin{table}[!t]
\scriptsize
\setlength{\tabcolsep}{4pt}
\renewcommand{\arraystretch}{0.65}

\begin{center}
\resizebox{1\linewidth}{!}{
\begin{tabular}{lcc|ccccccc}
\toprule

\multirow{2}{*}{Dataset}
& \multicolumn{2}{c|}{Diffusion-based PnP}
& \multicolumn{7}{c}{PnP-FBS} \\
\cmidrule(lr){2-3}
\cmidrule(lr){4-10}

& DiffPIR
& SGPnP
& FBS
& DRS
& IDF
& Noise2VST
& DnCNN
& Restormer
& \textbf{OURS} \\

\midrule

CBSD68
& $27.44/0.7315$
& $27.18/0.7773$
& $26.14/0.7276$
& $26.44/0.7300$
& $24.47/0.6251$
& $25.05/0.6479$
& $26.41/0.6800$
& $26.33/0.7161$
& $\mathbf{27.50}/\mathbf{0.7797}$ \\

Kodak24
& $\mathbf{28.82}/0.7114$
& $28.12/0.7761$
& $27.03/0.7412$
& $27.38/0.7410$
& $25.30/0.6573$
& $26.11/0.6779$
& $26.83/0.6344$
& $27.74/0.7339$
& $28.53/\mathbf{0.7850}$ \\

Urban100
& $\mathbf{27.03}/0.8146$
& $26.87/\mathbf{0.8472}$
& $23.87/0.7414$
& $24.39/0.7557$
& $22.09/0.6263$
& $23.66/0.7126$
& $25.02/0.6908$
& $25.50/0.7705$
& $25.63/0.8009$ \\

\bottomrule
\end{tabular}
}
\end{center}

\caption{PSNR/SSIM for deblurring with Levin kernel
1~\cite{levin2009understanding}. Diffusion-based methods, which do not provide contractivity guarantees, are reported separately.}
\label{tab:deblur_levin_red}
\end{table}

\noindent \textbf{Deblurring:} 
As shown in Table~\ref{tab:pnp_algorithms_denoisers}, our contractive denoiser achieves comparable performance across several PnP base algorithms for deblurring under a \(90\%\) sparse random blur. We therefore use PnP-FBS as the default backbone throughout, ensuring fair comparisons by isolating the effect of the denoiser.
In Tables~\ref{tab:deblur_sparse_pnp} and \ref{tab:deblur_levin_red}, we report PnP-FBS deblurring on standard color benchmarks under two forward operators \(\mathbf{B}\): (i) a \(90\%\) sparse random blur and (ii) the Levin ``kernel~1'' blur \cite{levin2009understanding}. Among the denoiser-based PnP-FBS methods, our contractive denoiser achieves the best performance across datasets under both blur settings, outperforming strong unconstrained denoisers. We additionally compare against the generative PnP baselines DiffPIR and SGPnP at fixed iteration counts, demonstrating competitive performance; unlike our denoiser, however, they do not enforce contractivity for convergent PnP restoration. These results suggest that enforcing stability (contractivity) while retaining high denoising quality translates into improved PnP reconstructions.
Visual results in Figs.~\ref{fig:deblur_motion_example} and \ref{fig:deblur_sparse} highlight the qualitative advantage of our contractive denoiser under severe blur: it recovers sharper edges and fine text while avoiding the oversmoothing of IDF/Noise2VST and the artifacts often introduced by unconstrained baselines (see zoomed regions).

\noindent \textbf{Superresolution:}
We evaluate color \(2\times\) super-resolution using PnP-FBS, where the forward operator \(\mathbf{B}\) is a known blur and \(\mathbf{S}\) denotes RGB downsampling. Under Gaussian blur (Table~\ref{tab:sr_color_pnp}), our provably contractive denoiser achieves the best PSNR/SSIM across all benchmarks, outperforming both \(1\)-Lipschitz baselines and strong unconstrained denoisers (DnCNN, Restormer). The same trend holds under anisotropic blur (Table~\ref{tab:sr_color_aniso}), indicating that the performance gains are robust to the choice of blur kernel $\B$. Qualitative results in Figs.~\ref{fig:sr_color_visual} and \ref{fig:superres1} provide additional examples under Gaussian blur settings:
our PnP reconstructions recover sharper text and edges while retaining fine textures. Compared with \(1\)-Lipschitz baselines, we avoid oversmoothing, and compared with unconstrained denoisers (e.g., DnCNN/Restormer) we reduce visually salient artifacts (see zoomed regions).

\begin{table}[!t]
\scriptsize
\setlength{\tabcolsep}{5pt}
\renewcommand{\arraystretch}{0.65}

\begin{center}
\resizebox{1\linewidth}{!}{
\begin{tabular}{lccccccc}
\toprule
Dataset
& FBS
& DRS
& IDF
& Noise2VST
& DnCNN
& Restormer
& \textbf{OURS} \\
\midrule

CBSD68
& $25.28/0.6835$
& $25.36/0.6863$
& $24.54/0.6267$
& $24.94/0.6498$
& $25.36/0.6670$
& $24.51/0.5822$
& $\mathbf{25.61/0.7046}$ \\

Kodak24
& $26.04/0.6994$
& $26.12/0.7018$
& $25.32/0.6575$
& $25.73/0.6705$
& $25.96/0.6580$
& $25.05/0.5588$
& $\mathbf{26.35/0.7184}$ \\

Urban100
& $22.56/0.6750$
& $22.71/0.6830$
& $21.79/0.6203$
& $22.62/0.6700$
& $22.83/0.6541$
& $22.34/0.5822$
& $\mathbf{22.89/0.6976}$ \\

\bottomrule
\end{tabular}
}
\end{center}

\caption{PnP-FBS superresolution with Gaussian blur ($9\times9$, $\sigma=2$) and noise level \(5\). Results are
reported as PSNR(dB)/SSIM.}
\label{tab:sr_color_pnp}
\end{table}

\begin{table}[!t]
\scriptsize
\setlength{\tabcolsep}{4pt}
\renewcommand{\arraystretch}{0.65}

\begin{center}
\resizebox{1\linewidth}{!}{
\begin{tabular}{lccccccc}
\toprule
Dataset
& FBS
& DRS
& IDF
& Noise2VST
& DnCNN
& Restormer
& \textbf{OURS} \\
\midrule

CBSD68
& $25.17/0.6773$
& $25.25/0.6800$
& $24.42/0.6203$
& $24.82/0.6427$
& $25.30/0.6633$
& $24.38/0.5731$
& $\mathbf{25.51/0.6986}$ \\

Kodak24
& $25.95/0.6945$
& $26.02/0.6965$
& $25.20/0.6518$
& $25.63/0.6652$
& $25.92/0.6555$
& $24.93/0.5498$
& $\mathbf{26.26/0.7135}$ \\

Urban100
& $22.42/0.6666$
& $22.57/0.6743$
& $21.64/0.6114$
& $22.48/0.6614$
& $22.74/0.6484$
& $22.21/0.5719$
& $\mathbf{22.76/0.6896}$ \\

\bottomrule
\end{tabular}
}
\end{center}

\caption{PnP-FBS superresolution with a $21\times21$ anisotropic blur
($\sigma_x=3$, $\sigma_y=1$, $\theta=0$). Results are reported as PSNR(dB)/SSIM.}
\label{tab:sr_color_aniso}
\end{table}

\subsection{Ablation study}

\begin{figure*}[!t]
\scriptsize
\renewcommand{\arraystretch}{0.72}

\begin{center}


\begin{minipage}[t]{0.29\linewidth}
\vspace{0pt}
\centering

\textbf{(a) Intermediate representations}\\[2pt]

\setlength{\tabcolsep}{1.5pt}
\renewcommand{\arraystretch}{0.85}

\begin{tabular}{ccc}
\includegraphics[width=0.30\linewidth]
{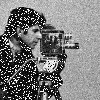}
&
\includegraphics[width=0.30\linewidth]
{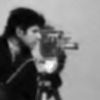}
&
\includegraphics[width=0.30\linewidth]
{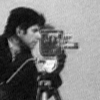}
\\[-1pt]

Noisy & Layer $2$ & Layer $8$ \\[2pt]

\includegraphics[width=0.30\linewidth]
{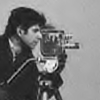}
&
\includegraphics[width=0.30\linewidth]
{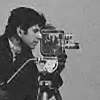}
&
\includegraphics[width=0.30\linewidth]
{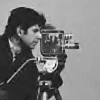}
\\[-1pt]

Layer $15$ & Layer $27$ & Layer $30$
\end{tabular}
\end{minipage}
\hfill
\begin{minipage}[t]{0.30\linewidth}
\vspace{0pt}
\centering

\textbf{(b) Component ablations}\\[2pt]

\setlength{\tabcolsep}{4pt}
\renewcommand{\arraystretch}{0.72}

\begin{tabular}{lcc}
\toprule
Variant & PSNR & Params \\
\midrule
No prox        & $30.63$ & $1.9$K \\
No conv        & $29.93$ & $363$K \\
LF-Thresh      & $32.87$ & $371$K \\
DnCNN          & $33.63$ & $668$K \\
Restormer      & $33.85$ & $26.1$M \\
Fixed $\alpha$ & $26.61$ & $365$K \\
\textbf{Ours}  & $32.74$ & $365$K \\
\bottomrule
\end{tabular}
\end{minipage}
\hfill
\begin{minipage}[t]{0.35\linewidth}
\vspace{0pt}
\centering

\textbf{(c) Learned step sizes}\\[2pt]

\setlength{\tabcolsep}{4pt}
\renewcommand{\arraystretch}{0.72}

\begin{tabular}{lccc}
\toprule
Layer & $\sigma=15$ & $\sigma=25$ & $\sigma=50$ \\
\midrule
$L_5$    & $0.3485$ & $0.2998$ & $0.0000$ \\
$L_{10}$ & $0.4847$ & $0.4359$ & $0.9406$ \\
$L_{15}$ & $0.0000$ & $0.1328$ & $0.9950$ \\
$L_{20}$ & $0.1698$ & $0.0000$ & $0.0000$ \\
$L_{25}$ & $0.9665$ & $0.9886$ & $0.9868$ \\
$L_{30}$ & $0.0801$ & $0.0607$ & $0.0511$ \\
\bottomrule
\end{tabular}
\end{minipage}

\vspace{5pt}


\begin{minipage}[t]{0.20\linewidth}
\vspace{0pt}
\centering

\textbf{(d) Layer count}\\[2pt]

\setlength{\tabcolsep}{4pt}
\renewcommand{\arraystretch}{0.72}

\begin{tabular}{ccc}
\toprule
$M$ & PSNR & SSIM \\
\midrule
$15$          & $32.41$          & $0.9032$ \\
$\mathbf{30}$ & $\mathbf{32.74}$ & $\mathbf{0.9115}$ \\
$45$          & $32.71$          & $0.9013$ \\
$60$          & $32.73$          & $0.9102$ \\
\bottomrule
\end{tabular}
\end{minipage}
\hfill
\begin{minipage}[t]{0.54\linewidth}
\vspace{0pt}
\centering

\textbf{(e) Complexity and runtime}\\[2pt]

\setlength{\tabcolsep}{4pt}
\renewcommand{\arraystretch}{0.72}

\resizebox{0.94\linewidth}{!}{
\begin{tabular}{lcccc}
\toprule
Model & Params & FLOPs & Memory & Time (ms) \\
\midrule
DnCNN
& $668$K
& $102.45$G
& $244.63$MB
& $\mathbf{5.40}$ \\

Restormer
& $26.1$M
& $330.44$G
& $1506.61$MB
& $502.13$ \\

DRUNet
& $32.6$M
& $336.58$G
& $431.76$MB
& $74.00$ \\

\textbf{Ours} ($s=32$)
& $\mathbf{365}$K
& $\mathbf{2.03}$G
& $\mathbf{70.77}$MB
& $175.75$ \\
\bottomrule
\end{tabular}
}
\end{minipage}
\hfill
\begin{minipage}[t]{0.20\linewidth}
\vspace{0pt}
\centering

\textbf{(f) Patch stride}\\[2pt]

\setlength{\tabcolsep}{5pt}
\renewcommand{\arraystretch}{0.72}

\begin{tabular}{cc}
\toprule
Stride $s$ & Time (ms) \\
\midrule
$32$ & $\mathbf{175.75}$ \\
$16$ & $419.31$ \\
$8$  & $1515.91$ \\
$4$  & $5886.29$ \\
\bottomrule
\end{tabular}
\end{minipage}

\end{center}

\caption{\textbf{Ablation and complexity analysis on CBSD68.}
(a) Intermediate representations across layers; (b) architectural
components; (c) learned step sizes $\alpha_k$; (d) network depth;
(e) complexity and runtime for a $481\times321$ RGB input; and
(f) runtime across patch strides.}
\label{fig:combined_ablations}
\end{figure*}

\noindent\textbf{Intermediate representations and depth.}
To better understand how the proposed contractive network restores image
structure, Fig.~\ref{fig:combined_ablations}(a) visualizes outputs from early, middle, and final layers for a fixed image. Lower-frequency structure is recovered in the early layers, while finer edges and high-frequency details emerge in deeper layers, suggesting a progressive reconstruction
process that combines stable coarse restoration with learned detail refinement. Table~\ref{fig:combined_ablations}(d) further studies network depth for color Gaussian denoising ($\sigma=15$). Performance improves with depth and saturates around $M=30$, our default setting, indicating that additional contractive layers increase restoration capacity up to a knee depth, beyond which the gains are marginal.

\noindent\textbf{Role of architectural components.}
Table~\ref{fig:combined_ablations}(b) examines the main design choices in the proposed layer: thresholding low-frequency components in the prox-wavelet block (\emph{LF-Thresh}), removing the prox-wavelet block (\emph{no prox}), removing convolutional refinement (\emph{no conv}), and fixing the step size to $\alpha=0.5$. Removing either block substantially degrades PSNR, with the largest drop for \emph{no prox}, confirming that the prox-wavelet block is essential rather than auxiliary. Thresholding low-frequency coefficients yields negligible gains despite increasing the parameter count, supporting our choice to regularize only high-frequency coefficients. Fixing $\alpha$ also degrades performance, consistent with Table.~\ref{fig:combined_ablations}(c), where the learned step sizes vary substantially across layers within the admissible range $(0,1)$.

\noindent\textbf{Complexity and runtime.}
As shown in Table~\ref{fig:combined_ablations}(e), our model has substantially fewer parameters, FLOPs, and memory requirements than the compared architectures. Table~\ref{fig:combined_ablations}(f) shows the
runtime--accuracy trade-off induced by patch overlap: reducing the stride increases overlap and computational cost, while $s=32$ provides the most efficient configuration evaluated. 

Overall, the ablations show that the proposed design balances denoising quality, stability, and computational efficiency while achieving high-quality PnP restoration. 

\section{Conclusion and Future Work}

We constructed a provably contractive denoiser that attains performance close to strong unconstrained baselines such as DnCNN and Restormer. Empirically, the model remains stable under realistic input perturbations while retaining contractivity; when deployed within PnP iterations, it yields convergent and high-quality reconstructions, whereas unconstrained denoisers can lead to divergence. Our results show that a provably contractive denoiser can substantially narrow the performance gap to unconstrained models. Beyond denoising, we aim to develop end-to-end contractive restoration networks that combine provable stability guarantees with strong reconstruction quality. A current limitation is that, in inverse problems, our method is deployed within iterative PnP algorithms, which can be slower than single-pass restoration networks. In addition, exact Lipschitz control of convolutional layers via FFT-based spectral normalization introduces extra computational overhead during training. Future work will focus on reducing this cost through more efficient normalization schemes and on designing task-specific contractive restoration networks that avoid iterative inference while retaining convergence and stability guarantees.

\bibliography{egbib}

@article{ROF92,
  author    = {Leonid I. Rudin and Stanley Osher and Emad Fatemi},
  title     = {Nonlinear total variation based noise removal algorithms},
  journal   = {Physica D: Nonlinear Phenomena},
  volume    = {60},
  number    = {1-4},
  pages     = {259--268},
  year      = {1992}
}

@article{SRCNN14,
  author  = {Chao Dong and Chen Change Loy and Kaiming He and Xiaoou Tang},
  title   = {Learning a Deep Convolutional Network for Image Super-Resolution},
  journal = {Proc. European Conference on Computer Vision},
  year    = {2014},
  pages   = {184--199}
}

@article{DnCNN17,
  author    = {Kai Zhang and Wangmeng Zuo and Yunjin Chen and Deyu Meng and Lei Zhang},
  title     = {Beyond a Gaussian Denoiser: Residual Learning of Deep CNN for Image Denoising},
  journal   = {IEEE Transactions on Image Processing},  
  volume    = {26},
  number    = {7},
  pages     = {3142--3155},
  year      = {2017}
}

@article{FFDNet18,
  author    = {Kai Zhang and Wangmeng Zuo and Lei Zhang},
  title     = {{FFDNet}: Toward a Fast and Flexible Solution for CNN-based Image Denoising},
  journal   = {IEEE Transactions on Image Processing},
  volume    = {27},
  number    = {9},
  pages     = {4608--4622},
  year      = {2018}
}

@article{Kim2016VDSR,
  author  = {Jiwon Kim and Jung Kwon Lee and Kyoung Mu Lee},
  title   = {Accurate Image Super-Resolution Using Very Deep Convolutional Networks},
  journal = {Proc. IEEE Conference on Computer Vision and Pattern Recognition},
  pages     = {1646--1654},
  year    = {2016}
}

@article{Lim2017EDSR,
  author  = {Bee Lim and Sanghyun Son and Heewon Kim and Seungjun Nah and Kyoung Mu Lee},
  title   = {Enhanced Deep Residual Networks for Single Image Super-Resolution},
  journal = {Proc. IEEE Conference on Computer Vision and Pattern Recognition Workshops, NTIRE},
  pages     = {136--144},
  year    = {2017}
}

@article{Liang2021SwinIR,
  author  = {Jingyun Liang and Jiezhang Cao and Guolei Sun and Kai Zhang and Luc Van Gool and Radu Timofte},
  title   = {{SwinIR}: Image Restoration Using Swin Transformer},
  journal = {Proc. IEEE/CVF International Conference on Computer Vision Workshops},
  year    = {2021},
  pages   = {1833--1844},
  doi     = {10.1109/ICCVW54120.2021.00210}
}

@article{chen2021pre,
  author  = {Hanting Chen and Yunhe Wang and Tianyu Guo and Chang Xu and Yiping Deng and Zhenhua Liu and Siwei Ma and Chunjing Xu and Chao Xu and Wen Gao},
  title   = {Pre-trained Image Processing Transformer},
  journal = {Proc. IEEE/CVF Conference on Computer Vision and Pattern Recognition},
  pages   = {12299--12310},
  year    = {2021}
}

@article{Restormer22,
  author  = {Syed Waqas Zamir and Aditya Arora and Salman Khan and Munawar Hayat and Fahad Shahbaz Khan and Ming-Hsuan Yang},
  title   = {Restormer: Efficient Transformer for High-Resolution Image Restoration},
  journal = {Proc. IEEE/CVF Conference on Computer Vision and Pattern Recognition},
  year    = {2022},
  pages   = {5728--5739}
}

@article{NAFNet22,
  title={Simple baselines for image restoration},
  author={Chen, Liangyu and Chu, Xiaojie and Zhang, Xiangyu and Sun, Jian},
  journal={European conference on computer vision},
  pages={17--33},
  year={2022},
  organization={Springer}
}

@article{Gregor2010LISTA,
  author  = {Karol Gregor and Yann LeCun},
  title   = {Learning Fast Approximations of Sparse Coding},
  journal = {Proc. International Conference on Machine Learning },
  year    = {2010},
  pages   = {399--406}
}

@article{Monga2021UnrollingSurvey,
  author  = {Vishal Monga and Yuelong Li and Yonina C. Eldar},
  title   = {Algorithm Unrolling: Interpretable, Efficient Deep Learning for Signal and Image Processing},
  journal = {IEEE Signal Processing Magazine},
  volume  = {38},
  number  = {2},
  pages   = {18--44},
  year    = {2021},
  doi     = {10.1109/MSP.2020.3016145}
}

@article{PnP13,
  author  = {Saiprasad Ravishankar Venkatakrishnan and Charles A. Bouman and Brendt Wohlberg},
  title   = {Plug-and-Play Priors for Model Based Reconstruction},
  journal = {Proc. IEEE Global Conference on Signal and Information Processing },
  year    = {2013},
  pages   = {945--948},
  doi     = {10.1109/GlobalSIP.2013.6737048}
}

@article{RED17,
  author  = {Yaniv Romano and Michael Elad and Peyman Milanfar},
  title   = {The Little Engine That Could: Regularization by Denoising (RED)},
  journal = {SIAM Journal on Imaging Sciences},
  volume  = {10},
  number  = {4},
  pages   = {1804--1844},
  year    = {2017},
  doi     = {10.1137/16M1102884}
}

@incollection{CombettesPesquet2011,
  author    = {Patrick L. Combettes and Jean-Christophe Pesquet},
  title     = {Proximal Splitting Methods in Signal Processing},
  booktitle = {Fixed-Point Algorithms for Inverse Problems in Science and Engineering},
  editor    = {Heinz H. Bauschke and Regina S. Burachik and Patrick L. Combettes and Valerio Elser and D. Russell Luke and Henry Wolkowicz},
  series    = {Springer Optimization and Its Applications},
  volume    = {49},
  pages     = {185--212},
  publisher = {Springer},
  year      = {2011},
  doi       = {10.1007/978-1-4419-9569-8_10}
}

@article{Boyd2011ADMM,
  author  = {Stephen Boyd and Neal Parikh and Eric Chu and Borja Peleato and Jonathan Eckstein},
  title   = {Distributed Optimization and Statistical Learning via the Alternating Direction Method of Multipliers},
  journal = {Foundations and Trends in Machine Learning},
  volume  = {3},
  number  = {1},
  pages   = {1--122},
  year    = {2011},
  doi     = {10.1561/2200000016}
}

@article{Chan2016PlugAndPlay,
  author  = {Stanley H. Chan and Xiran Wang and Omar A. Elgendy},
  title   = {Plug-and-Play ADMM for Image Restoration: Fixed-Point Convergence and Applications},
  journal = {IEEE Transactions on Computational Imaging},
  volume  = {3},
  number  = {1},
  pages   = {84--98},
  year    = {2017},  
  doi     = {10.1109/TCI.2016.2629286}
}

@article{renaud2024snore,
  title   = {Plug-and-Play Image Restoration with Stochastic deNOising REgularization},
  author  = {Renaud, Marien and Prost, Jean and Leclaire, Arthur and Papadakis, Nicolas},
  journal = {Proc. International Conference on Machine Learning },
  year    = {2024}
}

@article{ebner2024pnpconv,
  title        = {Plug-and-Play Image Reconstruction Is a Convergent Regularization Method},
  author       = {Ebner, Andrea and Haltmeier, Markus},
  journal      = {IEEE Transactions on Image Processing},
  year         = {2024},
  volume       = {33},
  pages        = {1476--1486}
}

@article{miyato2018spectral,
  title   = {Spectral Normalization for Generative Adversarial Networks},
  author  = {Takeru Miyato and Toshiki Kataoka and Masanori Koyama and Yuichi Yoshida},
  journal = {Proc. International Conference on Learning Representations},
  year    = {2018},
  url     = {https://openreview.net/forum?id=B1QRgziT-}
}

@article{Ryu2019PnPConvergence,
  author  = {Ernest K. Ryu and Jeffrey Liu and Stephen Wright and Wotao Yin and Xiaohan Chen and Zhiqiang Xu},
  title   = {Plug-and-Play Methods Provably Converge with Properly trained Denoisers},
  journal = {Proc. International Conference on Machine Learning },
  year    = {2019},
  pages   = {5546--5557}
}

@article{Anil2019GroupSort,
  author  = {Chetan R. Anil and James Lucas and Roger B. Grosse},
  title   = {Sorting Out Lipschitz Function Approximation},
  journal = {Proc. International Conference on Machine Learning},
  year    = {2019},
  pages   = {291--301}
}

@article{Sedghi2019Singular,
  author  = {Hanie Sedghi and Vineet Gupta and Philip M. Long},
  title   = {The Singular Values of Convolutional Layers},
  journal = {Proc. International Conference on Learning Representations},
  year    = {2019}
}

@article{Ducotterd2024JMLR,
  author  = {Stanislas Ducotterd and Michael Unser},
  title   = {Improving Lipschitz-Constrained Neural Networks by Learning Activation Functions},
  journal = {Journal of Machine Learning Research},
  volume  = {25},
  pages   = {1--44},
  year    = {2024}
}

@article{Cisse2017Parseval,
  author  = {Moustapha Cisse and Piotr Bojanowski and Edouard Grave and Yann Dauphin and Nicolas Usunier},
  title   = {Parseval Networks: Improving Robustness to Adversarial Examples},
  journal = {Proc. International Conference on Machine Learning},
  year    = {2017},
  pages   = {854--863}
}

@article{wang2020orthogonal,
  title   = {Orthogonal Convolutional Neural Networks},
  author  = {Jiayun Wang and Yubei Chen and Rudrasis Chakraborty and Stella X. Yu},
  journal = {Proc. IEEE/CVF Conference on Computer Vision and Pattern Recognition},
  pages   = {11505--11515},
  year    = {2020}
}

@article{Tsuzuku2018LMT,
  author  = {Yusuke Tsuzuku and Issei Sato and Masashi Sugiyama},
  title   = {Lipschitz-Margin Training: Scalable Certification of Perturbation Invariance for Deep Neural Networks},
  journal = {Proc. Advances in Neural Information Processing Systems},
  volume = {31},
  year    = {2018}
}

@article{Gouk2021LipschitzSurvey,
  author  = {Henry Gouk and Eibe Frank and Bernhard Pfahringer and Michael J. Cree},
  title   = {Regularisation of Neural Networks by Enforcing Lipschitz Continuity},
  journal = {Machine Learning},
  volume  = {110},
  number  = {2},
  pages   = {393--416},
  year    = {2021}
}

@article{behrmann2019invertible,
  title   = {Invertible Residual Networks},
  author  = {Jens Behrmann and Will Grathwohl and Ricky T. Q. Chen and David Duvenaud and J{\"o}rn-Henrik Jacobsen},
  journal = {Proc. International Conference on Machine Learning},
  pages   = {573--582},
  year    = {2019}
}

@article{Prach2024OneLipLayers,
  author  = {Bernd Prach and Fabio Brau and Giorgio Buttazzo and Christoph H. Lampert},
  title   = {1-Lipschitz Layers Compared: Memory, Speed, and Certifiable Robustness},
  journal = {Proc. IEEE/CVF Conference on Computer Vision and Pattern Recognition},
  year    = {2024},
  pages   = {24574--24583},
  doi     = {10.1109/CVPR52733.2024.02320}
}

@article{bredies2024learning,
  title={Learning Firmly Nonexpansive Operators},
  author={Bredies, Kristian and Chirinos-Rodriguez, Jonathan and Naldi, Emanuele},
  journal={arXiv preprint arXiv:2407.14156},
  year={2024}
}

@article{Chang2000Adaptive,
  author  = {Chang, S. Grace and Yu, Bin and Vetterli, Martin},
  title   = {Adaptive Wavelet Thresholding for Image Denoising and Compression},
  journal = {IEEE Transactions on Image Processing},
  year    = {2000},
  volume  = {9},
  number  = {9},
  pages   = {1532--1546},
  doi     = {10.1109/83.862633}
}

@article{Donoho1995SureShrink,
  author  = {Donoho, David L. and Johnstone, Iain M.},
  title   = {Adapting to Unknown Smoothness via Wavelet Shrinkage},
  journal = {Journal of the American Statistical Association},
  year    = {1995},
  volume  = {90},
  number  = {432},
  pages   = {1200--1224},
  doi     = {10.1080/01621459.1995.10476626}
}

@article{ribes2008linear,
  title={Linear inverse problems in imaging},
  author={Ribes, Alejandro and Schmitt, Francis},
  journal={IEEE Signal Processing Magazine},
  volume={25},
  number={4},
  pages={84--99},
  year={2008},
}

@incollection{engl2015regularization,
  title={Regularization of inverse problems},
  author={Engl, Heinz W and Ramlau, Ronny},
  booktitle={Encyclopedia of applied and computational mathematics},
  pages={1233--1241},
  year={2015},
  publisher={Springer}
}

@article{dong2011image,
  title={Image deblurring and super-resolution by adaptive sparse domain selection and adaptive regularization},
  author={Dong, Weisheng and Zhang, Lei and Shi, Guangming and Wu, Xiaolin},
  journal={IEEE Transactions on Image Processing},
  volume={20},
  number={7},
  pages={1838--1857},
  year={2011},
  publisher={IEEE}
}

@article{jagatap2019algorithmic,
  title={Algorithmic Guarantees for Inverse Imaging with Untrained Network Priors},
  author={Jagatap, Gauri and Hegde, Chinmay},
  journal = {Proc. Advances in Neural Information Processing Systems},
  pages = {14832--14842},
  year={2019}
}

@article{goujon2023neural,
  title={A neural-network-based convex regularizer for inverse problems},
  author={Goujon, Alexis and Neumayer, Sebastian and Bohra, Pakshal and Ducotterd, Stanislas and Unser, Michael},
  journal={IEEE Transactions on Computational Imaging},
  volume={9},
  pages={781--795},
  year={2023},
  publisher={IEEE}
}

@article{ParikhBoyd2014Proximal,
  author  = {Neal Parikh and Stephen Boyd},
  title   = {Proximal Algorithms},
  journal = {Foundations and Trends in Optimization},
  volume  = {1},
  number  = {3},
  pages   = {127--239},
  year    = {2014},
  doi     = {10.1561/2400000003}
}

@book{Mallat2009WaveletTour,
  author    = {St{\'e}phane Mallat},
  title     = {A Wavelet Tour of Signal Processing: The Sparse Way},
  publisher = {Academic Press},
  edition   = {3},
  year      = {2009}
}

@book{bauschke2017convex,
  title={{Convex Analysis and Monotone Operator Theory in Hilbert Spaces}},
  author={Bauschke, Heinz H and Combettes, Patrick L},
  address={New York, NY, USA},
  publisher={Springer},
  edition={2},
  year={2017},
}

@article{nair2024averaged,
  title={Averaged deep denoisers for image regularization},
  author={Nair, Pravin and Chaudhury, Kunal N},
  journal={Journal of Mathematical Imaging and Vision},
  volume={66},
  number={3},
  pages={362--379},
  year={2024},
  publisher={Springer}
}

@article{beck2009fast,
  title={A fast iterative shrinkage-thresholding algorithm for linear inverse problems},
  author={Beck, Amir and Teboulle, Marc},
  journal={SIAM Journal of Imaging Sciences},
  volume={2},
  number={1},
  pages={183--202},
  year={2009},
}

@article{levin2009understanding,
  title={Understanding and evaluating blind deconvolution algorithms},
  author={Levin, Anat and Weiss, Yair and Durand, Fredo and Freeman, William T},
  journal={Proc. IEEE Conference on Computer Vision and Pattern Recognition},
  pages={1964--1971},
  year={2009},  
}

@misc{Kodak24,
  author       = {Eastman Kodak Company},
  title        = {Kodak Lossless True Color Image Suite},
  year         = {1999},
  howpublished = {\url{http://r0k.us/graphics/kodak/}}
}

@article{huang2015single,
  title   = {Single Image Super-Resolution from Transformed Self-Exemplars},
  author  = {Jia-Bin Huang and Abhishek Singh and Narendra Ahuja},
  journal = {Proc. IEEE Conference on Computer Vision and Pattern Recognition},
  pages   = {5197--5206},
  year    = {2015}
}

@article{martin2001database,
  title   = {A Database of Human Segmented Natural Images and Its Application to Evaluating Segmentation Algorithms and Measuring Ecological Statistics},
  author  = {David Martin and Charless Fowlkes and Doron Tal and Jitendra Malik},
  journal = {Proc. IEEE International Conference on Computer Vision},
  volume  = {2},
  pages   = {416--423},
  year    = {2001}
}

@article{pielawski2020introducing,
  title={Introducing Hann windows for reducing edge-effects in patch-based image segmentation},
  author={Pielawski, Nicolas and W{\"a}hlby, Carolina},
  journal={PloS one},
  volume={15},
  number={3},
  pages={e0229839},
  year={2020},
  publisher={Public Library of Science San Francisco, CA USA}
}

@article{AgustssonTimofte2017DIV2K,
  author    = {Eirikur Agustsson and Radu Timofte},
  title     = {NTIRE 2017 Challenge on Single Image Super-Resolution: Dataset and Study},
  journal = {Proc. IEEE Conf. on Computer Vision and Pattern Recognition Workshops},
  year      = {2017}
}

@article{Zhang2011ColorDemosaicking,
  author  = {Lei Zhang and Xiaolin Wu and Antoni Buades and Xin Li},
  title   = {Color Demosaicking by Local Directional Interpolation and Non-Local Adaptive Thresholding},
  journal = {Journal of Electronic Imaging},
  volume  = {20},
  number  = {2},
  pages   = {023016},
  year    = {2011}
}

@article{kim2025idf,
  title={Idf: Iterative dynamic filtering networks for generalizable image denoising},
  author={Kim, Dongjin and Ko, Jaekyun and Ali, Muhammad Kashif and Kim, Tae Hyun},
  journal={Proc. IEEE/CVF International Conference on Computer Vision},
  pages={12180--12190},
  year={2025}
}

@article{herbreteau2025self,
  title={Self-calibrated variance-stabilizing transformations for real-world image denoising},
  author={Herbreteau, S{\'e}bastien and Unser, Michael},
  journal={Proc. IEEE/CVF International Conference on Computer Vision},
  pages={10496--10506},
  year={2025}
}

@article{xu2024provably,
  title={Provably robust score-based diffusion posterior sampling for plug-and-play image reconstruction},
  author={Xu, Xingyu and Chi, Yuejie},
  journal={Proc. Advances in Neural Information Processing Systems},
  volume={37},
  pages={36148--36184},
  year={2024}
}

@article{hurault2022proximal,
  title={Proximal denoiser for convergent plug-and-play optimization with nonconvex regularization},
  author={Hurault, Samuel and Leclaire, Arthur and Papadakis, Nicolas},
  journal={Proc. International Conference on Machine Learning},
  pages={9483--9505},
  year={2022}
}

@article{hurault2022gradient,
title={Gradient Step Denoiser for convergent Plug-and-Play},
author={Samuel Hurault and Arthur Leclaire and Nicolas Papadakis},
journal={Proc. International Conference on Learning Representations},
year={2022}
}

@article{jiang2025survey,
  title={A survey on all-in-one image restoration: Taxonomy, evaluation and future trends},
  author={Jiang, Junjun and Zuo, Zengyuan and Wu, Gang and Jiang, Kui and Liu, Xianming},
  journal={IEEE Transactions on Pattern Analysis and Machine Intelligence},
  volume={47},
  number={12},
  pages={11892--11911},
  year={2025}
}

@article{kamilov2023plug,
  title={Plug-and-play methods for integrating physical and learned models in computational imaging: Theory, algorithms, and applications},
  author={Kamilov, Ulugbek S and Bouman, Charles A and Buzzard, Gregery T and Wohlberg, Brendt},
  journal={IEEE Signal Processing Magazine},
  volume={40},
  number={1},
  pages={85--97},
  year={2023},
  publisher={IEEE}
}

@article{tan2025pnpOverview,
  title={Solving Imaging Inverse Problems Using Plug-and-Play Denoisers: Regularization and Optimization Perspectives},
  author={Tan, Hong Ye and Mukherjee, Subhadip and Tang, Junqi},
  journal={arXiv preprint arXiv:2509.03475},
  year={2025}
}

@article{daras2024survey,
  title={A Survey on Diffusion Models for Inverse Problems},
  author={Daras, Giannis and Chung, Hyungjin and Lai, Chieh-Hsin and Mitsufuji, Yuki and Ye, Jong Chul and Milanfar, Peyman and Dimakis, Alexandros G. and Delbracio, Mauricio},
  journal={arXiv preprint arXiv:2410.00083},
  year={2024}
}

@article{graikos2022diffusion,
  title={Diffusion models as plug-and-play priors},
  author={Graikos, Alexandros and Malkin, Nikolay and Jojic, Nebojsa and Samaras, Dimitris},
  journal={Proc. Advances in Neural Information Processing Systems},
  volume={35},
  pages={14715--14728},
  year={2022}
}

@inproceedings{zhu2023denoising,
  title={Denoising Diffusion Models for Plug-and-Play Image Restoration},
  author={Zhu, Yuanzhi and Zhang, Kai and Liang, Jingyun and Cao, Jiezhang and Wen, Bihan and Timofte, Radu and Van Gool, Luc},
  booktitle={Proceedings of the IEEE/CVF Conference on Computer Vision and Pattern Recognition Workshops},
  pages={3929--3938},
  year={2023}
}

@article{park2026stochastic,
  title={Stochastic generative plug-and-play priors},
  author={Park, Chicago Y and Chandler, Edward P and Hu, Yuyang and McCann, Michael T and Garcia-Cardona, Cristina and Wohlberg, Brendt and Kamilov, Ulugbek S},
  journal={arXiv preprint arXiv:2604.03603},
  year={2026}
}


\appendix

\section{Appendix}
\label{sec:appendix}

In this appendix, we provide additional experiments and analyses
that complement the results presented in the main paper.

\subsection{{Ablation Study}}
\label{app:ablation}

\begin{figure}[!htbp]
\begin{center}
\setlength{\tabcolsep}{1pt}
\renewcommand{\arraystretch}{1.0}
\scriptsize

\begin{tabular}{cc}

\begin{tabular}{c}
\includegraphics[width=0.27\linewidth]{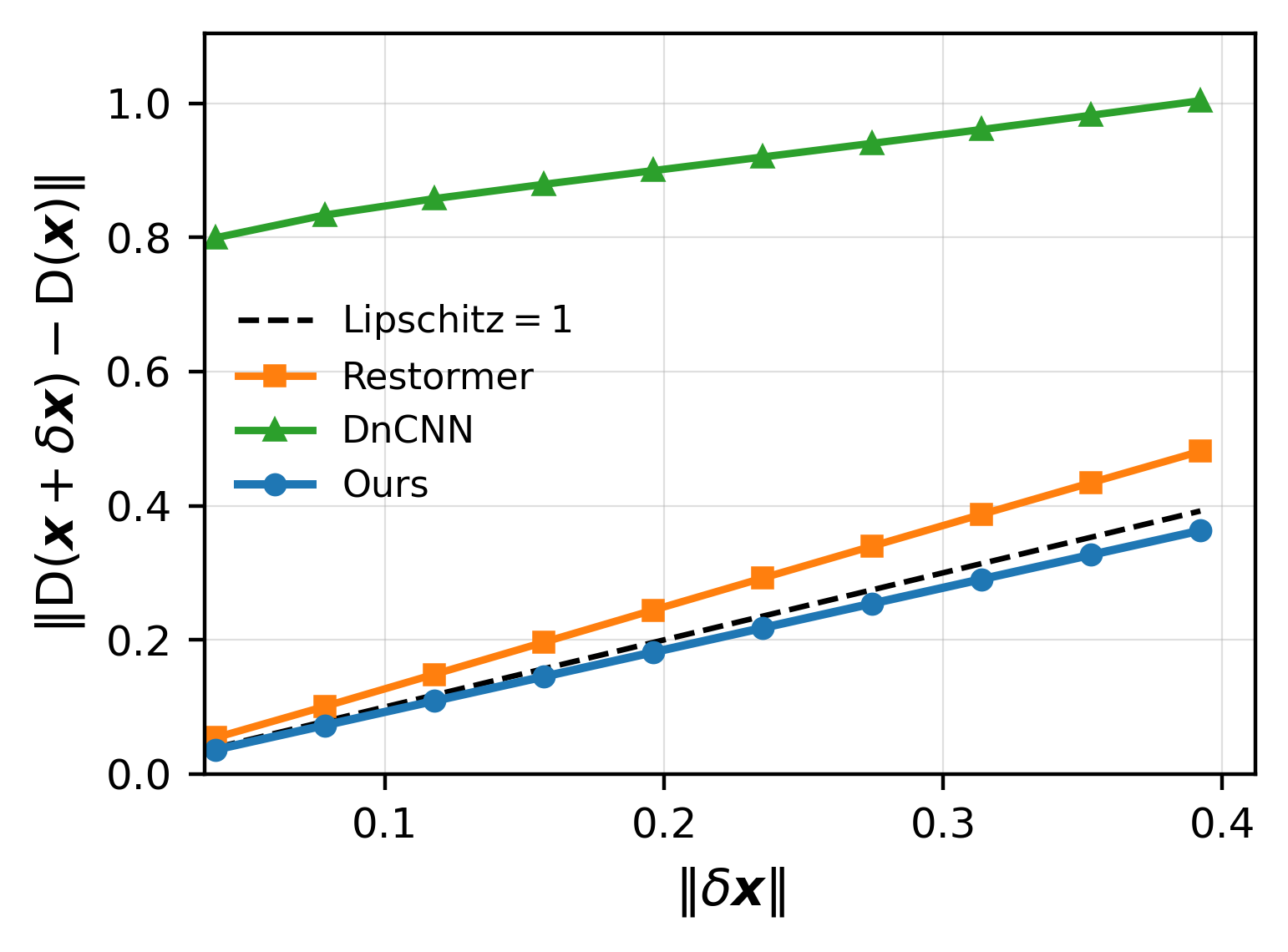} \\
Output vs.\ error norm.
\end{tabular}
&
\begin{tabular}{c}
\includegraphics[width=0.72\linewidth]{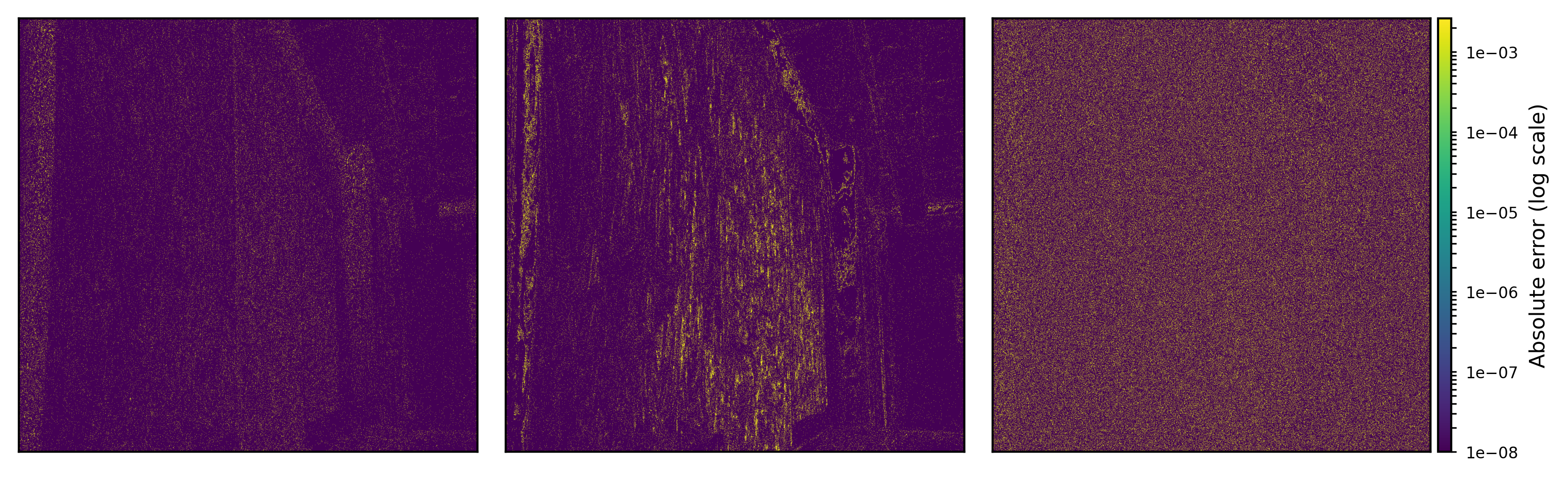} \\
Error heatmaps of $\D(\mathbf{x}+\delta\mathbf{x})-\D(\mathbf{x})$. 
Left to right: $\D$ (Ours, Restormer, DnCNN).
\end{tabular}

\end{tabular}
\end{center}
\caption{Perturbation analysis. Left: output variation versus perturbation norm. Right: spatial error heatmaps showing $\D(\mathbf{x}+\delta\mathbf{x})-\D(\mathbf{x})$ for different denoisers.}
\label{fig:perturbation_analysis}

\end{figure}

\noindent\textbf{Perturbation analysis.}
We analyze the sensitivity of different denoisers to adversarial
loss-gradient perturbations in Fig.~\ref{fig:perturbation_analysis}. The plot on the left reports the output error norm as a function of the perturbation magnitude $\|\delta x\|$. Compared with unconstrained denoisers such as Restormer and DnCNN, the proposed contractive denoiser exhibits a much slower growth in output deviation, indicating reduced sensitivity to input perturbations. The right figure visualizes the spatial error maps
$|D(x+\delta x)-D(x)|$ for different denoisers. Consistent with the quantitative trend, Restormer and DnCNN exhibit large perturbation-induced errors. In contrast, the proposed method produces nearly negligible error, demonstrating that the contractive design strongly suppresses the effect of input perturbations.

\begin{figure}
\begin{center}
\setlength{\tabcolsep}{3pt}
\renewcommand{\arraystretch}{1.0}
\scriptsize

\begin{tabular}{c c c c}

\multirow{3}{*}{
\begin{tabular}{c}
\includegraphics[width=0.22\linewidth]{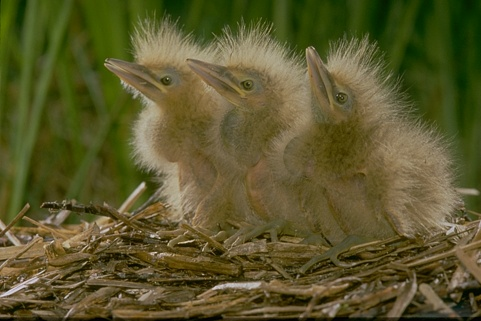} \\
Reference
\end{tabular}
}
&
\includegraphics[width=0.22\linewidth]{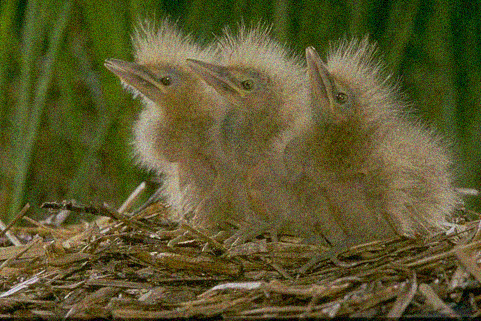} &
\includegraphics[width=0.22\linewidth]{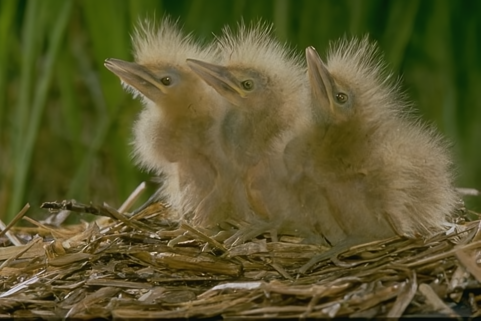} &
\includegraphics[width=0.22\linewidth]{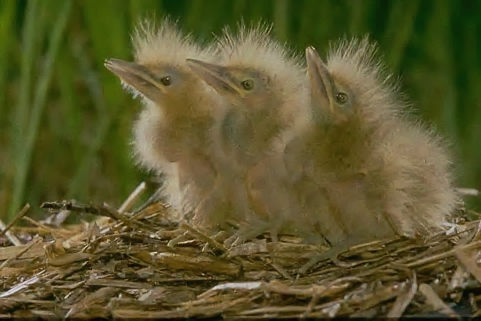} \\

&
Noisy ($24.74$) &
Restormer ($34.36$) &
Ours ($33.48$) \\[6pt]

&
\includegraphics[width=0.22\linewidth]{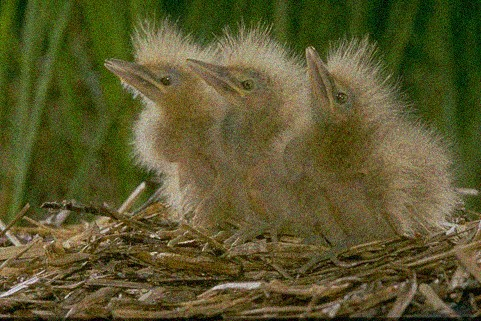} &
\includegraphics[width=0.22\linewidth]{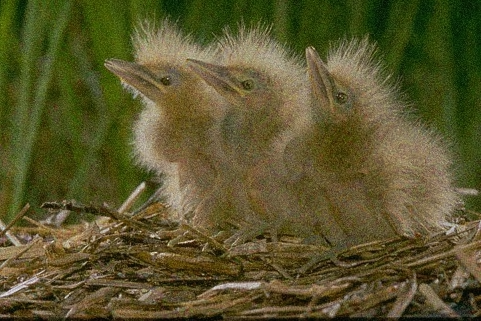} &
\includegraphics[width=0.22\linewidth]{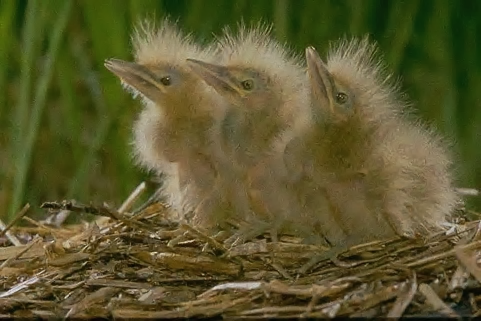} \\

&
Noisy+JPEG ($27.19$) &
Restormer ($27.98$) &
Ours ($32.43$)

\end{tabular}
\end{center}
\caption{JPEG robustness ($90\%$ quality). Our contractive denoiser remains stable, while Restormer degrades under mild compression. Values indicate PSNR in dB.}
\label{fig:jpeg_robustness_appendix}

\end{figure}

\begin{figure}[!t]
\begin{center}
\setlength{\tabcolsep}{3pt}
\renewcommand{\arraystretch}{1.0}
\scriptsize

\begin{tabular}{c c c c}

\multirow{3}{*}{
\begin{tabular}{c}
\includegraphics[width=0.22\linewidth]{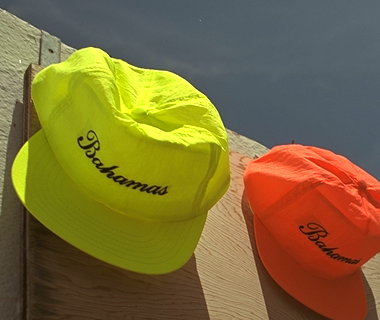} \\
Reference
\end{tabular}
}
&
\includegraphics[width=0.22\linewidth]{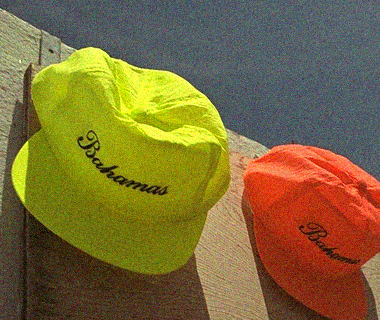} &
\includegraphics[width=0.22\linewidth]{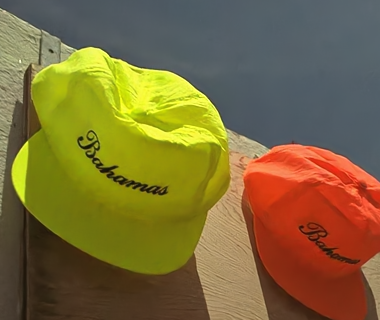} &
\includegraphics[width=0.22\linewidth]{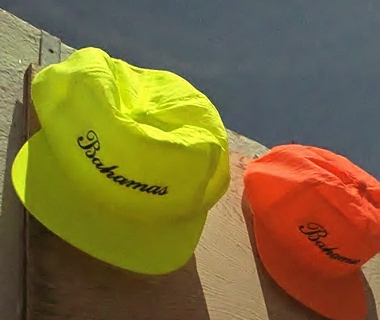} \\

&
Noisy ($24.97$) &
DnCNN ($36.84$) &
Ours ($35.66$) \\[6pt]

&
\includegraphics[width=0.22\linewidth]{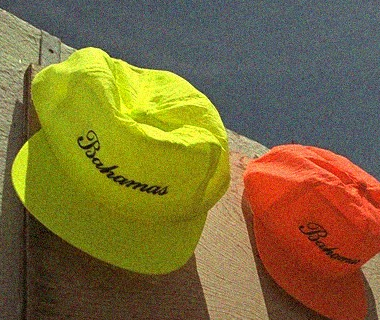} &
\includegraphics[width=0.22\linewidth]{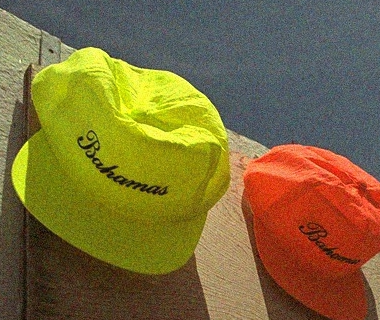} &
\includegraphics[width=0.22\linewidth]{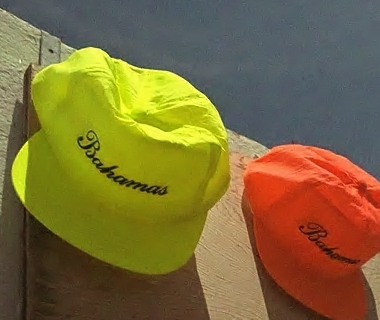} \\

&
Noisy+JPEG ($27.19$) &
DnCNN ($27.61$) &
Ours ($34.40$)

\end{tabular}
\end{center}
\caption{JPEG robustness ($90\%$ quality). Our contractive denoiser remains stable, while DnCNN degrades. PSNR shown.}
\label{fig:jpeg_robustness_kodim03}

\end{figure}

\begin{figure}[t]
\begin{center}
\setlength{\tabcolsep}{3pt}
\renewcommand{\arraystretch}{1.0}
\scriptsize

\begin{tabular}{c c c c}

\multirow{3}{*}{
\begin{tabular}{c}
\includegraphics[width=0.22\linewidth]{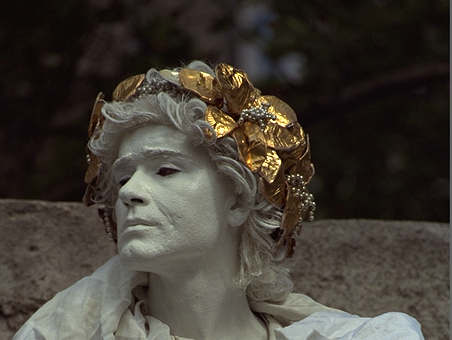} \\
Reference
\end{tabular}
}
&
\includegraphics[width=0.22\linewidth]{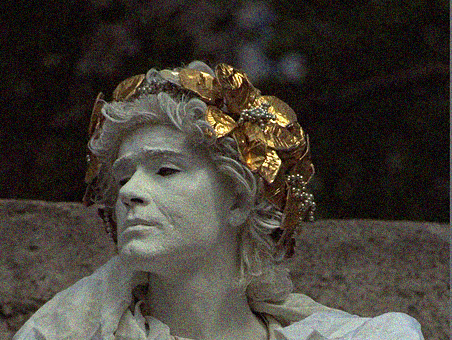} &
\includegraphics[width=0.22\linewidth]{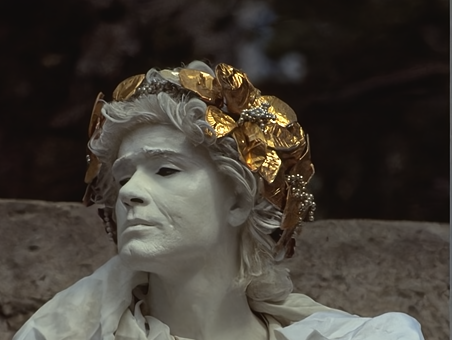} &
\includegraphics[width=0.22\linewidth]{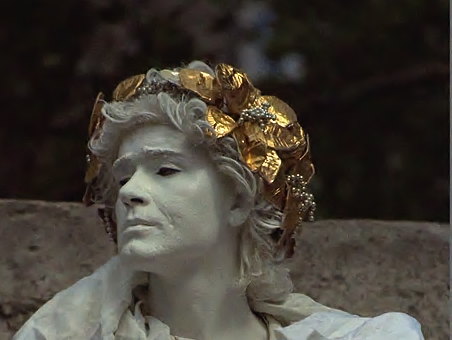} \\

&
Noisy ($24.97$) &
Restormer ($35.67$) &
Ours ($34.63$) \\[6pt]

&
\includegraphics[width=0.22\linewidth]{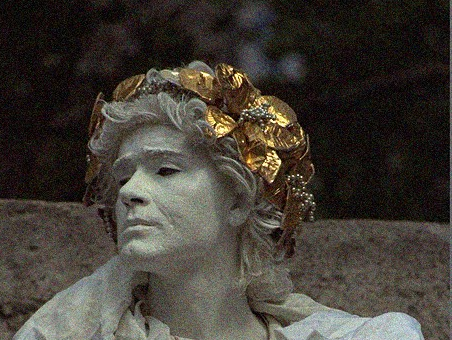} &
\includegraphics[width=0.22\linewidth]{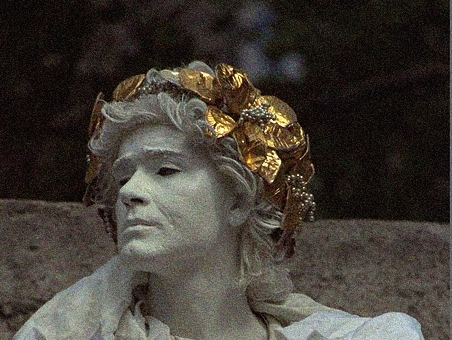} &
\includegraphics[width=0.22\linewidth]{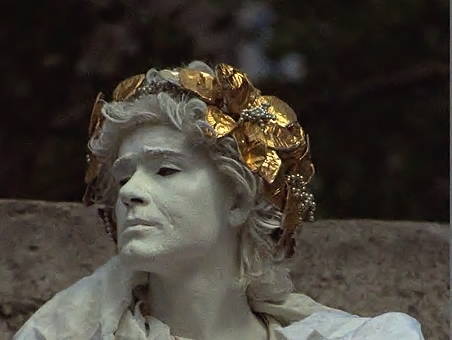} \\

&
Noisy+JPEG ($27.19$) &
Restormer ($27.92$) &
Ours ($33.60$)

\end{tabular}
\end{center}
\caption{JPEG robustness ($90\%$ quality). Our method remains stable under compression, while Restormer is highly sensitive to the perturbation.}
\label{fig:jpeg_robustness_kodim17}
\end{figure}

Next, we provide additional perturbation results, including JPEG compression in Figs.~\ref{fig:jpeg_robustness_appendix},\ref{fig:jpeg_robustness_kodim03} and \ref{fig:jpeg_robustness_kodim17}, where our model remains stable while DnCNN and Restormer exhibit amplified artifacts and reduced denoising quality.

\subsection{Denoising}\label{sec:appendix_denoising}

We report comprehensive color and grayscale Gaussian denoising results against both constrained and unconstrained models. For color denoising, we compare with the $1$-Lipschitz baselines D-FBS and D-DRS; the unconstrained models IDF, DnCNN, Noise2VST, and Restormer; and the pretrained models CLIP and DIL. Table~\ref{tab:color_denoising_appendix} further extends the evaluation beyond the datasets considered in the main manuscript to DIV$2$K~\cite{AgustssonTimofte2017DIV2K} and McMaster~\cite{Zhang2011ColorDemosaicking}. Since suitable pretrained models were unavailable for CLIP and DIL, their reported results are taken directly from the respective papers and are included only for the settings available therein. Across all datasets and noise levels, our model consistently outperforms the other $1$-Lipschitz models and substantially narrows the performance gap to unconstrained models, despite being strictly contractive.

We additionally extend the grayscale denoising evaluation in Table~\ref{tab:gray_denoising_appendix} by comparing with DnCNN, Restormer, MWCNN, and CRR-NN using both its $t$-step and proximal variants. Since publicly released CRR-NN grayscale models are available only for $\sigma=25$, we report its results only at this noise level. Our model remains competitive across datasets and noise levels while retaining its theoretical contractivity guarantee.

\begin{figure}[t]
\begin{center}
\setlength{\tabcolsep}{3pt}
\renewcommand{\arraystretch}{1.0}
\scriptsize

\begin{tabular}{c c c c}

\multirow{3}{*}{
\begin{tabular}{c}
\includegraphics[width=0.22\linewidth]{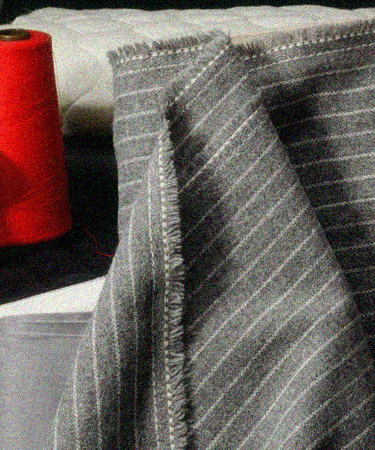} \\
Noisy ($24.94$ dB / $0.5705$)
\end{tabular}
}
&
\includegraphics[width=0.22\linewidth]{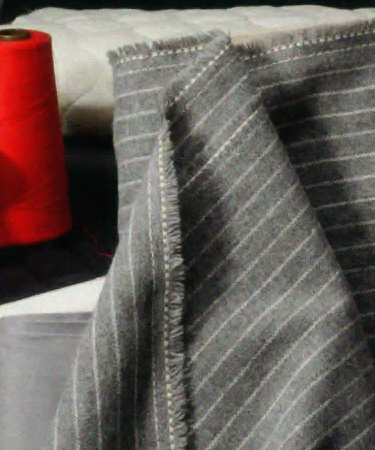} &
\includegraphics[width=0.22\linewidth]{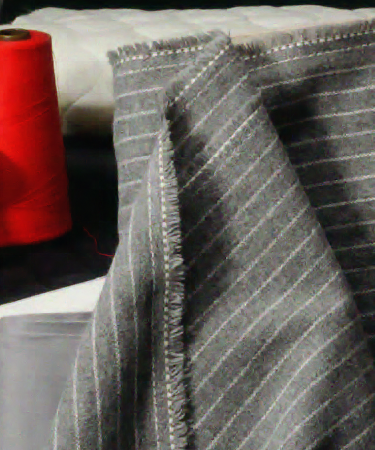} &
\includegraphics[width=0.22\linewidth]{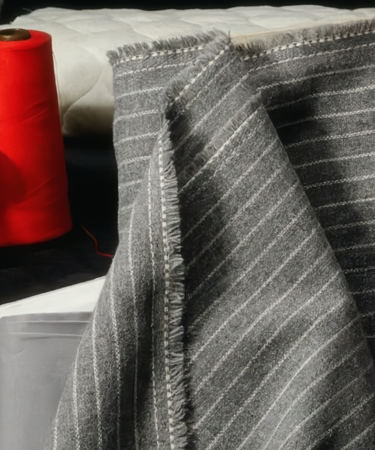} \\

&
D-FBS ($29.84$ dB / $0.8013$) &
D-DRS ($29.84$ dB / $0.8037$) &
Restormer ($33.12$ dB / $0.8880$) \\[6pt]

&
\includegraphics[width=0.22\linewidth]{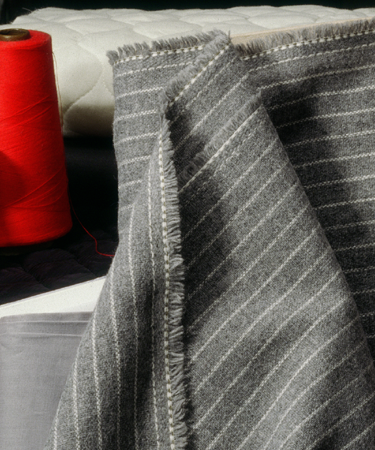} &
\includegraphics[width=0.22\linewidth]{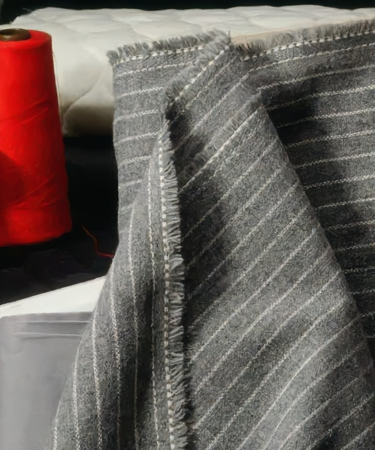} &
\includegraphics[width=0.22\linewidth]{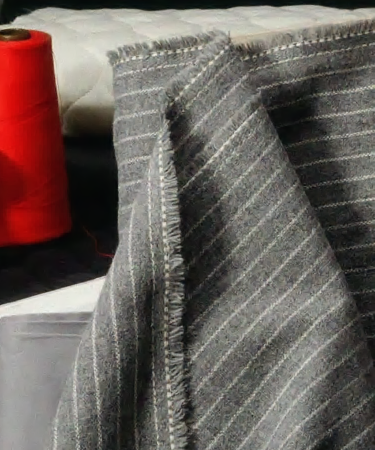} \\

&
Reference &
DnCNN ($33.02$ dB / $0.8862$) &
Ours ($32.36$ dB / $0.8730$)

\end{tabular}
\end{center}
\caption{Color Gaussian denoising on an image from the McMaster dataset ($\sigma=15/255$). Our contractive model preserves fine fabric textures comparable to Restormer, while D-FBS and D-DRS exhibit over-smoothing.}
\label{fig:color_gaussian1}

\end{figure}

\begin{figure}[t]
\begin{center}
\setlength{\tabcolsep}{3pt}
\renewcommand{\arraystretch}{1.0}
\scriptsize

\begin{tabular}{c c c c}

\multirow{3}{*}{
\begin{tabular}{c}
\includegraphics[width=0.22\linewidth]{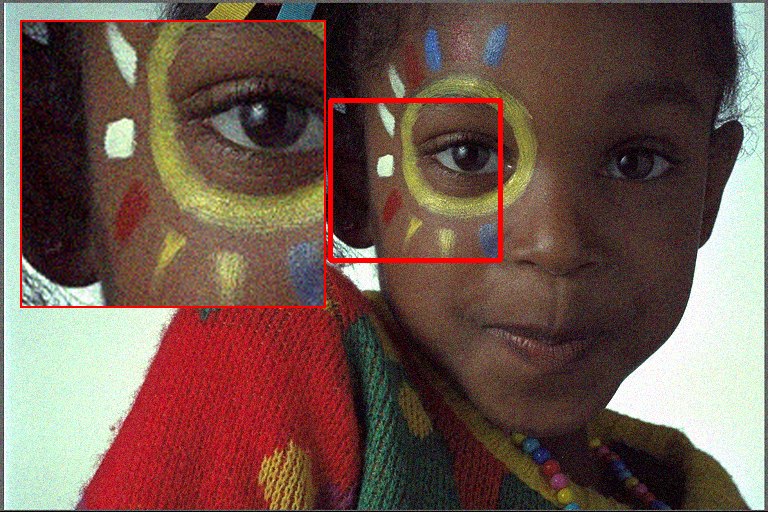} \\
Noisy ($24.97$ dB / $0.4352$)
\end{tabular}
}
&
\includegraphics[width=0.22\linewidth]{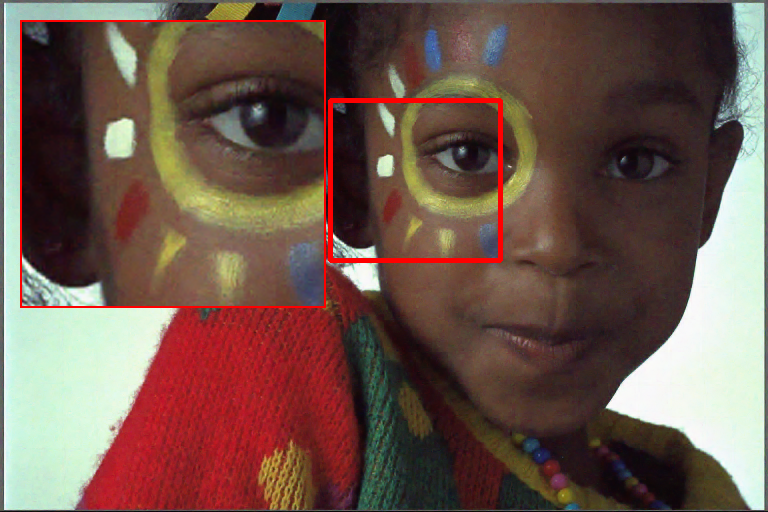} &
\includegraphics[width=0.22\linewidth]{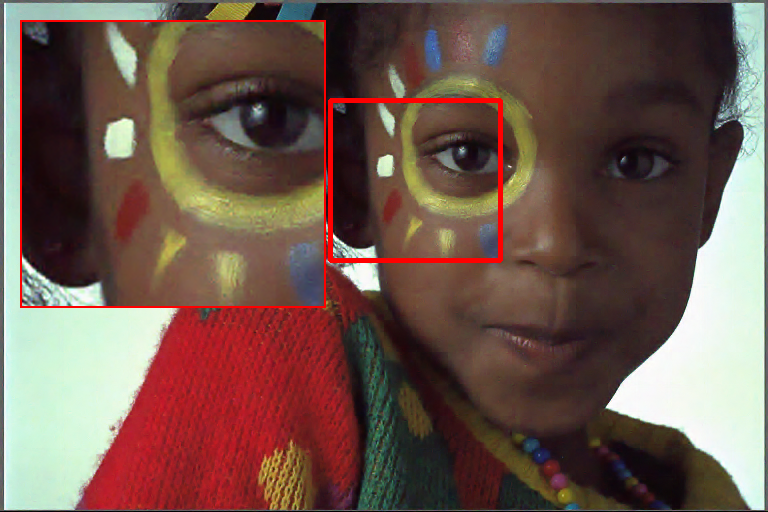} &
\includegraphics[width=0.22\linewidth]{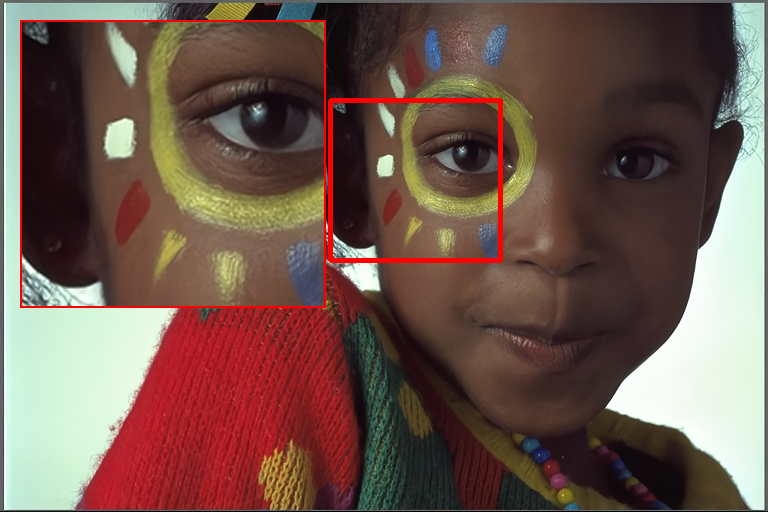} \\

&
D-FBS ($31.79$ dB / $0.8302$) &
D-DRS ($32.20$ dB / $0.8473$) &
Restormer ($34.80$ dB / $0.9099$) \\[6pt]

&
\includegraphics[width=0.22\linewidth]{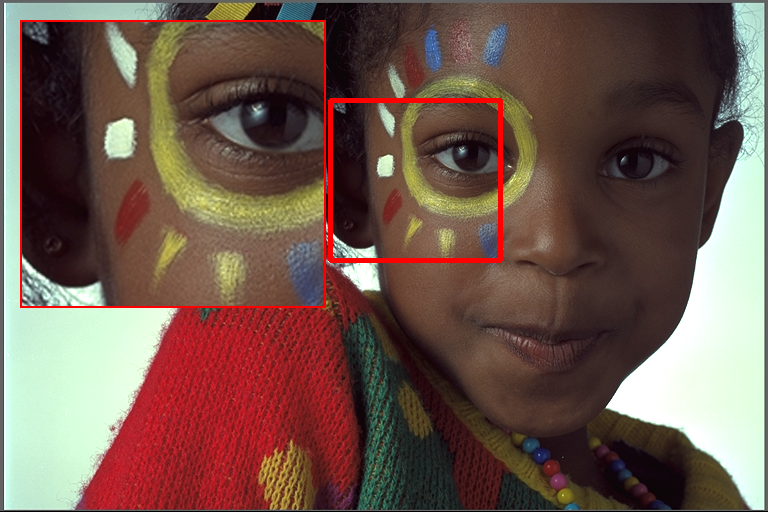} &
\includegraphics[width=0.22\linewidth]{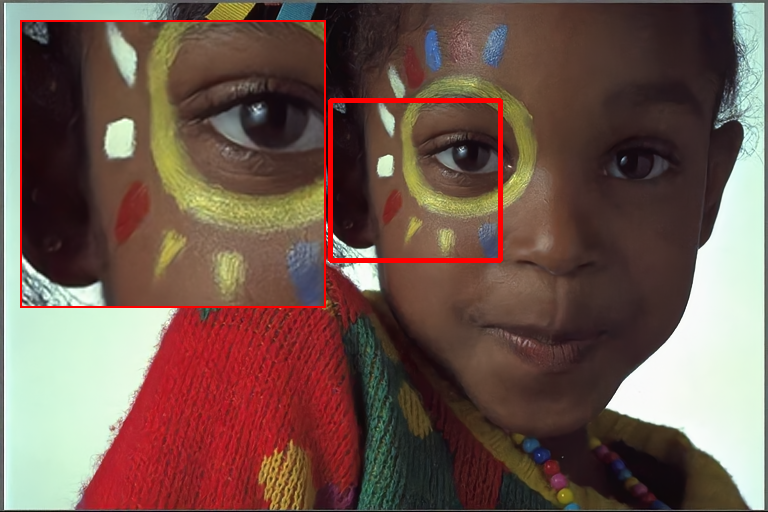} &
\includegraphics[width=0.22\linewidth]{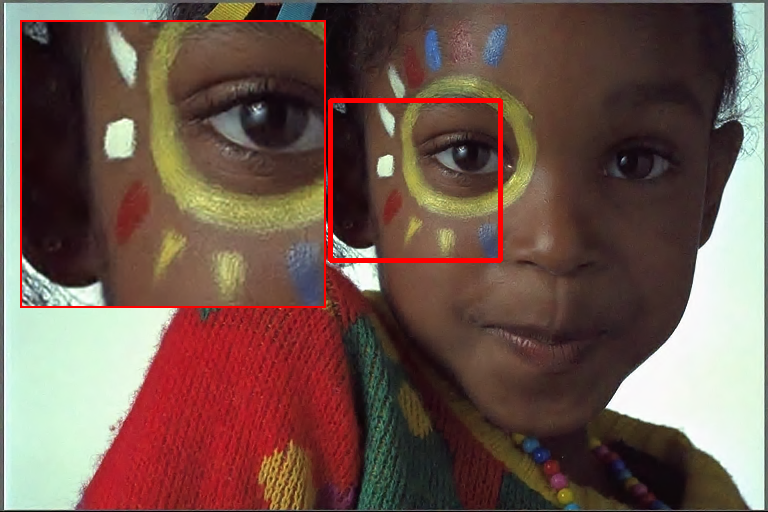} \\

&
Reference &
DnCNN ($34.75$ dB / $0.9115$) &
Ours ($34.02$ dB / $0.8962$)

\end{tabular}
\end{center}
\caption{Color Gaussian denoising on kodim15 (Kodak24). Our method preserves fine textures better than $1$-Lipschitz baselines while remaining competitive with DnCNN and Restormer. PSNR/SSIM shown.}
\label{fig:color_gaussian3}

\end{figure}

Finally, we present two additional qualitative examples in Figs.~\ref{fig:color_gaussian1} and~\ref{fig:color_gaussian3} on highly textured images. Unlike the other $1$-Lipschitz models, D-FBS and D-DRS, our contractive denoiser better preserves fine textures instead of excessively suppressing them and produces reconstructions visually comparable to unconstrained denoisers.

\begin{table*}[!t]
\begin{center}

\setlength{\tabcolsep}{3pt}
\scriptsize

\resizebox{\textwidth}{!}{
\begin{tabular}{lccc|cccc|cc}
\toprule

\multirow{2}{*}{\textbf{Dataset, $\sigma$}}
& \multicolumn{3}{c|}{\textbf{$1$-Lipschitz / Contractive Models}}
& \multicolumn{4}{c|}{\textbf{Unconstrained Models}}
& \multicolumn{2}{c}{\textbf{Pretrained Models}} \\

\cmidrule(lr){2-4}
\cmidrule(lr){5-8}
\cmidrule(lr){9-10}

& \textbf{D-FBS}
& \textbf{D-DRS}
& \textbf{Ours}
& \textbf{IDF}
& \textbf{DnCNN}
& \textbf{Noise2VST}
& \textbf{Restormer}
& \textbf{CLIP}
& \textbf{DIL} \\

\midrule

CBSD68, $\sigma{=}15$
& {\scriptsize $30.12/0.8409$}
& {\scriptsize $30.07/0.8526$}
& {\scriptsize $\mathbf{32.74/0.9115}$}
& {\scriptsize $32.18/0.8915$}
& {\scriptsize $33.63/0.9296$}
& {\scriptsize $33.71/0.9288$}
& {\scriptsize $\mathbf{33.85/0.9316}$}
& {\scriptsize $33.72/0.921$}
& {\scriptsize $32.50/0.906$} \\

CBSD68, $\sigma{=}25$
& {\scriptsize $27.54/0.7577$}
& {\scriptsize $27.51/0.7695$}
& {\scriptsize $\mathbf{29.78/0.8501}$}
& {\scriptsize $29.80/0.8435$}
& {\scriptsize $30.72/0.8806$}
& {\scriptsize $\mathbf{31.03/0.8877}$}
& {\scriptsize $30.90/0.8837$}
& {\scriptsize $31.02/0.878$}
& {\scriptsize $29.98/0.843$} \\

CBSD68, $\sigma{=}50$
& {\scriptsize $23.87/0.6206$}
& {\scriptsize $24.11/0.6266$}
& {\scriptsize $\mathbf{25.61/0.7211}$}
& {\scriptsize $25.73/0.7188$}
& {\scriptsize $26.52/0.7714$}
& {\scriptsize $25.93/0.7450$}
& {\scriptsize $\mathbf{26.65/0.7765}$}
& {\scriptsize $26.69/0.7311$}
& {\scriptsize $26.43/0.7170$} \\

\midrule

Kodak24, $\sigma{=}15$
& {\scriptsize $31.20/0.8315$}
& {\scriptsize $31.05/0.8465$}
& {\scriptsize $\mathbf{33.42/0.9020}$}
& {\scriptsize $32.87/0.8839$}
& {\scriptsize $34.37/0.9219$}
& {\scriptsize $\mathbf{34.64/0.9221}$}
& {\scriptsize $34.62/0.9237$}
& {\scriptsize $34.54/0.914$}
& {\scriptsize $33.37/0.912$} \\

Kodak24, $\sigma{=}25$
& {\scriptsize $28.44/0.7558$}
& {\scriptsize $28.58/0.7710$}
& {\scriptsize $\mathbf{30.67/0.8446}$}
& {\scriptsize $30.64/0.8410$}
& {\scriptsize $31.70/0.8837$}
& {\scriptsize $\mathbf{32.33/0.8898}$}
& {\scriptsize $31.93/0.8814$}
& {\scriptsize $31.86/0.871$}
& {\scriptsize $30.99/0.856$} \\

Kodak24, $\sigma{=}50$
& {\scriptsize $24.90/0.6380$}
& {\scriptsize $25.15/0.6484$}
& {\scriptsize $\mathbf{26.49/0.7285}$}
& {\scriptsize $26.60/0.7224$}
& {\scriptsize $27.63/0.7785$}
& {\scriptsize $27.44/0.7450$}
& {\scriptsize $\mathbf{27.82/0.7901}$}
& {\scriptsize $27.39/0.723$}
& {\scriptsize $27.46/0.736$} \\

\midrule

Urban100, $\sigma{=}15$
& {\scriptsize $29.53/0.8519$}
& {\scriptsize $29.73/0.8687$}
& {\scriptsize $\mathbf{31.96/0.9092}$}
& {\scriptsize $31.42/0.8929$}
& {\scriptsize $32.65/0.9199$}
& {\scriptsize $\mathbf{33.58/0.9253}$}
& {\scriptsize $33.16/0.9229$}
& {\scriptsize $33.07/0.918$}
& {\scriptsize $32.15/0.920$} \\

Urban100, $\sigma{=}25$
& {\scriptsize $26.34/0.7729$}
& {\scriptsize $26.70/0.7934$}
& {\scriptsize $\mathbf{28.94/0.8585}$}
& {\scriptsize $29.28/0.8571$}
& {\scriptsize $30.11/0.8837$}
& {\scriptsize $\mathbf{31.43/0.9032}$}
& {\scriptsize $30.55/0.8900$}
& {\scriptsize $30.72/0.893$}
& {\scriptsize $29.72/0.876$} \\

Urban100, $\sigma{=}50$
& {\scriptsize $22.02/0.6273$}
& {\scriptsize $22.52/0.6438$}
& {\scriptsize $\mathbf{24.40/0.7441}$}
& {\scriptsize $25.06/0.7547$}
& {\scriptsize $25.74/0.7990$}
& {\scriptsize $\mathbf{26.48/0.8234}$}
& {\scriptsize $26.07/0.8129$}
& {\scriptsize $26.27/0.769$}
& {\scriptsize $25.89/0.768$} \\

\midrule

McMaster, $\sigma{=}15$
& {\scriptsize $31.27/0.8497$}
& {\scriptsize $31.62/0.8491$}
& {\scriptsize $\mathbf{32.30/0.8574}$}
& {\scriptsize $31.98/0.8361$}
& {\scriptsize $33.08/0.8895$}
& {\scriptsize $33.48/0.8927$}
& {\scriptsize $\mathbf{33.51/0.8944}$}
& {\scriptsize $33.86/0.9100$}
& {\scriptsize $32.45/0.8530$} \\

McMaster, $\sigma{=}25$
& {\scriptsize $28.52/0.7717$}
& {\scriptsize $28.86/0.7821$}
& {\scriptsize $\mathbf{29.32/0.8023}$}
& {\scriptsize $29.01/0.7825$}
& {\scriptsize $30.54/0.8364$}
& {\scriptsize $31.77/0.8523$}
& {\scriptsize $\mathbf{31.82/0.8414}$}
& {\scriptsize $31.50/0.8600$}
& {\scriptsize $30.17/0.7910$} \\

McMaster, $\sigma{=}50$
& {\scriptsize $24.36/0.6352$}
& {\scriptsize $24.59/0.6414$}
& {\scriptsize $\mathbf{24.72/0.6661}$}
& {\scriptsize $24.61/0.6432$}
& {\scriptsize $25.92/0.7216$}
& {\scriptsize $\mathbf{26.12/0.7341}$}
& {\scriptsize $26.11/0.7271$}
& {\scriptsize $27.43/0.727$}
& {\scriptsize $26.61/0.6691$} \\

\midrule

DIV2K, $\sigma{=}15$
& {\scriptsize $31.21/0.8441$}
& {\scriptsize $31.75/0.8620$}
& {\scriptsize $\mathbf{33.71/0.9020}$}
& {\scriptsize $33.59/0.8944$}
& {\scriptsize $34.40/0.9131$}
& {\scriptsize $34.61/0.9132$}
& {\scriptsize $\mathbf{34.62/0.9144}$}
& {\scriptsize $-$}
& {\scriptsize $-$} \\

DIV2K, $\sigma{=}25$
& {\scriptsize $28.61/0.7784$}
& {\scriptsize $29.08/0.7963$}
& {\scriptsize $\mathbf{30.67/0.8361}$}
& {\scriptsize $30.41/0.8281$}
& {\scriptsize $31.67/0.8717$}
& {\scriptsize $31.84/0.8733$}
& {\scriptsize $\mathbf{31.87/0.8751}$}
& {\scriptsize $-$}
& {\scriptsize $-$} \\

DIV2K, $\sigma{=}50$
& {\scriptsize $24.40/0.6350$}
& {\scriptsize $24.65/0.6450$}
& {\scriptsize $\mathbf{25.91/0.7340}$}
& {\scriptsize $25.65/0.7123$}
& {\scriptsize $26.95/0.7828$}
& {\scriptsize $\mathbf{27.12/0.7913}$}
& {\scriptsize $27.09/0.7889$}
& {\scriptsize $-$}
& {\scriptsize $-$} \\

\bottomrule
\end{tabular}
}
\end{center}
\caption{\textbf{Color denoising (PSNR (dB)/SSIM).}
Our model outperforms the other $1$-Lipschitz/contractive models and narrows the gap with unconstrained models. CLIP and DIL results are taken directly from their respective papers due to the unavailability of the required pretrained models.}
\label{tab:color_denoising_appendix}
\end{table*}

\begin{table*}[!t]
\begin{center}

\setlength{\tabcolsep}{3pt}
\scriptsize

\resizebox{\textwidth}{!}{
\begin{tabular}{lccc|cc|ccc}
\toprule

\multirow{2}{*}{\textbf{Dataset, $\sigma$}}
& \multicolumn{3}{c|}{\textbf{$1$-Lipschitz / Contractive Models}}
& \multicolumn{2}{c|}{\textbf{Unconstrained Models}}
& \multicolumn{3}{c}{\textbf{Pretrained Models}} \\

\cmidrule(lr){2-4}
\cmidrule(lr){5-6}
\cmidrule(lr){7-9}

& \textbf{D-FBS}
& \textbf{D-DRS}
& \textbf{Ours}
& \textbf{DnCNN}
& \textbf{Restormer}
& \textbf{MWCNN}
& \textbf{CRR-NN ($t$-step)}
& \textbf{CRR-NN (prox.)} \\

\midrule

BSD68, $\sigma{=}15$
& {\scriptsize $29.89/0.8347$}
& {\scriptsize $29.79/0.8560$}
& {\scriptsize $\mathbf{30.52/0.8690}$}
& {\scriptsize $\mathbf{31.58/0.8934}$}
& {\scriptsize $31.47/0.8930$}
& {\scriptsize $31.86/0.8947$}
& {\scriptsize $-$}
& {\scriptsize $-$} \\

BSD68, $\sigma{=}25$
& {\scriptsize $27.31/0.7537$}
& {\scriptsize $27.24/0.7600$}
& {\scriptsize $\mathbf{27.82/0.7883}$}
& {\scriptsize $\mathbf{28.87/0.8254}$}
& {\scriptsize $28.75/0.8233$}
& {\scriptsize $29.41/0.8360$}
& {\scriptsize $28.06/0.7838$}
& {\scriptsize $28.05/0.7825$} \\

BSD68, $\sigma{=}50$
& {\scriptsize $24.21/0.6290$}
& {\scriptsize $24.18/0.6368$}
& {\scriptsize $\mathbf{24.40/0.6460}$}
& {\scriptsize $\mathbf{25.12/0.6846}$}
& {\scriptsize $24.35/0.6288$}
& {\scriptsize $26.53/0.7366$}
& {\scriptsize $-$}
& {\scriptsize $-$} \\

\midrule

Set12, $\sigma{=}15$
& {\scriptsize $30.41/0.8451$}
& {\scriptsize $30.39/0.8535$}
& {\scriptsize $\mathbf{31.43/0.8816}$}
& {\scriptsize $\mathbf{32.82/0.9044}$}
& {\scriptsize $32.69/0.9020$}
& {\scriptsize $33.15/0.9088$}
& {\scriptsize $-$}
& {\scriptsize $-$} \\

Set12, $\sigma{=}25$
& {\scriptsize $28.23/0.7924$}
& {\scriptsize $28.15/0.8136$}
& {\scriptsize $\mathbf{28.71/0.8219}$}
& {\scriptsize $\mathbf{30.32/0.8609}$}
& {\scriptsize $30.13/0.8574$}
& {\scriptsize $30.79/0.8711$}
& {\scriptsize $28.91/0.8157$}
& {\scriptsize $28.88/0.8100$} \\

Set12, $\sigma{=}50$
& {\scriptsize $24.53/0.6788$}
& {\scriptsize $24.48/0.6815$}
& {\scriptsize $\mathbf{25.19/0.7160}$}
& {\scriptsize $\mathbf{26.54/0.7654}$}
& {\scriptsize $26.15/0.6950$}
& {\scriptsize $27.74/0.8056$}
& {\scriptsize $-$}
& {\scriptsize $-$} \\

\midrule

Urban100, $\sigma{=}15$
& {\scriptsize $29.21/0.8478$}
& {\scriptsize $29.18/0.8774$}
& {\scriptsize $\mathbf{30.11/0.8851}$}
& {\scriptsize $\mathbf{32.36/0.9159}$}
& {\scriptsize $32.20/0.9139$}
& {\scriptsize $33.17/0.9357$}
& {\scriptsize $-$}
& {\scriptsize $-$} \\

Urban100, $\sigma{=}25$
& {\scriptsize $26.48/0.7803$}
& {\scriptsize $26.47/0.8058$}
& {\scriptsize $\mathbf{26.94/0.8118}$}
& {\scriptsize $\mathbf{29.36/0.8677}$}
& {\scriptsize $29.13/0.8630$}
& {\scriptsize $30.66/0.9026$}
& {\scriptsize $27.62/0.8252$}
& {\scriptsize $27.60/0.8209$} \\

Urban100, $\sigma{=}50$
& {\scriptsize $22.43/0.6461$}
& {\scriptsize $22.39/0.6339$}
& {\scriptsize $\mathbf{22.93/0.6690}$}
& {\scriptsize $\mathbf{24.75/0.7481}$}
& {\scriptsize $24.43/0.6798$}
& {\scriptsize $25.42/0.7371$}
& {\scriptsize $-$}
& {\scriptsize $-$} \\

\bottomrule
\end{tabular}
}
\end{center}
\caption{\textbf{Grayscale denoising (PSNR (dB)/SSIM).}
Our model delivers competitive denoising performance across all datasets and noise levels while retaining its theoretical contractivity guarantees. CRR-NN results are reported only at $\sigma=25$ due to the availability of its publicly released pretrained grayscale model. MWCNN results are obtained using publicly available pretrained models.}
\label{tab:gray_denoising_appendix}
\end{table*}

\begin{table}[!t]
\scriptsize
\setlength{\tabcolsep}{2.5pt}
\renewcommand{\arraystretch}{0.80}

\begin{center}
\resizebox{\linewidth}{!}{
\begin{tabular}{llccccccccccc}
\toprule
Dataset & Metric
& FBS 
& DRS 
& IDF 
& Noise2VST 
& DnCNN 
& Restormer
& CRR-NN ($t$-step)
& CRR-NN (prox.)
& MWCNN
& CLIP
& \textbf{OURS} \\
\midrule

\multirow{2}{*}{CBSD68}
& PSNR
& $25.28$ & $25.36$ & $24.54$ & $24.94$
& $25.36$ & $24.51$
& $24.66$ & $24.67$
& $24.36$ & $23.39$
& $\mathbf{25.61}$ \\

& SSIM
& $0.6835$ & $0.6863$ & $0.6267$ & $0.6498$
& $0.6670$ & $0.5822$
& $0.6851$ & $0.6857$
& $0.6675$ & $0.5861$
& $\mathbf{0.7046}$ \\
\midrule

\multirow{2}{*}{Kodak24}
& PSNR
& $26.04$ & $26.12$ & $25.32$ & $25.73$
& $25.96$ & $25.05$
& $25.94$ & $25.94$
& $25.72$ & $24.41$
& $\mathbf{26.35}$ \\

& SSIM
& $0.6994$ & $0.7018$ & $0.6575$ & $0.6705$
& $0.6580$ & $0.5588$
& $0.7023$ & $0.7036$
& $0.6946$ & $0.6188$
& $\mathbf{0.7184}$ \\
\midrule

\multirow{2}{*}{Urban100}
& PSNR
& $22.56$ & $22.71$ & $21.79$ & $22.62$
& $22.83$ & $22.34$
& $22.64$ & $22.64$
& $22.65$ & $21.38$
& $\mathbf{22.89}$ \\

& SSIM
& $0.6750$ & $0.6830$ & $0.6203$ & $0.6700$
& $0.6541$ & $0.5822$
& $0.6860$ & $0.6864$
& $0.6917$ & $0.6007$
& $\mathbf{0.6976}$ \\
\bottomrule
\end{tabular}
}
\end{center}

\caption{PnP-FBS super-resolution ($\times2$) with a $9\times 9$ Gaussian blur
kernel ($\sigma=2$) and noise level $\sigma_n=5/255$.}
\label{tab:sr_color_pnp1}
\end{table}

\begin{table}[!t]
\scriptsize
\setlength{\tabcolsep}{2.5pt}
\renewcommand{\arraystretch}{0.80}

\begin{center}
\resizebox{\linewidth}{!}{
\begin{tabular}{llccccccccccc}
\toprule
Dataset & Metric
& FBS 
& DRS 
& IDF 
& Noise2VST 
& DnCNN 
& Restormer
& CRR-NN ($t$-step)
& CRR-NN (prox.)
& MWCNN
& CLIP
& \textbf{OURS} \\
\midrule

\multirow{2}{*}{CBSD68}
& PSNR
& $25.17$ & $25.25$ & $24.42$ & $24.82$
& $25.30$ & $24.38$
& $24.92$ & $24.93$
& $24.68$ & $23.61$
& $\mathbf{25.51}$ \\

& SSIM
& $0.6773$ & $0.6800$ & $0.6203$ & $0.6427$
& $0.6633$ & $0.5731$
& $0.6904$ & $0.6911$
& $0.6728$ & $0.5892$
& $\mathbf{0.6986}$ \\
\midrule

\multirow{2}{*}{Kodak24}
& PSNR
& $25.95$ & $26.02$ & $25.20$ & $25.63$
& $25.92$ & $24.93$
& $26.12$ & $26.13$
& $26.00$ & $24.57$
& $\textbf{26.26}$ \\

& SSIM
& $0.6945$ & $0.6965$ & $0.6518$ & $0.6652$
& $0.6555$ & $0.5498$
& $0.7129$ & $0.7133$
& $0.7059$ & $0.6243$
& $\mathbf{0.7135}$ \\
\midrule

\multirow{2}{*}{Urban100}
& PSNR
& $22.42$ & $22.57$ & $21.64$ & $22.48$
& $22.74$ & $22.21$
& $22.55$ & $22.54$
& $22.69$ & $21.26$
& $\textbf{22.76}$ \\

& SSIM
& $0.6666$ & $0.6743$ & $0.6114$ & $0.6614$
& $0.6484$ & $0.5719$
& $0.6798$ & $0.6803$
& $0.6812$ & $0.5957$
& $\textbf{0.6896}$ \\
\bottomrule
\end{tabular}
}
\end{center}

\caption{PnP-FBS super-resolution ($\times2$) with a $21\times21$ anisotropic
Gaussian blur kernel ($\sigma_x=3$, $\sigma_y=1$) and noise level
$\sigma_n=5/255$.}
\label{tab:sr_color_aniso1}
\end{table}

\subsection{Superresolution}
\label{sec:appendix_superresolution}
\begin{figure}[!t]
\begin{center}
\setlength{\tabcolsep}{3pt}
\renewcommand{\arraystretch}{1.0}
\scriptsize

\begin{tabular}{c c c c}

\multirow{3}{*}{
\begin{tabular}{c}
\includegraphics[width=0.22\linewidth]{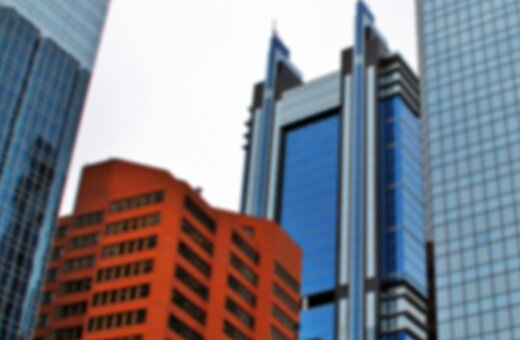} \\
Bicubic ($19.51$ dB / $0.5682$)
\end{tabular}
}
&
\includegraphics[width=0.22\linewidth]{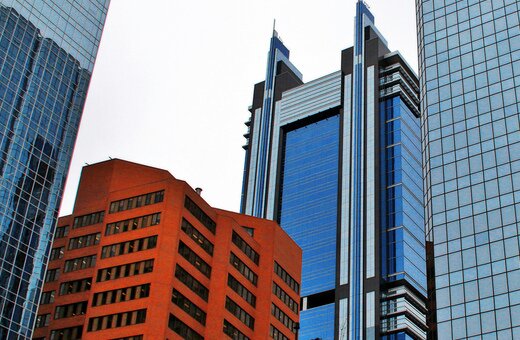} &
\includegraphics[width=0.22\linewidth]{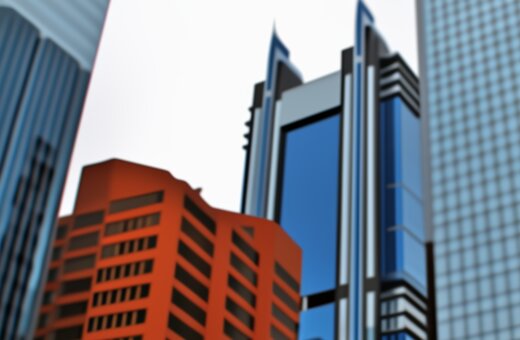} &
\includegraphics[width=0.22\linewidth]{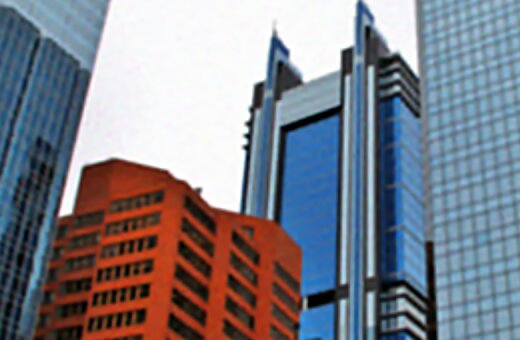} \\

&
Reference &
Noise2VST ($20.94$ dB / $0.5433$) &
D-FBS ($21.14$ dB / $0.5757$) \\[6pt]

&
\includegraphics[width=0.22\linewidth]{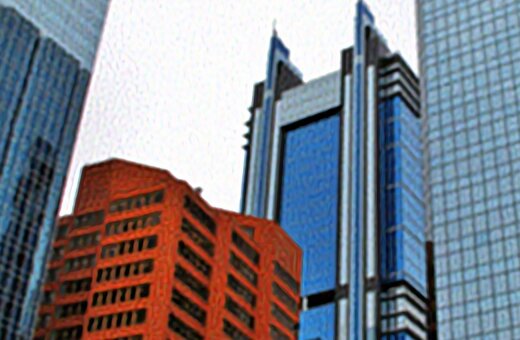} &
\includegraphics[width=0.22\linewidth]{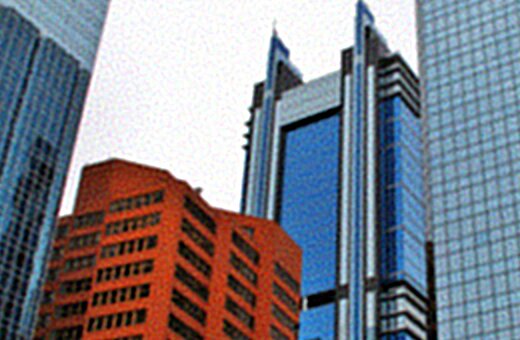} &
\includegraphics[width=0.22\linewidth]{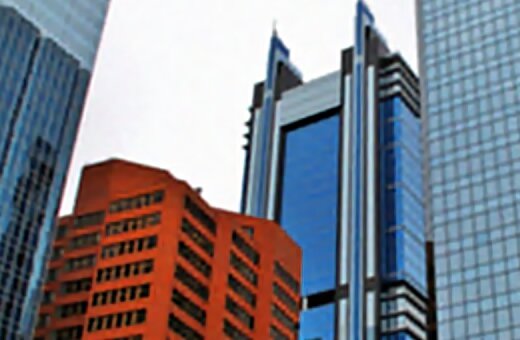} \\

&
DnCNN ($21.34$ dB / $0.5979$) &
Restormer ($21.28$ dB / $0.5558$) &
Ours ($21.35$ dB / $0.5945$)

\end{tabular}
\end{center}
\caption{PnP-FBS super-resolution under Gaussian blur ($25\times25$, $\sigma=1.6$). From bicubic input, our contractive denoiser restores clearer structures and achieves the highest PSNR and competitive SSIM.}
\label{fig:superres_gaussian}

\end{figure}

We next show the regularization capabilities of our denoiser, focusing on superresolution. The forward model is 
\(
\y = \S\B\x + \eta,
\)
where $\S$ is a fixed downsampling operator, $\B$ is a known blur, and
$\eta$ is AWGN for superresolution. We evaluate our model under additional blur settings and a more challenging $\times4$ super-resolution scenario. For $\times2$ super-resolution, we evaluate Gaussian, anisotropic Gaussian, defocus, and box blur in Tables~\ref{tab:sr_color_pnp1}, \ref{tab:sr_color_aniso1}, \ref{tab:defocus_super1}, and~\ref{tab:deblur_sparse_pnp2}, respectively. We compare against the $1$-Lipschitz baselines D-FBS and D-DRS, the unconstrained models IDF, Noise2VST, DnCNN, Restormer, CRR-NN ($t$-step and proximal variants), MWCNN, and CLIP. Our denoiser achieves the best or competitive PSNR/SSIM across CBSD68, Kodak24, and Urban100. We further evaluate $\times4$ super-resolution under anisotropic Gaussian, Gaussian, defocus, and box blur in Tables~\ref{tab:aniso_super}, \ref{tab:gaussian_super}, \ref{tab:defocus_super}, and \ref{tab:box_super} respectively, where our method again achieves the best or competitive results while preserving contraction. These results demonstrate that our contractive denoiser maintains strong reconstruction quality across diverse degradations. This trend is also reflected in the visual comparisons in Figures~\ref{fig:superres_gaussian} and~\ref{fig:superres2}, where our method restores sharper structures with fewer artifacts.

\begin{table}[!t]
\scriptsize
\setlength{\tabcolsep}{2.5pt}
\renewcommand{\arraystretch}{0.80}

\begin{center}
\resizebox{\linewidth}{!}{
\begin{tabular}{llccccccccccc}
\toprule
Dataset & Metric
& FBS
& DRS
& IDF
& Noise2VST
& DnCNN
& Restormer
& CRR-NN ($t$-step)
& CRR-NN (prox.)
& MWCNN
& CLIP
& \textbf{OURS} \\
\midrule

\multirow{2}{*}{CBSD68}
& PSNR
& $24.10$
& $24.14$
& $23.63$
& $23.78$
& $24.21$
& $23.81$
& $23.54$
& $23.55$
& $23.23$
& $22.70$
& $\textbf{24.24}$ \\

& SSIM
& $0.6122$
& $0.6123$
& $0.5778$
& $0.5874$
& $0.6001$
& $0.5468$
& $0.6093$
& $0.6100$
& $0.5911$
& $0.5371$
& $\textbf{0.6219}$ \\
\midrule

\multirow{2}{*}{Kodak24}
& PSNR
& $24.93$
& $24.97$
& $24.43$
& $24.66$
& $24.92$
& $24.52$
& $24.93$
& $24.94$
& $24.69$
& $23.88$
& $\textbf{25.08}$ \\

& SSIM
& $0.6404$
& $0.6399$
& $0.6149$
& $0.6187$
& $0.6050$
& $0.5445$
& $0.6478$
& $0.6489$
& $0.6346$
& $0.5857$
& $\textbf{0.6491}$ \\
\midrule

\multirow{2}{*}{Urban100}
& PSNR
& $21.86$
& $21.92$
& $21.48$
& $21.66$
& $22.15$
& $21.93$
& $21.35$
& $21.36$
& $21.19$
& $20.28$
& $\mathbf{22.20}$ \\

& SSIM
& $0.5618$
& $0.5643$
& $0.5279$
& $0.5437$
& $0.5698$
& $0.5322$
& $0.6047$
& $0.6056$
& $0.5980$
& $0.5260$
& $\textbf{0.6134}$ \\
\bottomrule
\end{tabular}
}
\end{center}

\caption{
PnP-FBS super-resolution ($\times2$) under defocus blur using a disk
kernel with radius $5$, kernel size $11\times11$, and noise level
$\sigma_n=5/255$.
}
\label{tab:defocus_super1}
\end{table}

\begin{table}[!t]
\scriptsize
\setlength{\tabcolsep}{2.5pt}
\renewcommand{\arraystretch}{0.80}

\begin{center}
\resizebox{\linewidth}{!}{
\begin{tabular}{llccccccccccc}
\toprule
Dataset & Metric
& FBS
& DRS
& IDF
& Noise2VST
& DnCNN
& Restormer
& CRR-NN ($t$-step)
& CRR-NN (prox.)
& MWCNN
& CLIP
& \textbf{OURS} \\
\midrule

\multirow{2}{*}{CBSD68}
& PSNR
& $24.86$
& $24.92$
& $24.25$
& $24.55$
& $25.06$
& $24.37$
& $23.56$
& $23.57$
& $23.21$
& $22.73$
& $\mathbf{25.15}$ \\

& SSIM
& $0.6581$
& $0.6598$
& $0.6110$
& $0.6291$
& $0.6474$
& $0.5745$
& $0.6098$
& $0.6106$
& $0.5880$
& $0.5355$
& $\mathbf{0.6757}$ \\
\midrule

\multirow{2}{*}{Kodak24}
& PSNR
& $25.68$
& $25.74$
& $25.04$
& $25.41$
& $25.78$
& $25.04$
& $24.99$
& $25.00$
& $24.70$
& $23.88$
& $\mathbf{25.97}$ \\

& SSIM
& $0.6793$
& $0.6805$
& $0.6442$
& $0.6538$
& $0.6436$
& $0.5594$
& $0.6496$
& $0.6508$
& $0.6331$
& $0.5847$
& $\mathbf{0.6955}$ \\
\midrule

\multirow{2}{*}{Urban100}
& PSNR
& $22.72$
& $22.85$
& $22.06$
& $22.67$
& $23.15$
& $22.72$
& $21.46$
& $21.46$
& $21.28$
& $20.30$
& $\mathbf{23.19}$ \\

& SSIM
& $0.6563$
& $0.6632$
& $0.6103$
& $0.6445$
& $0.6377$
& $0.5684$
& $0.6102$
& $0.6111$
& $0.6013$
& $0.5255$
& $\mathbf{0.6696}$ \\
\bottomrule
\end{tabular}
}
\end{center}

\caption{
PnP-FBS super-resolution ($\times2$) under box blur using a
$9\times9$ uniform kernel and noise level $\sigma_n=5/255$.
}
\label{tab:deblur_sparse_pnp2}
\end{table}

\begin{table}[!t]
\scriptsize
\setlength{\tabcolsep}{2.5pt}
\renewcommand{\arraystretch}{0.80}

\begin{center}
\resizebox{\linewidth}{!}{
\begin{tabular}{llccccccccccc}
\toprule
Dataset & Metric
& FBS
& DRS
& IDF
& Noise2VST
& DnCNN
& Restormer
& CRRNN($t$-step)
& CRRNN(prox.)
& MWCNN
& CLIP
& \textbf{OURS} \\
\midrule

\multirow{2}{*}{CBSD68}
& PSNR
& $24.25$
& $24.31$
& $24.10$
& $24.50$
& $23.95$
& $23.41$
& $24.13$
& $23.95$
& $24.08$
& $23.50$
& $\mathbf{24.52}$ \\

& SSIM
& $0.6117$
& $0.6248$
& $0.6046$
& $0.6305$
& $0.5763$
& $0.5213$
& $0.6163$
& $0.5940$
& $0.6189$
& $0.5794$
& $\mathbf{0.6421}$ \\
\midrule

\multirow{2}{*}{Kodak24}
& PSNR
& $24.98$
& $25.07$
& $24.84$
& $\mathbf{25.31}$
& $24.52$
& $23.95$
& $25.00$
& $24.77$
& $25.11$
& $24.38$
& $25.25$ \\

& SSIM
& $0.6244$
& $0.6413$
& $0.6363$
& $0.6545$
& $0.5660$
& $0.5045$
& $0.6349$
& $0.6023$
& $0.6519$
& $0.6101$
& $\mathbf{0.6565}$ \\
\midrule

\multirow{2}{*}{Urban100}
& PSNR
& $21.62$
& $21.68$
& $21.30$
& $\mathbf{22.02}$
& $21.51$
& $21.19$
& $21.65$
& $21.62$
& $21.82$
& $21.10$
& $21.82$ \\

& SSIM
& $0.5994$
& $0.6141$
& $0.5921$
& $0.6299$
& $0.5636$
& $0.5166$
& $0.6082$
& $0.5892$
& $\mathbf{0.6309}$
& $0.5788$
& $0.6273$ \\
\bottomrule
\end{tabular}
}
\end{center}

\caption{
PnP-FBS super-resolution ($\times$4) under anisotropic Gaussian blur ($\sigma_x=2.5$, $\sigma_y=1.5$, $\theta=45^\circ$, kernel size $21\times21$).
}
\label{tab:aniso_super}
\end{table}

\begin{table}[!t]
\scriptsize
\setlength{\tabcolsep}{2.5pt}
\renewcommand{\arraystretch}{0.80}

\begin{center}
\resizebox{\linewidth}{!}{
\begin{tabular}{llccccccccccc}
\toprule
Dataset & Metric
& FBS
& DRS
& IDF
& Noise2VST
& DnCNN
& Restormer
& CRR-NN ($t$-step)
& CRR-NN (prox.)
& MWCNN
& CLIP
& \textbf{OURS} \\
\midrule

\multirow{2}{*}{CBSD68}
& PSNR
& $24.29$
& $24.35$
& $24.15$
& $24.54$
& $23.99$
& $23.43$
& $24.19$
& $24.01$
& $24.14$
& $23.55$
& $\mathbf{24.57}$ \\

& SSIM
& $0.6133$
& $0.6264$
& $0.6070$
& $0.6327$
& $0.5768$
& $0.5204$
& $0.6180$
& $0.5956$
& $0.6213$
& $0.5823$
& $\mathbf{0.6439}$ \\
\midrule

\multirow{2}{*}{Kodak24}
& PSNR
& $25.00$
& $25.09$
& $24.90$
& $\mathbf{25.33}$
& $24.53$
& $23.95$
& $25.03$
& $24.80$
& $25.15$
& $24.39$
& $25.27$ \\

& SSIM
& $0.6253$
& $0.6426$
& $0.6394$
& $0.6560$
& $0.5652$
& $0.5025$
& $0.6359$
& $0.6030$
& $0.6537$
& $0.6129$
& $\mathbf{0.6577}$ \\
\midrule

\multirow{2}{*}{Urban100}
& PSNR
& $21.67$
& $21.73$
& $21.36$
& $\mathbf{22.07}$
& $21.56$
& $21.22$
& $21.71$
& $21.68$
& $21.88$
& $21.15$
& $21.87$ \\

& SSIM
& $0.6024$
& $0.6172$
& $0.5954$
& $\mathbf{0.6432}$
& $0.5655$
& $0.5172$
& $0.6112$
& $0.5919$
& $0.6349$
& $0.5826$
& $0.6304$ \\
\bottomrule
\end{tabular}
}
\end{center}

\caption{
PnP-FBS super-resolution ($\times$4) under Gaussian blur ($\sigma=2.0$, kernel size $11\times11$).
}
\label{tab:gaussian_super}
\end{table}

\begin{figure}[t]
\begin{center}
\setlength{\tabcolsep}{3pt}
\renewcommand{\arraystretch}{1.0}
\scriptsize

\begin{tabular}{c c c c}

\multirow{3}{*}{
\begin{tabular}{c}
\includegraphics[width=0.22\linewidth]{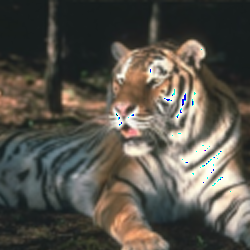} \\
Bicubic ($26.55$ / $0.8519$)
\end{tabular}
}
&
\includegraphics[width=0.22\linewidth]{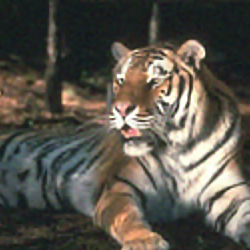} &
\includegraphics[width=0.22\linewidth]{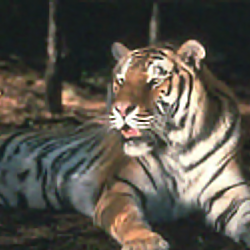} &
\includegraphics[width=0.22\linewidth]{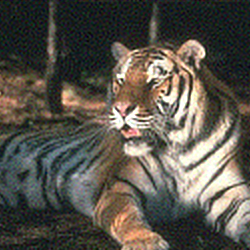} \\

&
D-FBS ($27.68$ / $0.8154$) &
D-DRS ($27.77$ / $0.8229$) &
Restormer ($26.21$ / $0.6671$) \\[6pt]

&
\includegraphics[width=0.22\linewidth]{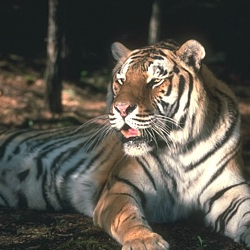} &
\includegraphics[width=0.22\linewidth]{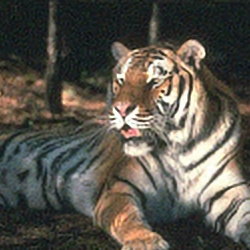} &
\includegraphics[width=0.22\linewidth]{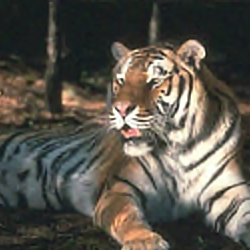} \\

&
Reference &
DnCNN ($28.27$ / $0.8042$) &
Ours ($28.52$ / $0.8549$)

\end{tabular}
\end{center}
\caption{$2\times$ super-resolution with PnP-FBS under Gaussian blur ($25\times25$, $\sigma=1.6$). Our contractive denoiser recovers sharper stripes and cleaner backgrounds, outperforming D-FBS/DRS and avoiding artifacts seen in Restormer.}
\label{fig:superres2}

\end{figure}

\begin{table}[!t]
\scriptsize
\setlength{\tabcolsep}{2.5pt}
\renewcommand{\arraystretch}{0.80}

\begin{center}
\resizebox{\linewidth}{!}{
\begin{tabular}{llccccccccccc}
\toprule
Dataset & Metric
& FBS
& DRS
& IDF
& Noise2VST
& DnCNN
& Restormer
& CRR-NN ($t$-step)
& CRR-NN (prox.)
& MWCNN
& CLIP
& \textbf{OURS} \\
\midrule

\multirow{2}{*}{CBSD68}
& PSNR
& $23.22$
& $23.29$
& $23.08$
& $23.22$
& $22.91$
& $22.68$
& $22.58$
& $22.40$
& $22.51$
& $22.11$
& $\mathbf{23.38}$ \\

& SSIM
& $0.5536$
& $0.5613$
& $0.5484$
& $0.5592$
& $0.5088$
& $0.4808$
& $0.5451$
& $0.5285$
& $0.5497$
& $0.5241$
& $\mathbf{0.5688}$ \\
\midrule

\multirow{2}{*}{Kodak24}
& PSNR
& $24.05$
& $24.16$
& $23.92$
& $24.17$
& $23.61$
& $23.39$
& $23.89$
& $23.74$
& $23.96$
& $23.42$
& $\mathbf{24.22}$ \\

& SSIM
& $0.5829$
& $0.5933$
& $0.5903$
& $0.5964$
& $0.5127$
& $0.4838$
& $0.5864$
& $0.5632$
& $0.5965$
& $0.5731$
& $\mathbf{0.5988}$ \\
\midrule

\multirow{2}{*}{Urban100}
& PSNR
& $20.38$
& $20.44$
& $20.15$
& $20.45$
& $20.30$
& $20.19$
& $20.19$
& $20.24$
& $20.26$
& $19.73$
& $\mathbf{20.49}$ \\

& SSIM
& $0.5243$
& $0.5337$
& $0.5208$
& $\mathbf{0.5409}$
& $0.4834$
& $0.4607$
& $0.5239$
& $0.5153$
& $0.5385$
& $0.4973$
& $0.5384$ \\
\bottomrule
\end{tabular}
}
\end{center}

\caption{
PnP-FBS super-resolution ($\times4$) under defocus blur
(disk kernel radius $=6$, kernel size $13\times13$).
}
\label{tab:defocus_super}
\end{table}
\begin{table}[!t]
\scriptsize
\setlength{\tabcolsep}{2.5pt}
\renewcommand{\arraystretch}{0.80}

\begin{center}
\resizebox{\linewidth}{!}{
\begin{tabular}{llccccccccccc}
\toprule
Dataset & Metric
& FBS
& DRS
& IDF
& Noise2VST
& DnCNN
& Restormer
& CRR-NN ($t$-step)
& CRR-NN (prox.)
& MWCNN
& CLIP
& \textbf{OURS} \\
\midrule

\multirow{2}{*}{CBSD68}
& PSNR
& $23.12$
& $23.19$
& $22.98$
& $23.10$
& $22.83$
& $22.61$
& $22.46$
& $22.28$
& $22.36$
& $22.04$
& $\mathbf{23.28}$ \\

& SSIM
& $0.5485$
& $0.5556$
& $0.5431$
& $0.5520$
& $0.5046$
& $0.4776$
& $0.5372$
& $0.5191$
& $0.5401$
& $0.5168$
& $\mathbf{0.5633}$ \\
\midrule

\multirow{2}{*}{Kodak24}
& PSNR
& $23.95$
& $24.05$
& $23.82$
& $24.05$
& $23.52$
& $23.31$
& $23.79$
& $23.65$
& $23.84$
& $23.35$
& $\mathbf{24.11}$ \\

& SSIM
& $0.5788$
& $0.5885$
& $0.5856$
& $0.5907$
& $0.5098$
& $0.4825$
& $0.5816$
& $0.5583$
& $0.5901$
& $0.5675$
& $\mathbf{0.5941}$ \\
\midrule

\multirow{2}{*}{Urban100}
& PSNR
& $20.29$
& $20.35$
& $20.04$
& $20.33$
& $20.23$
& $20.13$
& $20.10$
& $20.15$
& $20.14$
& $19.64$
& $\mathbf{20.41}$ \\

& SSIM
& $0.5191$
& $0.5278$
& $0.5145$
& $0.5330$
& $0.4803$
& $0.4583$
& $0.5173$
& $0.5091$
& $0.5293$
& $0.4886$
& $\mathbf{0.5333}$ \\
\bottomrule
\end{tabular}
}
\end{center}

\caption{
PnP-FBS super-resolution ($\times4$) under an $11\times11$ uniform blur.
}
\label{tab:box_super}
\end{table}

\subsection{Deblurring}
\label{sec:appendix_deblurring}
\begin{figure}[t]
\begin{center}
\setlength{\tabcolsep}{3pt}
\renewcommand{\arraystretch}{1.0}
\scriptsize

\begin{tabular}{c c c c}

\multirow{3}{*}{
\begin{tabular}{c}
\includegraphics[width=0.22\linewidth]{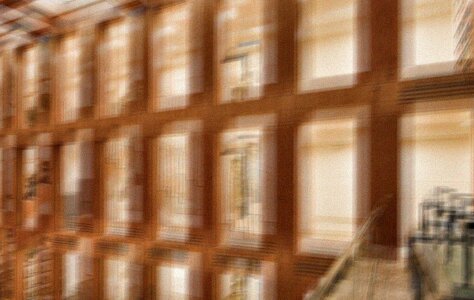} \\
Blurred ($10.29$ dB / $0.1003$)
\end{tabular}
}
&
\includegraphics[width=0.22\linewidth]{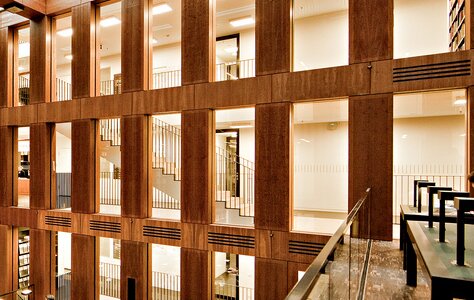} &
\includegraphics[width=0.22\linewidth]{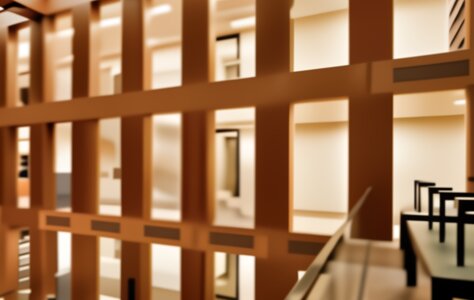} &
\includegraphics[width=0.22\linewidth]{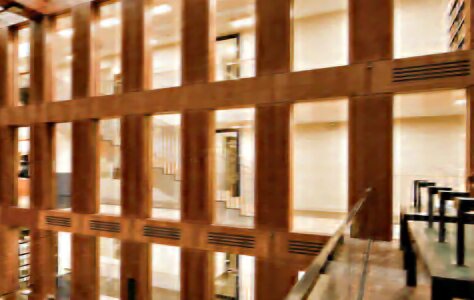} \\

&
Reference &
Noise2VST ($20.96$ dB / $0.6528$) &
D-FBS ($21.75$ dB / $0.6941$) \\[6pt]

&
\includegraphics[width=0.22\linewidth]{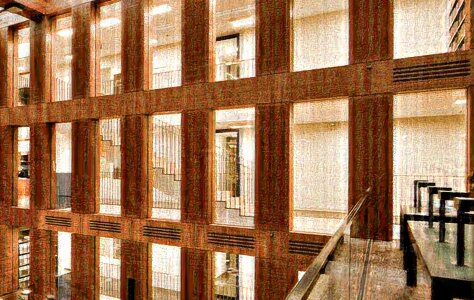} &
\includegraphics[width=0.22\linewidth]{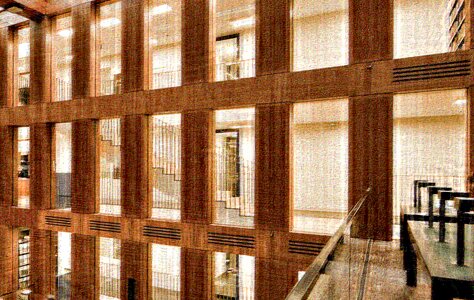} &
\includegraphics[width=0.22\linewidth]{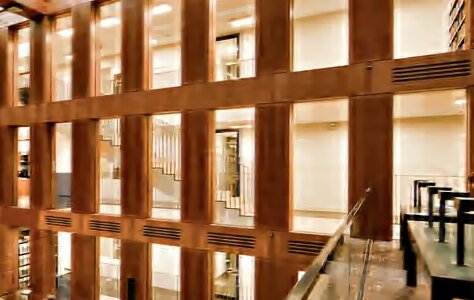} \\

&
DnCNN ($23.21$ dB / $0.7244$) &
Restormer ($22.60$ dB / $0.6804$) &
Ours ($23.31$ dB / $0.7610$)

\end{tabular}
\end{center}
\caption{PnP-FBS deblurring with Levin09 kernel (idx3). Our method achieves the best reconstruction and avoids artifacts seen in DnCNN and Restormer.}
\label{fig:kevin_blur_idx2}

\end{figure}

\begin{figure}[!t]
\begin{center}
\setlength{\tabcolsep}{3pt}
\renewcommand{\arraystretch}{1.0}
\scriptsize

\begin{tabular}{c c c c}

\multirow{3}{*}{
\begin{tabular}{c}
\includegraphics[width=0.22\linewidth]{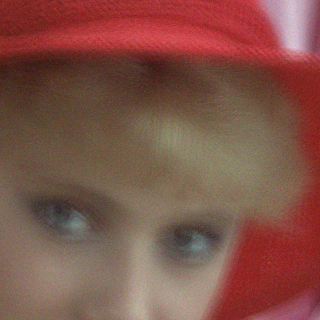} \\
Blurred ($22.62$ dB / $0.5130$)
\end{tabular}
}
&
\includegraphics[width=0.22\linewidth]{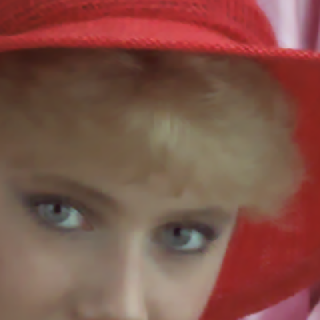} &
\includegraphics[width=0.22\linewidth]{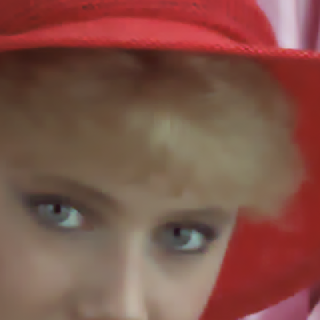} &
\includegraphics[width=0.22\linewidth]{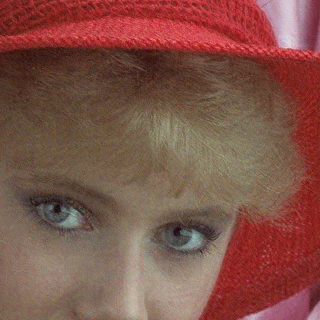} \\

&
D-FBS ($28.91$ dB / $0.7403$) &
D-DRS ($28.96$ dB / $0.7225$) &
Restormer ($30.12$ dB / $0.7629$) \\[6pt]

&
\includegraphics[width=0.22\linewidth]{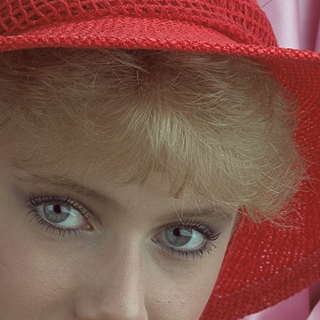} &
\includegraphics[width=0.22\linewidth]{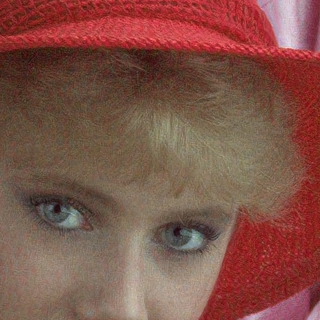} &
\includegraphics[width=0.22\linewidth]{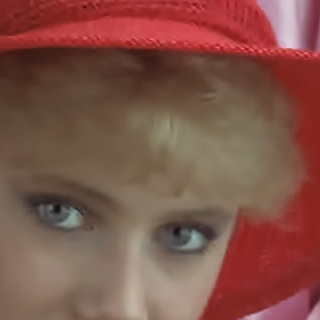} \\

&
Reference &
DnCNN ($29.71$ dB / $0.7245$) &
Ours ($29.85$ dB / $0.7594$)

\end{tabular}
\end{center}
\caption{Sparse-kernel deblurring ($15\times15$, $90\%$ zeros). Our method restores sharper facial details and cleaner reconstructions with fewer artifacts compared to other denoisers.}
\label{fig:sparse_blur}

\end{figure}

We next show the regularization capabilities of our denoiser, focusing on deblurring. The forward model is 
\(
\y = \B\x + \eta,
\)
where $\B$ is a known blur, and $\eta$ is AWGN for deblurring. We evaluate PnP-FBS deblurring under Gaussian, defocus, sparse, Levin kernel-1, Levin kernel-3, anisotropic Gaussian, diagonal motion, and box blur, as reported in Tables~\ref{tab:gaussian_blur}, \ref{tab:defocus_blurr}, \ref{tab:deblur_sparse_pnp1}, \ref{tab:deblur_levin_red1}, \ref{tab:levin_blur}, \ref{tab:aniso_blurr}, \ref{tab:motion_blurr}, and~\ref{tab:box_blurr}, respectively. We compare against the $1$-Lipschitz baselines D-FBS and D-DRS, the unconstrained models IDF, Noise2VST, DnCNN, and Restormer, CRR-NN ($t$-step and proximal variants), MWCNN, and CLIP. Across the evaluated datasets and blur settings, our contractive denoiser consistently achieves the best or competitive PSNR/SSIM, demonstrating that enforcing contraction does not compromise deblurring quality. Qualitative comparisons are shown in Figures~\ref{fig:kevin_blur_idx2}, \ref{fig:sparse_blur}, \ref{fig:motion_blur_pnp}, and \ref{fig:levin_idx3_deblur}. Across different blur types and textured scenes, our method restores sharper structures while avoiding the over-smoothing observed in $1$-Lipschitz baselines and the artifacts produced by unconstrained models.

\begin{table}[!t]
\scriptsize
\setlength{\tabcolsep}{2.5pt}
\renewcommand{\arraystretch}{0.80}

\begin{center}
\resizebox{\linewidth}{!}{
\begin{tabular}{llccccccccccc}
\toprule
Dataset & Metric
& FBS
& DRS
& IDF
& Noise2VST
& DnCNN
& Restormer
& CRR-NN ($t$-step)
& CRR-NN (prox.)
& MWCNN
& CLIP
& \textbf{OURS} \\
\midrule

\multirow{2}{*}{CBSD68}
& PSNR
& $25.79$
& $25.84$
& $25.04$
& $25.36$
& $25.75$
& $25.20$
& $25.89$
& $25.88$
& $25.16$
& $23.81$
& $\mathbf{26.03}$ \\

& SSIM
& $0.7170$
& $0.7163$
& $0.6576$
& $0.6795$
& $0.7167$
& $0.6486$
& $0.7178$
& $0.7172$
& $0.6714$
& $0.5981$
& $\mathbf{0.7329}$ \\
\midrule

\multirow{2}{*}{Kodak24}
& PSNR
& $26.48$
& $26.58$
& $25.76$
& $26.13$
& $26.55$
& $25.86$
& $26.60$
& $26.59$
& $26.06$
& $24.29$
& $\mathbf{26.76}$ \\

& SSIM
& $0.7313$
& $0.7300$
& $0.6856$
& $0.6976$
& $0.7159$
& $0.6330$
& $0.7338$
& $0.7333$
& $0.6947$
& $0.6230$
& $\mathbf{0.7454}$ \\
\midrule

\multirow{2}{*}{Urban100}
& PSNR
& $23.28$
& $23.41$
& $22.47$
& $23.20$
& $23.67$
& $23.26$
& $23.25$
& $23.17$
& $23.13$
& $21.25$
& $\mathbf{23.69}$ \\

& SSIM
& $0.7244$
& $0.7313$
& $0.6665$
& $0.7105$
& $0.7259$
& $0.6584$
& $0.7154$
& $0.7121$
& $0.6980$
& $0.6065$
& $\mathbf{0.7420}$ \\
\bottomrule
\end{tabular}
}
\end{center}

\caption{
PnP-FBS image deblurring under Gaussian blur
($\sigma=2.0$, kernel size $9\times9$) and noise level
$\sigma_n=5/255$.
}
\label{tab:gaussian_blur}
\end{table}

\begin{table}[!t]
\scriptsize
\setlength{\tabcolsep}{2.5pt}
\renewcommand{\arraystretch}{0.80}

\begin{center}
\resizebox{\linewidth}{!}{
\begin{tabular}{llccccccccccc}
\toprule
Dataset & Metric
& FBS
& DRS
& IDF
& Noise2VST
& DnCNN
& Restormer
& CRR-NN ($t$-step)
& CRR-NN (prox.)
& MWCNN
& CLIP
& \textbf{OURS} \\
\midrule

\multirow{2}{*}{CBSD68}
& PSNR
& $24.59$
& $24.66$
& $23.98$
& $24.17$
& $24.88$
& $23.88$
& $24.69$
& $24.69$
& $24.07$
& $23.08$
& $\mathbf{24.91}$ \\

& SSIM
& $0.6414$
& $0.6404$
& $0.5977$
& $0.6111$
& $0.6293$
& $0.5330$
& $0.6446$
& $0.6443$
& $0.6060$
& $0.5535$
& $\mathbf{0.6559}$ \\
\midrule

\multirow{2}{*}{Kodak24}
& PSNR
& $25.44$
& $25.54$
& $24.78$
& $25.10$
& $\mathbf{25.81}$
& $24.67$
& $25.60$
& $25.59$
& $25.08$
& $23.96$
& $\mathbf{25.81}$ \\

& SSIM
& $0.6686$
& $0.6672$
& $0.6344$
& $0.6416$
& $0.6245$
& $0.5152$
& $0.6736$
& $0.6734$
& $0.6403$
& $0.5913$
& $\mathbf{0.6825}$ \\
\midrule

\multirow{2}{*}{Urban100}
& PSNR
& $21.80$
& $21.98$
& $21.15$
& $21.70$
& $\mathbf{22.60}$
& $22.03$
& $21.96$
& $21.88$
& $21.65$
& $20.28$
& $22.54$ \\

& SSIM
& $0.6278$
& $0.6321$
& $0.5838$
& $0.6182$
& $0.6196$
& $0.5446$
& $0.6362$
& $0.6279$
& $0.6136$
& $0.5358$
& $\mathbf{0.6515}$ \\
\bottomrule
\end{tabular}
}
\end{center}

\caption{
PnP-FBS image deblurring under defocus blur using a disk kernel
with radius $5$, kernel size $11\times11$, and noise level
$\sigma_n=5/255$.
}
\label{tab:defocus_blurr}
\end{table}

\begin{table}[!t]
\scriptsize
\setlength{\tabcolsep}{2.5pt}
\renewcommand{\arraystretch}{0.80}

\begin{center}
\resizebox{\linewidth}{!}{
\begin{tabular}{llccccccccccc}
\toprule
Dataset & Metric
& FBS 
& DRS 
& IDF 
& Noise2VST 
& DnCNN 
& Restormer
& CRR-NN ($t$-step)
& CRR-NN (prox.)
& MWCNN
& CLIP
& \textbf{OURS} \\
\midrule

\multirow{2}{*}{CBSD68}
& PSNR
& $26.62$ & $27.03$ & $24.26$ & $24.72$
& $27.11$ & $25.35$
& $25.35$ & $25.27$
& $23.71$ & $22.35$
& $\mathbf{28.40}$ \\

& SSIM
& $0.7403$ & $0.7438$ & $0.6009$ & $0.6130$
& $0.7179$ & $0.7013$
& $0.6772$ & $0.6724$
& $0.5738$ & $0.5070$
& $\mathbf{0.8063}$ \\
\midrule

\multirow{2}{*}{Kodak24}
& PSNR
& $27.45$ & $27.91$ & $25.24$ & $25.82$
& $27.45$ & $25.24$
& $26.75$ & $26.67$
& $25.29$ & $23.50$
& $\mathbf{29.21}$ \\

& SSIM
& $0.7456$ & $0.7451$ & $0.6430$ & $0.6490$
& $0.6772$ & $0.6692$
& $0.7121$ & $0.7082$
& $0.6334$ & $0.5602$
& $\mathbf{0.8102}$ \\
\midrule

\multirow{2}{*}{Urban100}
& PSNR
& $24.59$ & $25.48$ & $22.56$ & $23.33$
& $25.86$ & $25.26$
& $23.29$ & $23.19$
& $22.19$ & $19.79$
& $\mathbf{26.33}$ \\

& SSIM
& $0.7600$ & $0.7741$ & $0.6390$ & $0.7369$
& $0.7419$ & $0.7397$
& $0.7017$ & $0.6971$
& $0.6309$ & $0.4953$
& $\mathbf{0.8099}$ \\
\bottomrule
\end{tabular}
}
\end{center}

\caption{
PnP-FBS image deblurring with a $90\%$ sparse kernel and noise level
$\sigma_n=5/255$.
}
\label{tab:deblur_sparse_pnp1}
\end{table}

\begin{table}
\scriptsize
\setlength{\tabcolsep}{2.5pt}
\renewcommand{\arraystretch}{0.80}

\begin{center}
\resizebox{\linewidth}{!}{
\begin{tabular}{llccccccccccc}
\toprule
Dataset & Metric
& FBS 
& DRS 
& IDF 
& Noise2VST 
& DnCNN 
& Restormer
& CRR-NN ($t$-step)
& CRR-NN (prox.)
& MWCNN
& CLIP
& \textbf{OURS} \\
\midrule

\multirow{2}{*}{CBSD68}
& PSNR
& $26.14$ & $26.44$ & $24.47$ & $25.05$
& $26.41$ & $26.33$
& $26.70$ & $26.63$
& $25.16$ & $23.78$
& $\mathbf{27.50}$ \\

& SSIM
& $0.7276$ & $0.7300$ & $0.6251$ & $0.6479$
& $0.6800$ & $0.7161$
& $0.7447$ & $0.7415$
& $0.6513$ & $0.5820$
& $\mathbf{0.7797}$ \\
\midrule

\multirow{2}{*}{Kodak24}
& PSNR
& $27.03$ & $27.38$ & $25.30$ & $26.11$
& $26.83$ & $27.74$
& $27.84$ & $27.78$
& $26.51$ & $24.78$
& $\mathbf{28.53}$ \\

& SSIM
& $0.7412$ & $0.7410$ & $0.6573$ & $0.6779$
& $0.6344$ & $0.7339$
& $0.7619$ & $0.7593$
& $0.6905$ & $0.6166$
& $\mathbf{0.7850}$ \\
\midrule

\multirow{2}{*}{Urban100}
& PSNR
& $23.87$ & $24.39$ & $22.09$ & $23.66$
& $25.02$ & $25.50$
& $24.72$ & $24.64$
& $24.09$ & $21.67$
& $\mathbf{25.63}$ \\

& SSIM
& $0.7414$ & $0.7557$ & $0.6263$ & $0.7126$
& $0.6908$ & $0.7705$
& $0.7635$ & $0.7606$
& $0.7199$ & $0.6035$
& $\mathbf{0.8009}$ \\
\bottomrule
\end{tabular}
}
\end{center}

\caption{
PnP-FBS image deblurring with kernel 1 from
Levin et al.~\cite{levin2009understanding} and noise level
$\sigma_n=5/255$.
}
\label{tab:deblur_levin_red1}
\end{table}

\begin{figure*}[!t]
\begin{center}
\setlength{\tabcolsep}{3pt}
\renewcommand{\arraystretch}{1.0}
\scriptsize

\begin{tabular}{c c c c}

\multirow{3}{*}{
\begin{tabular}{c}
\includegraphics[width=0.22\linewidth]{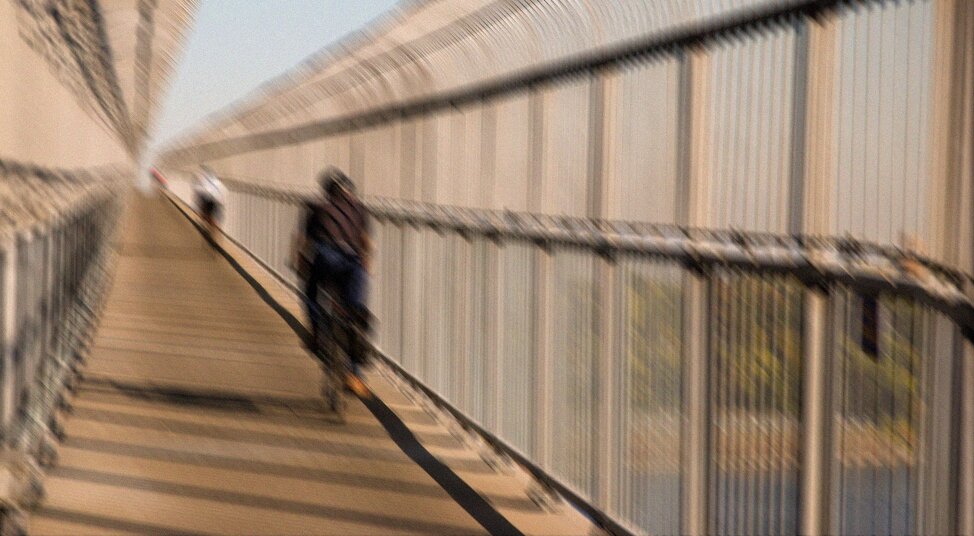} \\
Blurred ($13.14$ / $0.2021$)
\end{tabular}
}
&
\includegraphics[width=0.22\linewidth]{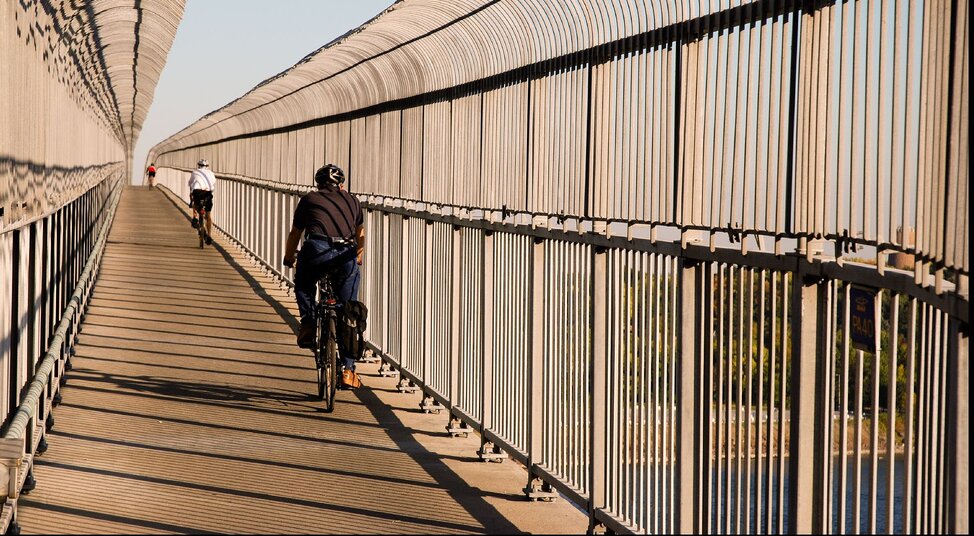} &
\includegraphics[width=0.22\linewidth]{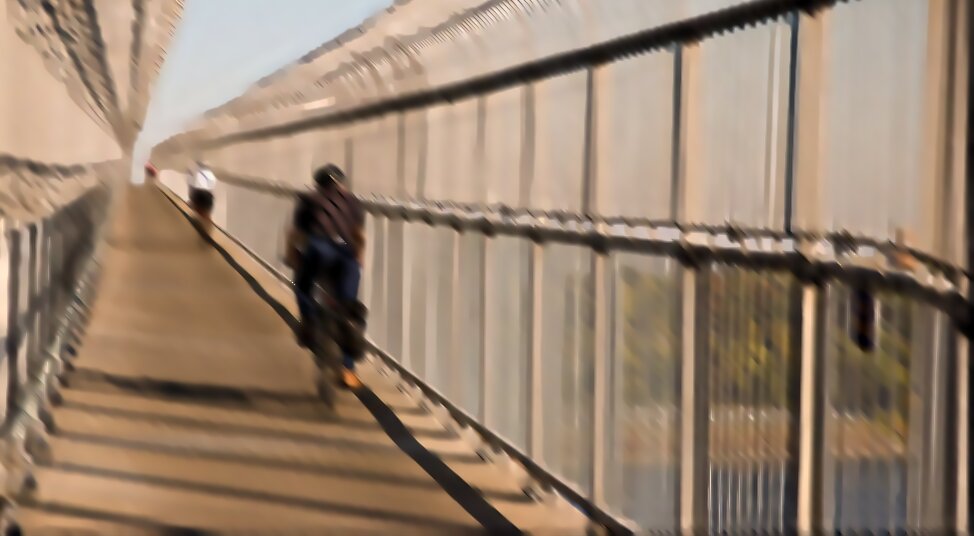} &
\includegraphics[width=0.22\linewidth]{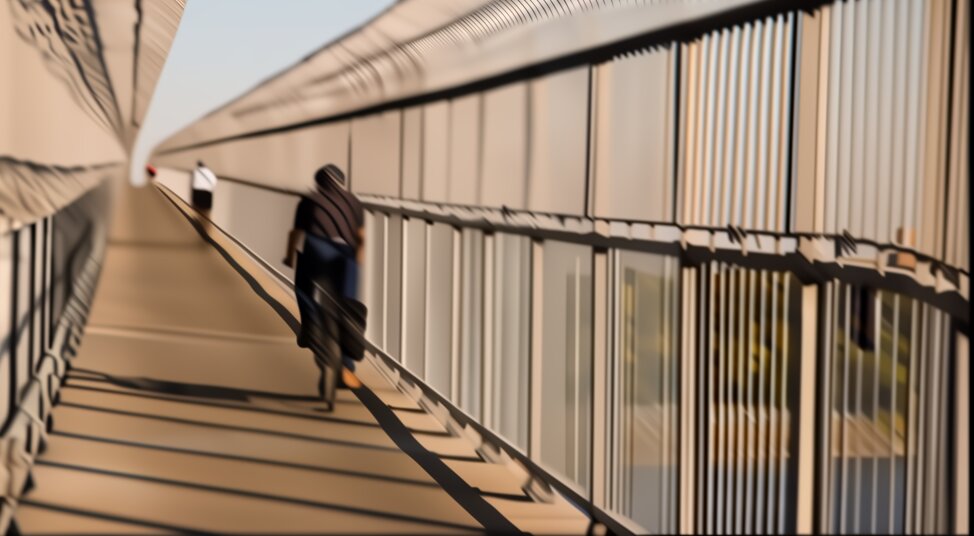} \\

&
Reference &
IDF ($15.49$ / $0.3337$) &
Noise2VST ($17.08$ / $0.4821$) \\[6pt]

&
\includegraphics[width=0.22\linewidth]{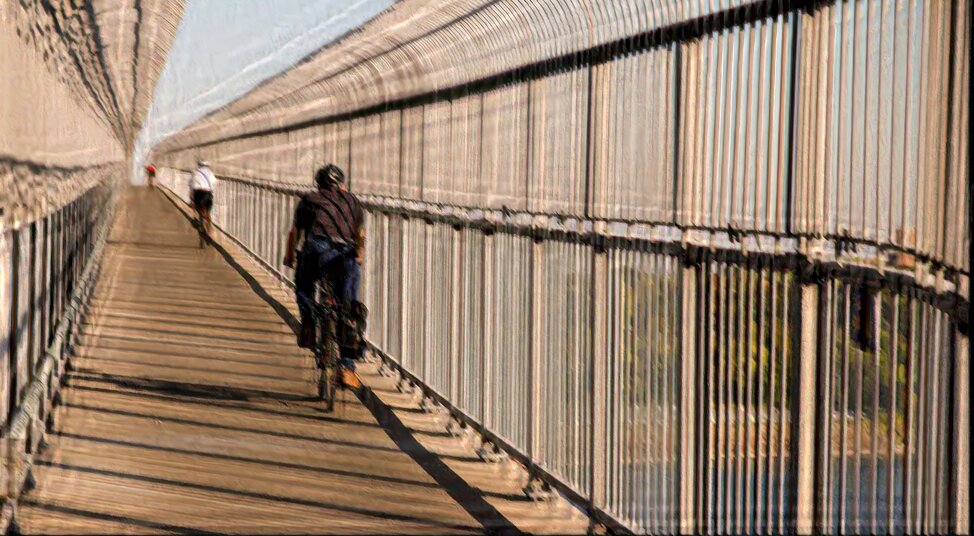} &
\includegraphics[width=0.22\linewidth]{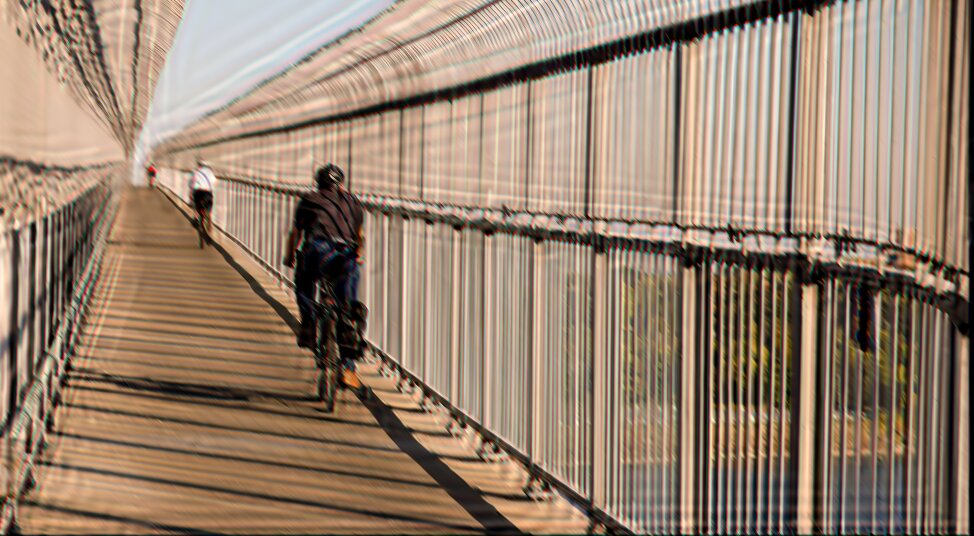} &
\includegraphics[width=0.22\linewidth]{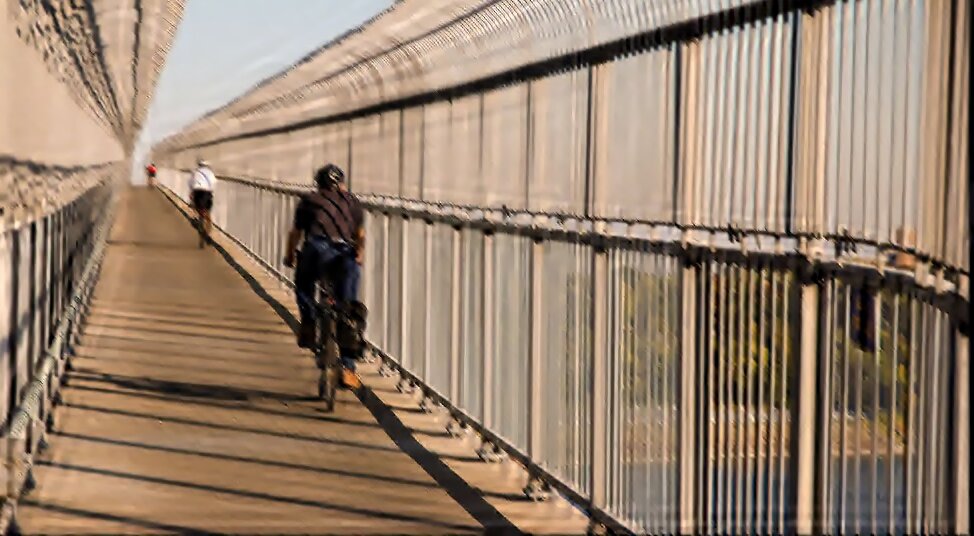} \\

&
DnCNN ($18.63$ / $0.6227$) &
Restormer ($17.74$ / $0.5552$) &
Ours ($18.75$ / $0.5429$)

\end{tabular}
\end{center}
\caption{PnP-FBS deblurring under diagonal motion blur ($15\times15$). Our method restores clearer details while avoiding artifacts in DnCNN/Restormer and the over-smoothing of Noise2VST and IDF.}
\label{fig:motion_blur_pnp}

\end{figure*}

\begin{table}
\scriptsize
\setlength{\tabcolsep}{2.5pt}
\renewcommand{\arraystretch}{0.80}

\begin{center}
\resizebox{\linewidth}{!}{
\begin{tabular}{llccccccccccc}
\toprule
Dataset & Metric
& FBS
& DRS
& IDF
& Noise2VST
& DnCNN
& Restormer
& CRR-NN ($t$-step)
& CRR-NN (prox.)
& MWCNN
& CLIP
& \textbf{OURS} \\
\midrule

\multirow{2}{*}{CBSD68}
& PSNR
& $25.98$
& $26.28$
& $24.22$
& $24.86$
& $26.17$
& $24.15$
& $26.85$
& $26.82$
& $25.56$
& $24.17$
& $\mathbf{27.30}$ \\

& SSIM
& $0.7146$
& $0.7181$
& $0.6079$
& $0.6329$
& $0.6691$
& $0.5626$
& $0.7511$
& $0.7502$
& $0.6830$
& $0.6083$
& $\mathbf{0.7683}$ \\
\midrule

\multirow{2}{*}{Kodak24}
& PSNR
& $26.90$
& $27.27$
& $25.08$
& $25.94$
& $26.59$
& $24.61$
& $27.69$
& $27.67$
& $26.55$
& $24.81$
& $\mathbf{28.33}$ \\

& SSIM
& $0.7303$
& $0.7317$
& $0.6447$
& $0.6643$
& $0.6249$
& $0.5188$
& $0.7635$
& $0.7629$
& $0.7078$
& $0.6298$
& $\mathbf{0.7767}$ \\
\midrule

\multirow{2}{*}{Urban100}
& PSNR
& $23.86$
& $24.38$
& $21.51$
& $23.50$
& $24.77$
& $23.45$
& $24.81$
& $24.77$
& $24.19$
& $21.93$
& $\mathbf{25.08}$ \\

& SSIM
& $0.7159$
& $0.7384$
& $0.6097$
& $0.6919$
& $0.6747$
& $0.5948$
& $0.7677$
& $0.7671$
& $0.7314$
& $0.6285$
& $\mathbf{0.7743}$ \\
\bottomrule
\end{tabular}
}
\end{center}

\caption{
PnP-FBS image deblurring with kernel 3 from
Levin et al.~\cite{levin2009understanding} and noise level
$\sigma_n=5/255$.
}
\label{tab:levin_blur}
\end{table}

\begin{table}[!t]
\scriptsize
\setlength{\tabcolsep}{2.5pt}
\renewcommand{\arraystretch}{0.80}

\begin{center}
\resizebox{\linewidth}{!}{
\begin{tabular}{llccccccccccc}
\toprule
Dataset & Metric
& FBS
& DRS
& IDF
& Noise2VST
& DnCNN
& Restormer
& CRR-NN ($t$-step)
& CRR-NN (prox.)
& MWCNN
& CLIP
& \textbf{OURS} \\
\midrule

\multirow{2}{*}{CBSD68}
& PSNR
& $25.81$
& $25.88$
& $24.96$
& $25.47$
& $25.87$
& $24.96$
& $25.91$
& $25.91$
& $25.33$
& $23.59$
& $\mathbf{26.14}$ \\

& SSIM
& $0.7246$
& $0.7249$
& $0.6608$
& $0.6853$
& $0.7180$
& $0.6251$
& $0.7264$
& $0.7259$
& $0.6787$
& $0.6000$
& $\mathbf{0.7441}$ \\
\midrule

\multirow{2}{*}{Kodak24}
& PSNR
& $26.70$
& $26.78$
& $25.81$
& $26.37$
& $26.78$
& $25.72$
& $26.84$
& $26.83$
& $26.37$
& $24.02$
& $\mathbf{27.03}$ \\

& SSIM
& $0.7469$
& $0.7467$
& $0.6928$
& $0.7106$
& $0.7190$
& $0.6112$
& $0.7507$
& $0.7503$
& $0.7102$
& $0.6289$
& $\mathbf{0.7657}$ \\
\midrule

\multirow{2}{*}{Urban100}
& PSNR
& $23.01$
& $23.17$
& $22.10$
& $23.16$
& $23.32$
& $22.80$
& $23.17$
& $23.13$
& $23.16$
& $20.93$
& $\mathbf{23.36}$ \\

& SSIM
& $0.7080$
& $0.7115$
& $0.6496$
& $0.7006$
& $0.6945$
& $0.6165$
& $0.7144$
& $0.7102$
& $0.7008$
& $0.6015$
& $\mathbf{0.7280}$ \\
\bottomrule
\end{tabular}
}
\end{center}

\caption{
PnP-FBS image deblurring under anisotropic Gaussian blur
(kernel size $21\times21$, $\sigma_x=3.0$, $\sigma_y=1.0$)
and noise level $\sigma_n=5/255$.
}
\label{tab:aniso_blurr}
\end{table}

\begin{table}[!t]
\scriptsize
\setlength{\tabcolsep}{2.5pt}
\renewcommand{\arraystretch}{0.80}

\begin{center}
\resizebox{\linewidth}{!}{
\begin{tabular}{llccccccccccc}
\toprule
Dataset & Metric
& FBS
& DRS
& IDF
& Noise2VST
& DnCNN
& Restormer
& CRR-NN ($t$-step)
& CRR-NN (prox.)
& MWCNN
& CLIP
& \textbf{OURS} \\
\midrule

\multirow{2}{*}{CBSD68}
& PSNR
& $24.47$
& $24.63$
& $23.46$
& $23.72$
& $\mathbf{25.08}$
& $23.96$
& $24.51$
& $24.49$
& $23.64$
& $23.13$
& $24.98$ \\

& SSIM
& $0.6538$
& $0.6570$
& $0.5857$
& $0.5945$
& $\mathbf{0.6885}$
& $0.6473$
& $0.6551$
& $0.6532$
& $0.5925$
& $0.5652$
& $0.6851$ \\
\midrule

\multirow{2}{*}{Kodak24}
& PSNR
& $25.31$
& $25.51$
& $24.36$
& $24.79$
& $\mathbf{26.10}$
& $24.96$
& $25.45$
& $25.43$
& $24.84$
& $23.99$
& $25.89$ \\

& SSIM
& $0.6748$
& $0.6762$
& $0.6250$
& $0.6320$
& $0.6962$
& $0.6640$
& $0.6791$
& $0.6776$
& $0.6342$
& $0.5932$
& $\mathbf{0.7012}$ \\
\midrule

\multirow{2}{*}{Urban100}
& PSNR
& $21.88$
& $22.11$
& $20.81$
& $21.83$
& $22.49$
& $22.23$
& $21.74$
& $21.69$
& $21.48$
& $20.34$
& $\mathbf{22.58}$ \\

& SSIM
& $0.6455$
& $0.6562$
& $0.5878$
& $0.6250$
& $0.6657$
& $0.6325$
& $0.6325$
& $0.6312$
& $0.6067$
& $0.5440$
& $\mathbf{0.6824}$ \\
\bottomrule
\end{tabular}
}
\end{center}

\caption{
PnP-FBS image deblurring under diagonal motion blur using a
$15\times15$ kernel with non-zero values along the main diagonal
and noise level $\sigma_n=5/255$.
}
\label{tab:motion_blurr}
\end{table}

\begin{table}[!t]
\scriptsize
\setlength{\tabcolsep}{2.5pt}
\renewcommand{\arraystretch}{0.80}

\begin{center}
\resizebox{\linewidth}{!}{
\begin{tabular}{llccccccccccc}
\toprule
Dataset & Metric
& FBS
& DRS
& IDF
& Noise2VST
& DnCNN
& Restormer
& CRR-NN ($t$-step)
& CRR-NN (prox.)
& MWCNN
& CLIP
& \textbf{OURS} \\
\midrule

\multirow{2}{*}{CBSD68}
& PSNR
& $24.73$
& $24.83$
& $24.01$
& $24.25$
& $24.96$
& $23.63$
& $24.85$
& $24.84$
& $24.14$
& $23.08$
& $\mathbf{25.12}$ \\

& SSIM
& $0.6478$
& $0.6470$
& $0.5980$
& $0.6123$
& $0.6282$
& $0.5119$
& $0.6512$
& $0.6509$
& $0.6070$
& $0.5527$
& $\mathbf{0.6663}$ \\
\midrule

\multirow{2}{*}{Kodak24}
& PSNR
& $25.63$
& $25.77$
& $24.84$
& $25.22$
& $25.96$
& $24.42$
& $25.83$
& $25.83$
& $25.20$
& $23.96$
& $\mathbf{26.11}$ \\

& SSIM
& $0.6748$
& $0.6736$
& $0.6353$
& $0.6434$
& $0.6244$
& $0.4908$
& $0.6805$
& $0.6802$
& $0.6421$
& $0.5910$
& $\mathbf{0.6930}$ \\
\midrule

\multirow{2}{*}{Urban100}
& PSNR
& $22.03$
& $22.24$
& $21.21$
& $21.92$
& $22.66$
& $22.06$
& $22.18$
& $22.16$
& $21.85$
& $20.28$
& $\mathbf{22.69}$ \\

& SSIM
& $0.6535$
& $0.6645$
& $0.5950$
& $0.6268$
& $0.6240$
& $0.5261$
& $0.6476$
& $0.6471$
& $0.6220$
& $0.5361$
& $\mathbf{0.6672}$ \\
\bottomrule
\end{tabular}
}
\end{center}

\caption{
PnP-FBS image deblurring under box blur using a $9\times9$ uniform
kernel and noise level $\sigma_n=5/255$.
}
\label{tab:box_blurr}
\end{table}

\begin{figure*}[t]
\begin{center}
\setlength{\tabcolsep}{3pt}
\renewcommand{\arraystretch}{1.0}
\scriptsize

\begin{tabular}{c c c c}

\multirow{3}{*}{
\begin{tabular}{c}
\includegraphics[width=0.22\linewidth]{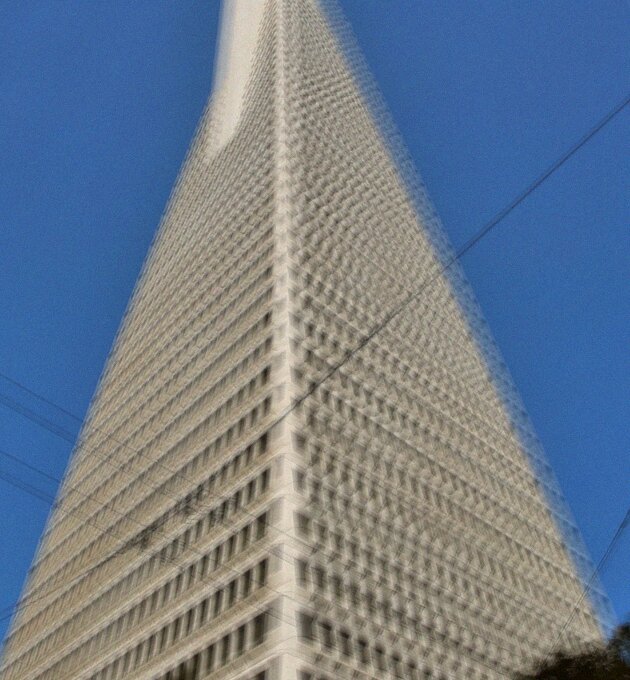} \\
Blurred ($12.20$ / $0.3653$)
\end{tabular}
}
&
\includegraphics[width=0.22\linewidth]{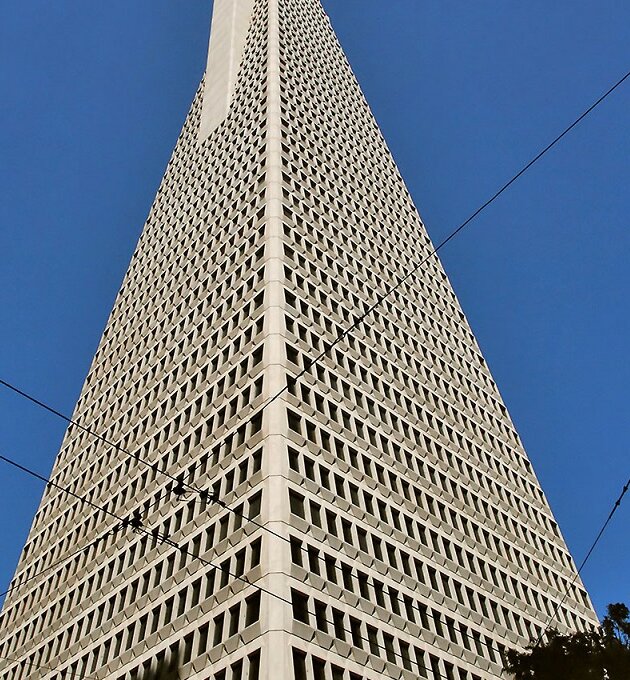} &
\includegraphics[width=0.22\linewidth]{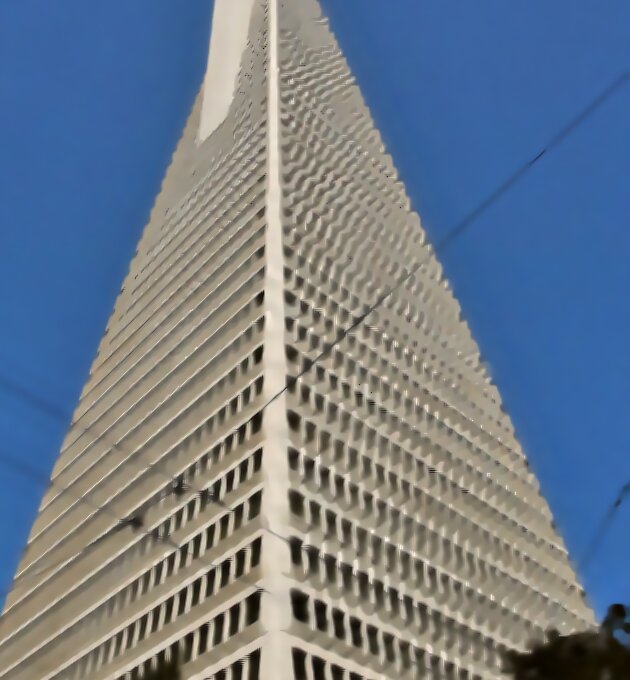} &
\includegraphics[width=0.22\linewidth]{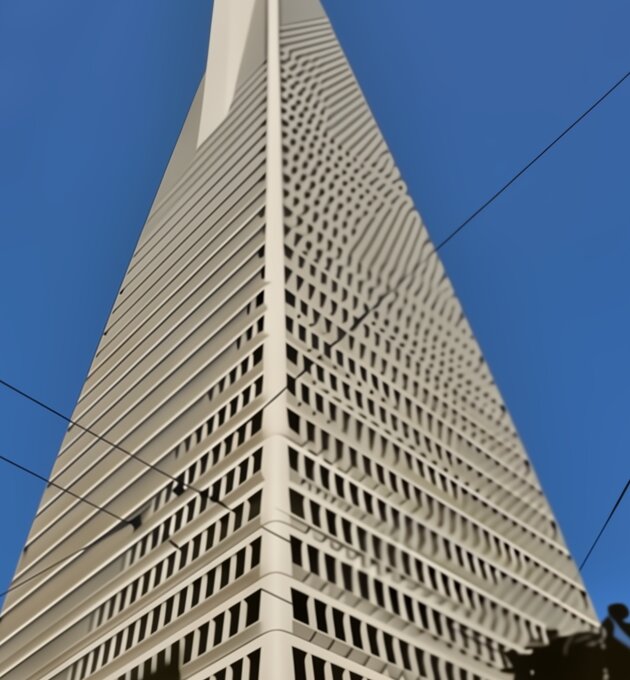} \\

&
Reference &
IDF ($14.89$ / $0.5567$) &
Noise2VST ($15.70$ / $0.6413$) \\[6pt]

&
\includegraphics[width=0.22\linewidth]{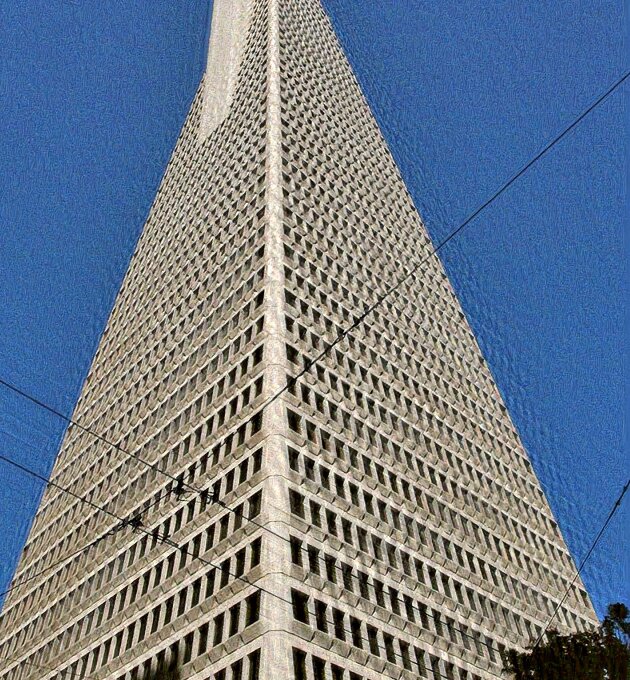} &
\includegraphics[width=0.22\linewidth]{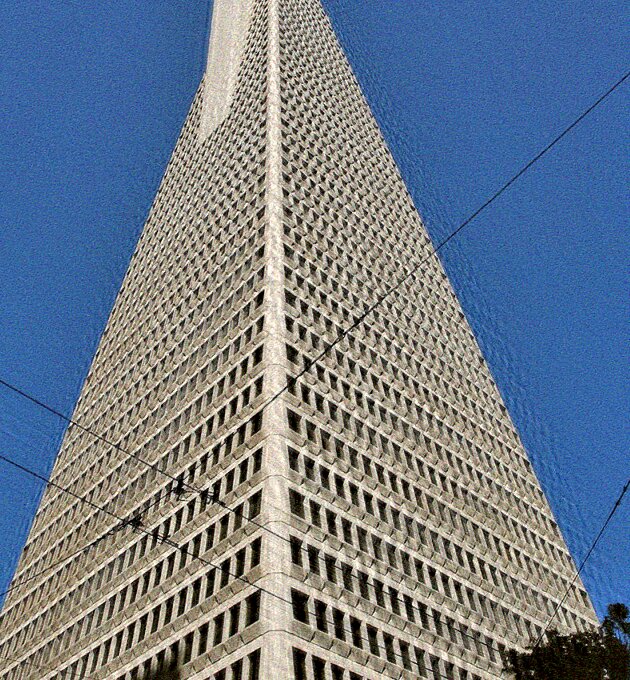} &
\includegraphics[width=0.22\linewidth]{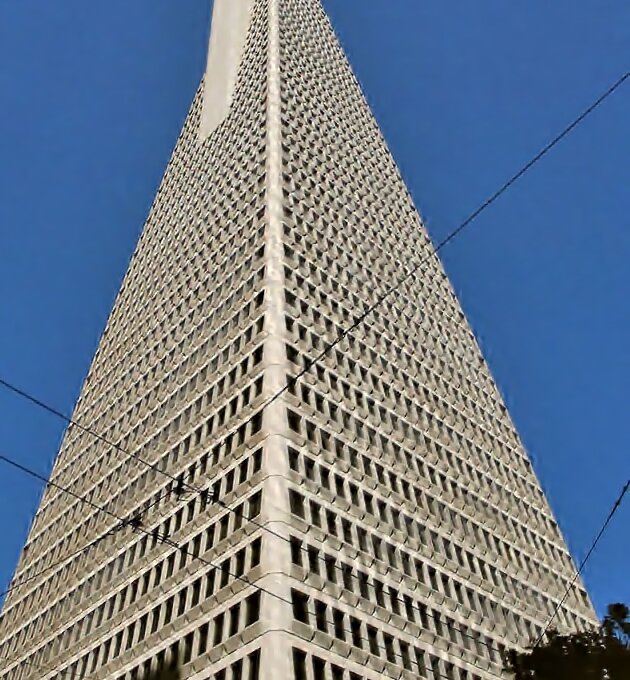} \\

&
DnCNN ($19.62$ / $0.6326$) &
Restormer ($18.96$ / $0.5767$) &
Ours ($20.09$ / $0.8353$)

\end{tabular}
\end{center}
\caption{PnP-FBS deblurring with the Levin09 kernel (idx3). Our contractive denoiser reconstructs sharper structures while avoiding ringing artifacts seen around the edges in DnCNN and Restormer and the over-smoothing of IDF and Noise2VST.}
\label{fig:levin_idx3_deblur}

\end{figure*}

\end{document}